\documentclass[journal]{IEEEtran}
\ifCLASSINFOpdf

\else

\fi
\usepackage{textcomp,booktabs}       
\usepackage[usenames,dvipsnames]{color}  
\usepackage{colortbl}  
\definecolor{mygray}{gray}{.9}  
\definecolor{mypink}{rgb}{.99,.91,.95}  
\definecolor{mycyan}{cmyk}{.3,0,0,0}  
\usepackage{amssymb}
\usepackage{fixltx2e}
\usepackage{url}
\usepackage{graphicx}
\usepackage{amsmath}
\usepackage{amsfonts}
\usepackage{mathtools}
\usepackage{nicefrac}
\usepackage{caption}
\usepackage{subcaption}
\usepackage{cite}
\usepackage{filecontents}
\usepackage{xr}
\usepackage[ruled,linesnumbered,vlined]{algorithm2e}
\DeclareMathOperator*{\argmin}{arg\,min}
\begin{document}
	\title{Nonlinear Dimensionality Reduction for Data Visualization: An Unsupervised Fuzzy Rule-based Approach}
	\author{Suchismita~Das
		and Nikhil~R.~Pal,~\IEEEmembership{Fellow,~IEEE}
		\thanks{Suchismita~Das and  Nikhil~R.~Pal are with the Electronics and Communication Sciences Unit, Indian Statistical Institute, Calcutta, 700108, India e-mail: (\{suchismitasimply,nrpal59\}@gmail.com).}
		\thanks{Manuscript received XXXX XX, 2020; revised XXXX XX, XXXX.}}
	\markboth{}%
	{Das \MakeLowercase{\textit{et al.}}: Nonlinear Dimensionality Reduction for Data Visualization: An Unsupervised Fuzzy Rule-based Approach}

	\maketitle
	
	\begin{abstract}
		 Here, we propose an unsupervised fuzzy rule-based dimensionality reduction method primarily for data visualization. It considers the following important issues relevant to dimensionality reduction-based data visualization: (i) preservation of neighborhood relationships, (ii) handling data on a non-linear manifold, (iii) the capability of predicting projections for new test data points, (iv) interpretability of the system, and (v) the ability to reject test points if required. For this, we use a first-order Takagi-Sugeno type model. We generate rule antecedents using clusters in the input data. In this context, we also propose a new variant of the Geodesic $c$-means clustering algorithm. We estimate the rule parameters by minimizing an error function that preserves the inter-point geodesic distances (distances over the manifold) as Euclidean distances on the projected space. We apply the proposed method on three synthetic and three real-world data sets and visually compare the results with four other standard data visualization methods. The obtained results show that the proposed method behaves desirably and performs better than or comparable to the methods compared with. The proposed method is found to be robust to the initial conditions. The predictability of the proposed method for test points is validated by experiments. We also assess the ability of our method to reject output points when it should. Then, we extend this concept to provide a general framework for learning an \emph{unsupervised} fuzzy model for data projection with different objective functions. To the best of our knowledge, this is the first attempt to manifold learning using unsupervised fuzzy modeling.
	\end{abstract}
	
	\IEEEpeerreviewmaketitle
	
	\section{Introduction}\label{sec:introduction}
	\IEEEPARstart{V}{isualization} is one of the prominent exploratory data analysis schemes as it provides insights into the data. We come across high dimensional data in various real-world problems related to, as examples,  finance, meteorology, computer vision, medical imaging, multimedia information processing, and text mining \cite{alvarez2017kernel,meng2008nonlinear,talwalkar2008large,van2009dimensionality,wang2018perception}. However, plotting more than three dimensions directly is not feasible. Data visualization schemes provide ways to visualize high dimensional data. They can be roughly divided into two categories. Methods that fall under the first category provide some mechanism to display more than two dimensions graphically. The Chernoff faces \cite{chernoff1973use} is an example of this category. The second category reduces the dimensionality of the data to two or three.  They aim to represent the data in a lower dimension keeping the `relevant' information of the original data as intact as possible. Note that, the schemes under the first category do not explicitly extract/summarize any information of the data. On the other hand, dimensionality reduction-based schemes, try to carry the information present in the original data to its lower dimensional representation. Dimensionality reduction for data visualization via projection can be achieved in many ways such as Principal Component Analysis (PCA) \cite{duda2001pattern}, Multi-Dimensional Scaling (MDS) \cite{borg2003modern}, and manifold learning \cite{cayton2005algorithms}. Some of these methods are linear; for example, PCA, canonical correlation analysis \cite{hotelling1936relations}, linear discriminant analysis \cite{fisher1936use}, factor analysis \cite{akaike1987factor}, locality preserving projections \cite{he2004locality}, which are not suitable if the data set has non-linear structures. Note that, a special case of linear projection is feature selection \cite{tuv2009feature, chandrashekar2014survey,nag2016multiobjective, chung2018feature}. When data are projected by feature selection, the features in the reduced space maintain their original identity, while in other cases of projections, the new features are difficult to interpret. Non-linear projections such as Sammon's projection \cite{sammon1969nonlinear} and manifold learning algorithms preserve some geometric properties of the data. Although the physical meaning of such features is difficult to comprehend, they produce more useful visualization. A class of non-linear data projection methods for visualization can be categorized as manifold learning algorithms. In a p-dimensional manifold, each point has a local neighborhood that is homeomorphic to the Euclidean space of the same dimension. In manifold learning for data visualization, the objective is to produce a low dimensional (usually two or three) representation of the high dimensional data by preserving the local neighborhood (local geometry). This helps to understand the intrinsic dimensionality of the manifold.  There have been many attempts to manifold learning \cite{roweis2000nonlinear,tenenbaum2000global,donoho2003hessian,belkin2003laplacian}. 
	
	Whether linear or non-linear, dimensionality reduction methods optimize some objective function to get a low dimensional representation. Objective functions could be either convex or non-convex. Dimensionality reduction methods having convex objectives do not suffer from getting stuck at local optima. However, the study in \cite{van2009dimensionality} suggests that for dimensionality reduction, convex methods do not necessarily perform better than non-convex methods. This study also claims that non-convex dimensionality reduction methods, like multi-layer auto-encoders \cite{hinton2006reducing}, perform well.  
	Another important aspect is whether the dimensionality reduction method is equipped with predictability or not. Methods that are parametric such as PCA and multi-layer auto-encoder based methods provide a direct mapping from the high dimensional space to the low dimensional space. Thus the trained parametric model can produce the lower dimensional representation for any test points. With non-parametric methods, such predictions for new points are not possible.  For example, methods such as local linear embedding (LLE)\cite{roweis2000nonlinear}, isometric mapping (ISOMAP)\cite{tenenbaum2000global} and $t$-distributed stochastic neighbor embedding ($t$-SNE) \cite{maaten2008visualizing} are non-parametric and as such do not have predictability. However, in \cite{bengio2004out} authors proposed some out of sample extension of methods like LLE and ISOMAP. 
	
	Fuzzy rule-based models are parametric models that are extensively used in different machine learning tasks such as control, classification, and forecasting \cite{takagi1985fuzzy,ng1997fuzzy,bezdek1984fcm}. They learn a function from a given set of training points and directly predict outputs for test points. Fuzzy rule-based systems can handle non-linear relationship between input and output. Moreover, they are easy to understand and develop. They provide systems that are ``explainable"/``comprehensible" at least to some extent. They store the knowledge as a set of easy-to-interpret fuzzy rules. For these characteristics, fuzzy rule-based systems seem to be a suitable candidate for implementing dimensionality reduction based data visualization models. However, the literature is not rich in this area. Apart from the work in \cite{pal2002fuzzy} we could not find any investigation employing fuzzy rule-based systems for data visualization applications through dimensionality reduction.
	
	Here, we have proposed a fuzzy rule-based dimensionality reduction method for visualization of data. We have used a first order Takagi Sugeno (TS) \cite{takagi1985fuzzy} type model. In the proposed model, a fuzzy rule represents a small region of the input space and its associated output region is represented by a hyper-plane. The rule antecedents model the local geometry of the input data while the consequents model the lower dimensional representation of the input. The rule parameters are learned by minimizing an objective function that preserves approximate distances over the manifold on or near which the data lie. The contributions in this work can be summarized as follows:
	\begin{enumerate}
		\item To the best of our knowledge, this is the first unsupervised method based on fuzzy rules for nonlinear manifold learning.
		\item Most unsupervised methods cannot project test data points (points not used in the training data), but ours can.
		\item To the best of our knowledge, this is the first method of nonlinear data projection that has the reject option.
		\item The proposed method also enjoys some level of interpretability and because of the structure of the fuzzy reasoning, it is not likely to make a poor generalization.
		\item The proposed method is quite robust to the initial condition.
		\item The proposed model provides a general framework for utilizing fuzzy systems for unsupervised dimensionality reduction. We can use different objective functions to preserve different characteristics of the original data in the projected space. For example, in \cite{das2019unsupervised}, we have used the  Sammons' Stress \cite{sammon1969nonlinear} as the objective function.
		\item We also proposed a modified form of k-means clustering in the proposed method that ensures that the cluster centers are always on the manifold and thereby providing better initial rules.
		\item Finally, the proposed method performs better or comparable to several non-fuzzy methods.
	\end{enumerate}
	The rest of the paper is organized as follows. In section~\ref{sec:literature}, we give a short literature review of dimensionality reduction methods for data visualization. Section~\ref{sec:proposed} elaborates the proposed method. Section~\ref{sec:experiment} discusses the experiments and results. We finally conclude in section~\ref{sec:conclusion}.
	
	\section{Literature Review \label{sec:literature}}
	
	Let $\mathbf{X}=\{\mathbf{x_i}=(x_{i1},x_{i2},\cdots,x_{id_{h}}) \in \mathcal{R}^{d_{h}}: i \in \{1,2,\cdots,n\}\}$ be the input data set, where, $n$ is the number of instances and $d_{h}$ is the dimension of the data. To visualize the data, it will be mapped to a $d_{l}$-dimensional space, where $d_{l}<d_{h}$. Let the lower dimensional data be represented by $\mathbf{Y}=\{\mathbf{y_i} = (y_{i1},y_{i2},\cdots,y_{id_{l}})\in \mathcal{R}^{d_{l}}: i \in \{1,2,\cdots ,n\}\}$. Dimensionality reduction methods estimate a map from $R^{d_{h}}$ to $R^{d_{l}}$ keeping some relevant information of $\mathbf{X} $ as intact as possible in $\mathbf{Y}$. Generally, for visualization purpose, $d_{l}$ is chosen as two or three. Other than visualization, dimensionality reduction methods are also applied in data de-noising, compressing, extracting suitable features for classification, clustering and so on \cite{cunningham2015linear,van2009dimensionality,meng2008nonlinear}. We discuss here only the methods which attempt to preserve the structure of the data and aid in visualization. The methods aiming at other applications are not discussed here.
	
	Dimensionality reduction methods could be divided into two groups: linear and non-linear.  Principal component analysis (PCA), classical multi-dimensional scaling (MDS) are among extensively used linear methods \cite{cunningham2015linear}. They focus on preserving the large distances of the original space in the obtained lower dimensional space. However, for data sets where points lie near or on a non-linear manifold, preserving small distances or local structure is more important. In those scenarios, linear mapping based methods fail to represent the desirable characteristics of the data. Non-linear dimensionality reduction methods like LLE\cite{roweis2000nonlinear} and ISOMAP\cite{tenenbaum2000global} attempt to preserve the local structure of the data and consequently unfold the non-linear structure. In case of classification where the structure preservation may not be important, non-linear methods are also advantageous. Consider a data set consisting of points from a sphere and a spherical shell surrounding the sphere, where the sphere and shell represent two classes. These two classes cannot be separated by a hyperplane in the original space. Their projections obtained by linear dimensionality reduction methods cannot also be separated linearly into two classes even if the classes are well separated. But there can be non-linear methods which can project such data into a space where the two classes are linearly separable. 
	Linear methods are extensively used despite their limitations in handling non-linear structure present in the data. The most widespread linear method is PCA \cite{duda2001pattern}. PCA maximizes the data variance or equivalently minimizes the error in the reconstruction of the original data from the generated lower dimensional data. A special case of PCA is classical (linear) MDS \cite{borg2003modern}. Classical MDS,  precisely the `maximal variance' PCA aims to maximize the scatter of the lower dimensional projection to yield the most informative projection.
	
	Maximum auto-correlation factors (MAF) \cite{switzer1984Min, larsen2002decomposition} preserves temporally interesting structure present in the input to the projected output data. This method also considers a linear projection. MAF assumes that there is an underlying $d_{l}$ dimensional temporal signal which is smooth and the remaining $d_{h}-d_{l}$ dimensions are noise with little temporal correlation. MAF estimates the orthogonal projection $M^{T}$ which maximizes the correlation between adjacent output points $\mathbf{y}_{t},\mathbf{y}_{t+\delta}  $ through an objective function \cite{larsen2002decomposition}.
	
	We note that all the above mentioned methods consider a global objective. This leads to distortions when subjected to data with non-linear structure and in the presence of outliers \cite{cunningham2015linear}. To avoid this, a popular alternative is to consider the local neighborhood structure of the data.
	Two such popular methods are locality preserving projections\cite{he2004locality} based on Laplacian Eigenmaps\cite{belkin2003laplacian} and neighborhood preserving embedding \cite{he2005neighborhood} based  on LLE. 
	A detailed review of various linear dimensionality reduction methods is available in\cite{cunningham2015linear}. We discuss non-linear methods next.
	
	 A group of non-linear methods tries to preserve the inter-point distances in high dimensional space as the inter-point distances in the estimated low dimensional space as much as possible. Examples of such methods are Sammon's projection\cite{sammon1969nonlinear}, curvilinear component analysis (CCA)\cite{demartines1997curvilinear}, ISOMAP\cite{tenenbaum2000global}, curvilinear distance analysis (CDA)\cite{lee2000robust}.
	
	Sammon's projection extends the idea of classical multidimensional scaling, i.e., preservation of pairwise distances. The following cost function is minimized with respect to $\mathbf{y_i} $s,  $i \in \{1,2,\cdots ,n\}$
	\begin{equation} \label{eq: sammon}
     \textstyle E=( \textstyle  \nicefrac{1}{\sum_{i}\sum_{j} d_{ij}} ) \sum_{i} \sum_{j \neq i} \textstyle  \nicefrac{\left( d_{ij}-\lVert\mathbf{y}_i-\mathbf{y}_j\rVert\right)^{2}}{d_{ij}} .
	\end{equation}
	where $d_{ij}$ represents the pairwise Euclidean distance between the high dimensional data points $\mathbf{x}_{i}$ and $\mathbf{x}_{j}$. 
	
	Sammon's projection, however, tends to fail on data sets such as a Swiss Roll\cite{roweis2000nonlinear}. In Swiss Roll, the Euclidean distance between two points can be small but their distance over the manifold on which the data points reside can be large. To model such data sets we should consider the pairwise distance computed using the distances over the manifold, i.e., the geodesic distance. In differential geometry, a geodesic is a curve defining the shortest path between a pair of points on a surface or in a Riemannian manifold. The length of the geodesic defines the geodesic distance between the two points. Thus it is the shortest distance between two points while moving along the surface. The geodesic generalizes the notion of ``straight line" on a Riemannian manifold. Mathematically, the distance between two points p and q over a Riemannian manifold is defined to be the infimum of the lengths $L(\gamma)$ over all the piece-wise smooth curve segments $\gamma$ from $p$ to $q$ \cite{lee2001introduction}. 
	
	ISOMAP is one of the methods which considers the geodesic distance as the distance measure for the input space. It applies the classical MDS over the input geodesic distance matrix to compute the lower dimensional embedding $\mathbf{Y}$. The vectors $\mathbf{y}_{i}$s, $i \in \{1,2,\cdots ,n\}$, are chosen to minimize the cost function $\textstyle  E= \textstyle  \lVert \tau(\mathbf{D}_{G})-\tau(\mathbf{D}_{Y})\rVert_{L^2} $. Here, $\mathbf{D}_{G}$ denotes the matrix of geodesic distance over the input space, $\mathbf{D}_{Y}$ denotes the matrix of Euclidean distance over the output space. $\lVert \mathbf{A} \rVert_{L^2} $ is the $L^2$ matrix norm, $\lVert \mathbf{A} \rVert_{L^2}=\sqrt{\sum_{ij}A_{ij}^{2}}$. The $\tau$ operator converts distances to inner-products \cite{tenenbaum2000global}.
	
	Almost at the same time of ISOMAP, another manifold learning algorithm was introduced named LLE. Although this method makes use of neighborhood, the underlying concept is quite different. For data points residing on or approximately on a manifold, we can assume a small neighborhood of a point to be linear, i.e., points in a small neighborhood are on a linear patch. The entire manifold is made of numerous such small linear patches. This local linear model for each data point is estimated in the LLE. The low dimensional data  representation is computed in such a way that the local geometry of the original space is preserved. This is done in two steps. First, ${w}_{ij}$s are estimated by minimizing $\Phi_{i}\left( \mathbf{x}\right) = \textstyle  \lVert\mathbf{x}_{i}-\sum_{\mathbf{x}_{j} \in \mathcal{N}(\mathbf{x}_{i})} {w}_{ij}\mathbf{x}_{j}\rVert^{2}$ for $i= 1 \text{ to } n$, where $\mathcal{N}(\mathbf{x}_{i})$ represents the set of neighbors of the point $\mathbf{x}_{i}$. Second, after the ${w}_{ij}$s are estimated for each $\mathbf{x}_{i}$s, the corresponding lower dimensional representation $\mathbf{y}_{i}$s are estimated minimizing the objective function $ \psi_{i}\left( \mathbf{y}\right)= \textstyle  \lVert\mathbf{y}_{i}-\sum_{\mathbf{x}_{j} \in \mathcal{N}(\mathbf{x}_{i})} {w}_{ij}\mathbf{y}_{j}\rVert^{2}$ keeping ${w}_{ij}$s constant, for $i= 1 \text{  to } n$.
	
	The Laplacian Eigenmaps \cite{belkin2003laplacian} is another approach similar to the LLE. Here the local property is characterized by the distances of a point to its neighbors. The  weight $w_{ij}$ associated with a data point $\mathbf{x}_{i}$ and its neighbor $\mathbf{x}_j\in \mathcal{N}(\mathbf{x_{i}})$
	is computed using (\ref{eq: Laplacian Eigenmap1}).
	\begin{equation} \label{eq: Laplacian Eigenmap1}
	\textstyle  w_{ij}=\Bigg\{\begin{array}{ll}
	\textstyle  e^{-\nicefrac{\lVert\mathbf{x}_{i}-\mathbf{x}_{j}\rVert^{2}}{2 \sigma^{2}}} \text{ if }\mathbf{x}_{j}  \in   \mathcal{N}(\mathbf{x}_{i})\\0 \hspace{40pt}\text{ otherwise } \end{array} ,
	\end{equation}
	where $\sigma > 0$ is the spread of the Gaussian function. The corresponding lower dimensional representations $\mathbf{y}_{i}$s are computed by minimizing the cost function $\phi(\mathbf{Y}) =\sum_{ij} w_{ij}\lVert\mathbf{y}_{i}-\mathbf{y}_{j}\rVert^{2}$.
	
	Another method named $t$-SNE \cite{maaten2008visualizing} is able to create a single map that reveals structure at many different scales. The $t$-SNE converts the Euclidean inter-point distances in the high dimensional input space to symmetrized conditional probabilities $p_{ij} = \dfrac{p_{j|i}+p_{i|j}}{2n}$, where $n$ is number of data points and $p_{j|i}$ signifies the conditional probability that $\mathbf{x}_{i}$ would pick $\mathbf{x}_{j}$ as its neighbor. The conditional probability $p_{j|i}$ follows a Gaussian distribution. 
	Similarly, the joint probability of the lower dimensional outputs, $\mathbf{y}_i$ and $\mathbf{y}_j$ is represented by $q_{ij}$ and it is considered to follow the student's $t$ distribution with one degree of freedom. 
	The lower dimensional representations $\mathbf{y}_{i}$s are obtained by minimizing the Kullback-Leibler divergence between the joint probability distributions of the input and its lower dimensional representation.
%
	There are several other non-linear methods. To name a few: Kernel PCA \cite{scholkopf1997kernel}, Maximum variance unfolding (MVU) \cite{weinberger2006introduction}, Diffusion Maps \cite{coifman2006diffusion}, Hessian LLE \cite{donoho2003hessian}, and Local tangent space analysis \cite{zhang2004principal}. A common drawback of all of these methods is that they are non-parametric in nature and consequently, they cannot predict projections for test data points.

	Now we shall discuss some parametric models. 
	The study in \cite{pal1998two} proposed a learning system based on multi-layer perceptron (MLP) for structure preserving dimensionality reduction. In this work, Sammon's projections, of the sampled input or of the input prototypes generated by a self organizing map were used as target outputs in a supervised gradient descent based learning scheme. The investigations \cite{jain1992artificial,mao1995artificial} explored unsupervised learning of MLP for dimensionality reduction. In \cite{jain1992artificial} Sammon's stress function was used as the learning objective whereas in \cite{mao1995artificial} multiple networks were proposed with different objective functions. In \cite{hinton2006reducing, demers1993non} multi-layer autoencoders with an odd number of hidden layers were employed for dimensionality reduction. 
	To reduce the original data to $d_{l}$ dimension, the middle hidden layer is chosen to have $d_{l}$ nodes. After training the autoencoder, when an input $\mathbf{x}_i$ is applied,  the $d_{l}$-dimensional representation obtained at the middle hidden layer is taken as the corresponding dimensionality reduced representation $\mathbf{y}_{i}$.  However, these models do not take into account the interrelationship between data points and the underlying manifold structure.  The work in \cite{wang2014generalized} proposes a scheme called ``generalized autoencoder" (GAE) which incorporates concepts of manifold learning. The GAE extends the traditional autoencoder in two aspects: $(i)$ Each instance $\mathbf{x}_i$ reconstructs a set of instances $\{\mathbf{x}_j\}$ rather than reconstructing only itself. $(ii)$ The reconstruction error of an instance $ \mathbf{x}_i$ is weighted by a relational function of $\mathbf{x}_i$ and $ \mathbf{x}_j$  defined on the learned manifold. GAE learns the underlying manifold by minimizing the weighted distance between the original and reconstructed ones.

	Works exploring fuzzy rule-based systems to reduce the data dimensionality for visualization purposes is scarce. In \cite{pal2002fuzzy} a \emph{supervised} fuzzy rule-based model is proposed for structure preserving dimensionality reduction. Here given the high dimensional inputs and its lower dimensional projections (generated by any method such as Sammon's projection), a fuzzy rule-based system is used to learn the projecting map. In this context both Mamdani-Assilian (MA) \cite{mamdani1975experiment} and TS \cite{takagi1985fuzzy} models have been used.
	
	Apart from these computational intelligence based works, there are studies to find out explicit map between higher and lower dimensional data. In \cite{qiao2012explicit}, the authors proposed an algorithm named neighborhood preserving polynomial embedding (NPPE). The method considers the objective function of LLE as its learning objective where the lower dimensional output vectors are represented as a polynomial function of higher dimensional input vectors. The unknown coefficients of the polynomial function are estimated from the modified LLE objective function by solving a generalized eigenvalue problem.
	
	\section{Proposed Method \label{sec:proposed}}
	In this work, we aim to produce a lower dimensional representation of the original data in such a way that the local neighborhood structure present in original space is preserved. We also want to transform the data globally in such a way that if the data lie on any non-linear high dimensional manifold, it would be flattened in the lower dimensional space preserving the local neighborhood structure of the input space.

	For this, we consider a fuzzy rule-based system as the most appropriate tool. Although an MA model could be used, considering greater flexibility we use the TS model. Since this is an \emph{unsupervised} method the main challenge is to define a suitable objective function. The second challenge is how to generate the initial rule base. Typically, we use a clustering algorithm to generate the initial rules. But when the data points are on a manifold whose intrinsic dimension is lower than the original dimension of the data, the $c$-means or the fuzzy $c$-means may find cluster centers that are significantly away from the manifold.
	
	As mentioned earlier the input data set is denoted by $\mathbf{X}=\{\mathbf{x}_{1}, \mathbf{x}_{2},\cdots,\mathbf{x}_{n}\}, \mathbf{x}_{i} \in \mathbb{R}^{d_{h}} $ . The projected output data set is denoted by  $\mathbf{Y}=\{\mathbf{y}_{1}, \mathbf{y}_{2},\cdots,\mathbf{y}_{n}\}, \mathbf{y}_{i} \in \mathbb{R}^{d_{l}}$ where, $\mathbf{y}_{i}$ is the projection of $\mathbf{x}_{i}, i=\{1,2,\cdots,n\}$. For $m$th output variable, let there be $n_{c}$ rules of the form
	\begin{flalign}\label{eq:rule}
	\textstyle  &R_{km}:\text{ If } x_{1} \text{ is } F_{k1} \text{ AND } x_{2} \text{ is } F_{k2}\text{ AND } \cdots \text{ AND } \nonumber\\
	\textstyle  &x_{d_{h}} \text{ is } F_{kd_{h}}\text{ then } y_{m}^{k}= \textstyle a_{km0}+ \textstyle \sum_{q=1}^{d_{h}} a_{kmq}x_{q}
	\end{flalign}
	where $k=1,2,\cdots,n_{c};\hspace{2pt} m=1,2,\cdots,d_{l}$; $F_{kq}$ is the $k$th fuzzy set (linguistic value) defined on the $q$th feature and $ a_{kmq}$'s are consequent parameters. Let us define the matrix $A$ $=$ $(A_{1}, A_{2},\cdots A_{n_{c}})^{T}$ where, $A_{k}$ $=$ $\left(a_{k10}, a_{k11}, \ldots ,\right.$ $\left.a_{k1d_{h}}, a_{k20},a_{k21},\ldots,\right.$ $\left.a_{k2d_{h}},\ldots,a_{kd_{l}0},a_{kd_{l}1},\ldots, a_{kd_{l}d_{h}} \right)$. Let the firing strength of the rule $R_{km}$ for the point $\mathbf{x}_{i}$ be $\alpha_{k,i}; k=1,2,\cdots,n_{c}$. Note that, for an antecedent, `$x_{1} \text{ is } F_{k1} \text{ AND } x_{2} \text{ is } F_{k2}\text{ AND } \cdots \text{ AND } x_{d_{h}} \text{ is } F_{kd_{h}}$' there are $d_{l}$ consequents, `$y_{m}^{k}= \textstyle a_{km0}+ \textstyle \sum_{q=1}^{d_{h}} a_{kmq}x_{q} $'; $m=1,2,\cdots,d_{l}$, resulting in $d_{l}$ rules. But the firing strength for each of these $d_{l}$ rules remain the same ($\alpha_{k,i}$).
	\begin{subequations}
		\label{eq:output}
	\begin{flalign}
	&y_{im}=\textstyle \sum_{k=1}^{n_{c}}\left(\nicefrac{\textstyle \alpha_{k,i} }{\sum_{p=1}^{n_{c}}\alpha_{p,i}}\right)y_{im}^{k}\label{eq:output_1}\\
	&=\textstyle \sum_{k=1}^{n_{c}}\left(\nicefrac{\textstyle \alpha_{k,i} }{\sum_{p=1}^{n_{c}}\alpha_{p,i}}\right)\left(a_{km0}+\sum_{q=1}^{d_{h}} a_{kmq}x_{iq} \right)\label{eq:output_2}
	\end{flalign}	
	\end{subequations}
	where, $i=1,2,\cdots,n; \hspace{4pt} m=1,2,\cdots,d_{l}$. Typically, for designing a fuzzy rule-based system, the target outputs for the training data are known and the rule base parameters are estimated by minimizing the square error defined by the target outputs and the estimated outputs.
    But for us the target output is \emph{not} available. So we need to define a suitable objective function that can help to learn the manifold of the input data $\mathbf{X}$. One promising approach would be to preserve the geodesic neighborhood relationship, i.e., the geodesic distance structure of the manifold on to the projected lower dimension. Let $gd_{ij}^{\mathbf{X}}$ be the geodesic distance between $\mathbf{x}_{i}$ and $\mathbf{x}_{j}$, $\mathbf{x}_{i}, \mathbf{x}_{j} \in \mathbb{R}^{d_{h}}$ and $ed_{ij}^{\mathbf{Y}}$ be the euclidean distance between $\mathbf{y}_{i}$ and $\mathbf{y}_{j}$, $\mathbf{y}_{i}, \mathbf{y}_{j} \in \mathbb{R}^{d_{l}}$. A good objective function to estimate $\mathbf{y}_{i}$s is (\ref{eq: error})
	\begin{equation} \label{eq: error}
	\textstyle  E=\sum_{i=1}^{n-1} \sum_{j=i+1}^{n} \nicefrac{\left( gd_{ij}^{\mathbf{X}}-ed_{ij}^{\mathbf{Y}}\right)^{2}}{gd_{ij}^{\mathbf{X}}}.
	\end{equation}
	The objective (\ref{eq: error}) is introduced by Yang in \cite{yang2004sammon}. Note that (\ref{eq: error}) is similar to Sammon's stress function which uses Euclidean distance in both high and low dimensional spaces. In place of (\ref{eq: error}) we can use other functions also. Next, we address the issue of identification of the rule base.
	\subsection{Initial Rule Extraction}
	When both input-output data are provided there are many ways of generating an initial rule base and its refinement \cite{sugeno1993fuzzy,wang1992generating,cordon1999two,mitra2000neuro,pal2002fuzzy,pal2008simultaneous,chen2012integrated}. Some of the popular methods use evolutionary algorithm \cite{cordon1999two} or clustering \cite{pal2002fuzzy,pal2008simultaneous,chen2012integrated} of input-output data. Here we do not have the output data. So we cluster the input data into a predefined number of clusters and translate each cluster into a rule (in particular, to a rule antecedent). A natural choice for the clustering algorithm appears to be the $c$-means (often called $k$-means) or the fuzzy $c$-means (FCM). The proposed method being a fuzzy rule-based scheme, the obvious choice seems to be the FCM clustering algorithm. But we did not do so for two reasons: First, for the FCM, the cluster centers may not fall on the data manifold. Since our idea is to use a cluster centroid to model a set of points in the neighborhood of the cluster centroid, it is better to have the centroid on the data manifold. Second, the geodesic distance is not defined in terms of an inner-product induced norm and hence the FCM convergence theory does not hold for the geodesic distance based FCM.   So we use a slightly modified $c$-means called Geodesic $c$-Means (GCM)  algorithm. Here we cluster the input, aiming to obtain some representative points on the input manifold. To extract information regarding the input manifold structure, we use geodesic distance as the dissimilarity measure for clustering. Like the conventional $c$-means, data points are assigned to a cluster using the minimum distance (here minimum geodesic distance) criterion and the cluster centroids are computed as the mean vector of the points in a cluster. Since such centroids may not lie on the manifold, we use an extra step. For each computed centroid, we find the input data point closest to the centroid and use that data point as a cluster centroid. We note here that the use of geodesic distance in the $c$-means clustering algorithm has been investigated in other studies \cite{asgharbeygi2008geodesic,kim2007soft,feil2007geodesic} but their approaches are different from ours. In \cite{asgharbeygi2008geodesic} the authors introduced a class of geodesic distance which took into account local density information and employed that geodesic distance in the $c$-means clustering algorithm. The study in \cite{kim2007soft} incorporated geodesic distance in a soft kernel $c$-means algorithm. The authors in \cite{feil2007geodesic} integrated the geodesic distance measure into the fuzzy $c$-means clustering algorithm. The algorithmic description of  our clustering algorithm (GCM) is given in Algorithm \ref{alg:GeoCMeans}.
	To estimate the geodesic distance we have followed an approach similar to that in ISOMAP\cite{tenenbaum2000global}. First, we construct a neighborhood graph of the input data points based on the Euclidean distance. For approximating geodesic distance, every edge is assigned with a weight/cost which is basically the Euclidean distance between the pair of points on which the edge is incident. For any point, the geodesic distances to its neighboring points are approximated by the Euclidean distance. The geodesic distance between two points which are not neighbors, i.e., not connected by an edge is approximated by evaluating the shortest path on the neighborhood graph. For the shortest path distance estimation, we use the Floyd-Warshall algorithm \cite{floyd1962algorithm}.
	
	The $k$th cluster centroid, $\mathbf{v}_{k}=(v_{k1},v_{k2},\cdots,v_{kd_{h}})$, obtained by GCM is translated into the antecedent of the $k$th rule, $R_{km}$ as follows.
	\begin{flalign}
	\textstyle & \text{If } x_{1} \text{ is ``close to } v_{k1}" \text{ AND } x_{2} \text{ is ``close to } v_{k2}" \nonumber\\
	\textstyle &\text{AND } \cdots x_{d_{h}} \text{ is ``close to } v_{kd_{h}}".
	\end{flalign}
	The fuzzy set $F_{kq}$ in (\ref{eq:rule}) is defined linguistically as `` close to $v_{kq}$"; $q=1,2,\cdots,d_{h}$. Thus the $k$th rule given in (\ref{eq:rule}) becomes
	\begin{flalign}\label{eq:rule_centroid}
	\textstyle &R_{km}:\text{If } x_{1} \text{ is ``close to } v_{k1}" \text{ AND } x_{2} \text{ is ``close to } v_{k2}" \text{ AND } \nonumber \\
	\textstyle &\ldots\text{AND } x_{d_{h}} \text{ is ``close to } v_{kd_{h}}" \text{ then }\textstyle   y_{m}^{k}= \textstyle \sum_{q=0}^{d_{h}} a_{kmq}x_{q}
	\end{flalign}
	where, $x_0=1$, $k=1,2,\cdots,n_{c}; \hspace{2pt} m=1,2,\cdots,d_{l}.$ We model `$x_{q}$ is ``close to $v_{kq}$"' using a Gaussian membership function with the center at $v_{kq}$. Although any $\pi$-type membership function can be used, to exploit the differentiability property, we use Gaussian membership functions. This also makes the antecedent of the $k$th fuzzy rule to model a hyper-ellipsoidal volume centering $\mathbf{v}_{k}$ in the input space. For the $i$th point, the membership to the set ``close to $v_{kq}$" is computed as
	\begin{equation} \label{eq:membership}
	\mu_{kq,i}=\exp\left\lbrace -\nicefrac{(x_{iq}-v_{kq})^2}{2\sigma_{kq}^2}\right\rbrace
	\end{equation}
	Here, $\sigma_{kq}$ is the spread of the Gaussian centered at $v_{kq}$ which is the membership function of the fuzzy set $F_{kq}$. We use product as the $T$-norm to aggregate $\mu_{kq,i}$'s to calculate $\alpha_{k,i}$.
	\begin{equation}\label{eq:firing_strength}
	\textstyle \alpha_{k,i}= \prod_{q=1}^{d_{h}}\mu_{kq,i}.
	\end{equation}
	We denote the matrix of cluster centroids as $V=\left( \mathbf{v}_{1},\mathbf{v}_{2},\cdots,\mathbf{v}_{n_{c}}\right)^{T}=\left[v_{kq} \right]_{n_{c}\times d_{h}} $. Similarly, the matrix of spreads $\Sigma=\left( \mathbf{\sigma}_{1},
	\mathbf{\sigma}_2,\cdots,\mathbf{\sigma_{n_{c}}}\right)^{T}=\left[\sigma_{kq} \right]_{n_{c}\times d_{h}} $. An initial estimate of $\sigma_{kq}$ can be taken as the standard deviation of $q$th feature in the $k$th cluster or can be initialized using some other method. The set of consequent parameters $A$ is initialized randomly. Note that these are initial choices which will be refined during the training. 
	
    \subsection{The Objective Function and its Optimization}
    To learn the lower dimensional representations $\mathbf{Y}$ of the given input $\mathbf{X}$, the error function defined in (\ref{eq: error}) is 
    minimized 
    with respect to the rule base parameters $V,\Sigma,$ and $A$. Using (\ref{eq:membership}) and (\ref{eq:firing_strength}) in (\ref{eq:output_2}) we obtain the following relation.
    \begin{equation}\label{eq:output_as_a_function}
     y_{im}= \textstyle \sum_{k=1}^{n_{c}}\frac{\textstyle \left( \prod_{q=1}^{d_{h}}\exp\left\lbrace -\frac{(x_{iq}-v_{kq})^2}{2\sigma_{kq}^2}\right\rbrace\right) \left(\sum_{q=0}^{d_{h}} a_{kmq}x_{iq} \right) }{\sum_{p=1}^{n_{c}}\prod_{q=1}^{d_{h}}\exp\left\lbrace -\frac{(x_{il}-v_{pq})^2}{2\sigma_{pq}^2}\right\rbrace}
    \end{equation}
    Here, $x_{i0}=1$. Now, $\mathbf{y}_{i}=\left(  y_{i1}, y_{i2}, \cdots, y_{id_{l}}\right) $. Hence $\mathbf{y}_{i}$ is dependent on $v_{kq}, \sigma_{kq},$ and $a_{kmq}; \forall k=1 \text{ to } n_{c}, q=1 \text{ to } d_{h} \text{ and } m=1 \text{ to } d_{l}$ which forms $V,\Sigma \text{ and } A$, respectively. It is evident that we can write the lower dimensional vectors $\mathbf{y}_{i}$s as a function of the rule base parameters $V,\Sigma,$ and $A$.
    \begin{equation}\label{eq:output_as_a_function_1}
    \textstyle\mathbf{y}_{i}=f(V,\Sigma,A;\mathbf{x}_{i})
    \end{equation}
    Replacing $ed_{ij}^{\mathbf{Y}}$ in (\ref{eq: error}) as $\lVert\mathbf{y}_{i}-\mathbf{y}_{j}\rVert$ and then using (\ref{eq:output_as_a_function_1}) we get
    \begin{flalign} \label{eq:error_with_y}
      & E=\textstyle \sum_{i=1}^{n-1} \sum_{j=i+1}^{n} \nicefrac{\textstyle \left( gd_{ij}^{\mathbf{X}}-\lVert\mathbf{y}_{i}-\mathbf{y}_{j}\rVert\right)^{2}}{gd_{ij}^{\mathbf{X}}}\nonumber\\
     &=\textstyle\sum_{i=1}^{n-1} \sum_{j=i+1}^{n} \nicefrac{\left( gd_{ij}^{\mathbf{X}}-\lVert f(V,\Sigma,A;\mathbf{x}_{i})-f(V,\Sigma,A;\mathbf{x}_{j})\rVert\right)^{2}}{gd_{ij}^{\mathbf{X}}} .
    \end{flalign}
    The optimum values of $V,\Sigma,$ and $A$ are obtained as
    \begin{equation}\label{eq:optimization}
    \textstyle (V_{opt},\Sigma_{opt},A_{opt})=\argmin_{V,\Sigma,A} \left( E\right)  .
    \end{equation}
    To minimize the error function $E$ in (\ref{eq:optimization}), a momentum based gradient descent approach is used here. The overall process is summarized in Algorithm~\ref{alg:GeoNLMFRS}.
    \begin{algorithm*}[tb] 
    	\DontPrintSemicolon
    	\SetAlgoLined
    	\caption{Unsupervised Fuzzy Rule-based Dimensionality Reduction}\label{alg:GeoNLMFRS}
    	Determine Geodesic distance $gd^{\mathbf{X}}$ between every pair of input data points in $\mathbf{X}$. \;
    	Cluster the input data $\mathbf{X}$ into $n_c$ clusters using the Geodesic  distance based $c$-means clustering algorithm (Algorithm \ref{alg:GeoCMeans}). Let the obtained cluster centroids be $\mathbf{v}_{1},\mathbf{v}_{2},\cdots,\mathbf{v}_{n_{c}};$ $\mathbf{V}=\left( \mathbf{v}_{1},\mathbf{v}_{2},\cdots,\mathbf{v}_{n_{c}}\right)^{T}$. \;
    	Translate each cluster $\left( \mathbf{v}_{k}\right)$ into a rule as explained in (\ref{eq:rule_centroid}) and (\ref{eq:membership}) . Initialize the centers of the Gaussian membership functions (described in (\ref{eq:membership})) using $\mathbf{V}$ and spreads by $\mathbf{\Sigma}$. Here we use predefined initial value for $\sigma_{kq} \forall k,q$. The consequent parameters $\mathbf{A}$ are initialized randomly. \;
    	Initialize iteration count $t=1$. $\Delta\mathbf{V}^{t},\Delta\mathbf{\Sigma}^{t},\text{ and }\Delta\mathbf{A}^{t}$ are updates at time step $t$ corresponding to centers, spreads and consequent parameters. $\Delta\mathbf{V}^{0},\Delta\mathbf{\Sigma}^{0},\text{ and } \Delta\mathbf{A}^{0}$ are set to zeros.\;
    	
    	\While {$ t < Maximum Iteration $} {
    		Differentiate objective function $E$ in (\ref{eq: error}) with respect to $\mathbf{V}, \mathbf{\Sigma}$ and $\mathbf{A}$ to obtain$\nabla_{\mathbf{V}}E,\nabla_{\mathbf{\Sigma}}E$ and $ \nabla_{\mathbf{A}}E$. Calculate $\Delta\mathbf{V}^{t},\Delta\mathbf{\Sigma}^{t},\Delta\mathbf{A}^{t}$ and update $\mathbf{V}^{t}$, $\mathbf{\Sigma}^{t}$ and $\mathbf{A}^{t}$ as follows:
    		\begin{flalign}
    		&\Delta\mathbf{V}^{t}= -\nabla_{\mathbf{V}}E+\alpha \Delta\mathbf{V}^{t-1}; \Delta\mathbf{\Sigma}^{t}= -\nabla_{\mathbf{\Sigma}}E+\alpha \Delta\mathbf{\Sigma}^{t-1};  \Delta\mathbf{A}^{t}= -\nabla_{\mathbf{A}}E+\alpha \Delta\mathbf{A}^{t-1}\\  
    		&\mathbf{V}^{t}=\mathbf{V}^{t-1}+\eta\Delta\mathbf{V}^{t};  \mathbf{\Sigma}^{t}=\mathbf{\Sigma}^{t-1}+\eta\Delta\mathbf{\Sigma}^{t};  \mathbf{A}^{t}=\mathbf{A}^{t-1}+\eta\Delta\mathbf{A}^{t}   
    		\end{flalign}
    		
    		$\alpha >0$ and $\eta >0$ are momentum and learning coefficient, respectively.\;
    	}
    	\Return outputs  $\mathbf{Y}$ and fuzzy rule base parameters $\mathbf{V}$, $\mathbf{\Sigma}$, $\mathbf{A}$. \;    	
    \end{algorithm*}

	\subsection{Scalability Analysis}
	Scalability is a desirable property for every scheme  concerning both data size $(n)$ and dimension $(d_{h})$.  The error function (\ref{eq: error}) of our method considers all the inter-point distances, i.e., a total of $\nicefrac{n(n-1)}{2} $ distances. In each iteration to update the rule parameters $\nicefrac{n(n-1)}{2} $ values need to be computed. So for a total of $t$ iterations, the computational complexity of the proposed method is O($n^{2}t$) for a fixed input dimension. In Table \ref{tab:complexity}, we have compared the computational complexity of the proposed method with four standard dimensionality reduction based data visualization schemes: Sammon's projection, ISOMAP, LLE, and $t$-SNE. We have also used these four schemes in our experiments. Note that, LLE and ISOMAP perform convex optimization whereas Sammon's projection, $t$-SNE as well as our proposed method perform non-convex optimization with iterative steps.
	\begin{table}[!tb]
		\caption{Computational complexity of various methods}\label{tab:complexity}
		\begin{center}
			\begin{tabular}{ccc}
				\hline \hline
				{} & \multicolumn{2}{c}{ Complexity }\\
				{ Method} &  { Computational}&  { Space}\\
				\hline
				Proposed Method& O($n^{2}t$) & O($n^{2}$)\\
				Sammon's Projection\cite{van2009dimensionality} & O($n^{2}t$)  & O($n^{2}$)\\
				ISOMAP\cite{van2009dimensionality} & O($n^{3}$) & O($n^{2}$)\\
				LLE\cite{van2009dimensionality} & O($pn^{2}$) & O($pn^{2}$)\\
				t-SNE\cite{maaten2008visualizing} & O($n^{2}t$) & O($n^{2}$)\\
				\hline \\
				\multicolumn{3}{c}{ $n$= number of instances, $t$= total number of iterations,  }\\
				\multicolumn{3}{c}{ $p$= ratio of nonzero to total number of elements in sparse matrix}
			\end{tabular}
		\end{center}
	\end{table}
	  Table \ref{tab:complexity} demonstrates that our proposed method has the same computational complexity as the other two iterative methods. In terms of space requirement, like most of the other methods, our method also exhibits a complexity of O($n^{2}$). Since both the complexities are quadratic functions of the number of data points, for a very large $n$, it would be difficult to use the proposed method directly. Use of GPU can reduce the effective complexities significantly. Moreover, a proper sampling scheme \cite{pal1998two,pal2002complexity} that produces a subset of the original data which adequately represents the original data distribution may be used. It is also important to understand the dependency of an algorithm with the dimension of the input feature vectors $(d_{h})$. In our method, the number of rule parameters varies linearly with  the input dimension. So both the computational and space complexity scale up linearly with the number of input features. The rule firing strength $\alpha_{k,i}$ is calculated by combining the antecedent clauses (memberships) using product. The number of atomic antecedent clauses in a rule equals the number of input features. For data sets having a large number of features, it may be possible that for many features, the  memberships are sufficiently low for all the rules. Since these membership values are multiplied while calculating the firing strengths, for some inputs no rule may fire with significant strength, creating an undesired situation. So effectively any fuzzy rule-based system involving a large number of features (say, 10,000 or more) encounter numerical difficulties.  
	
	\section{Experimentation and Results \label{sec:experiment}}
	\subsection{Data Set Description}
\begin{figure*}[!tb]
		\centering
		\begin{subfigure}{.33\textwidth}
			\centering
			\includegraphics[width=.8\linewidth]{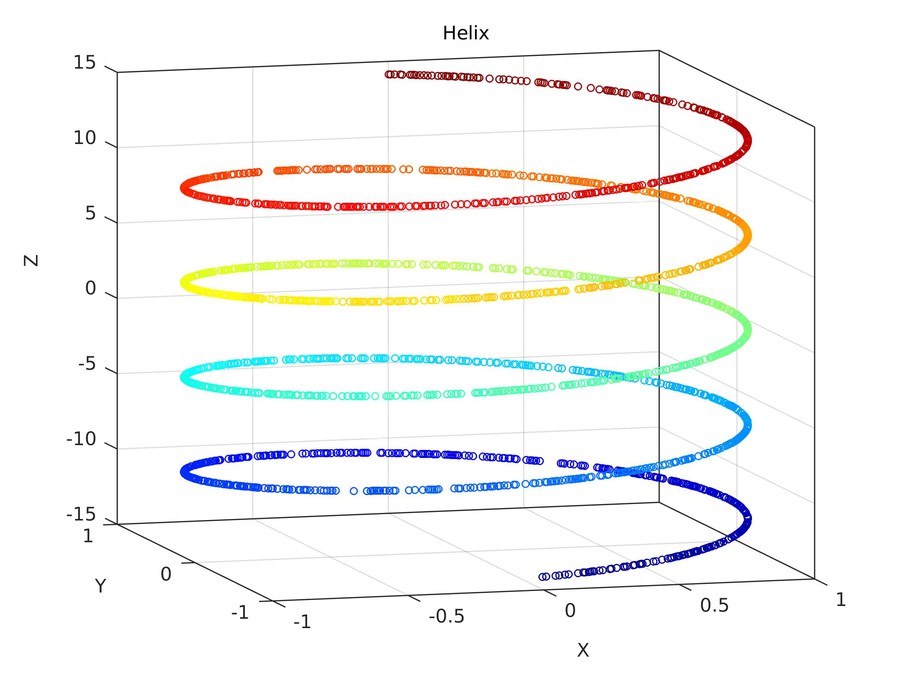}
			\caption{Original Data}
			\label{fig:helix}
		\end{subfigure}%
		\begin{subfigure}{.33\textwidth}
			\centering
			\includegraphics[width=.8\linewidth]{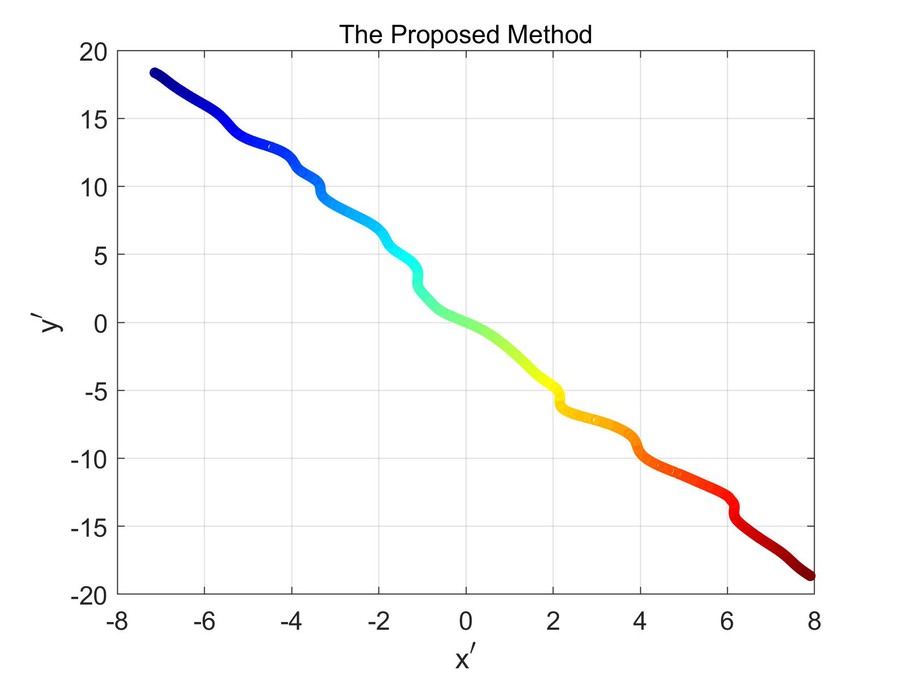}
			\caption{Proposed Method}
			\label{fig:helix proposed}
		\end{subfigure}%
		\begin{subfigure}{.33\textwidth}
			\centering
			\includegraphics[width=.8\linewidth]{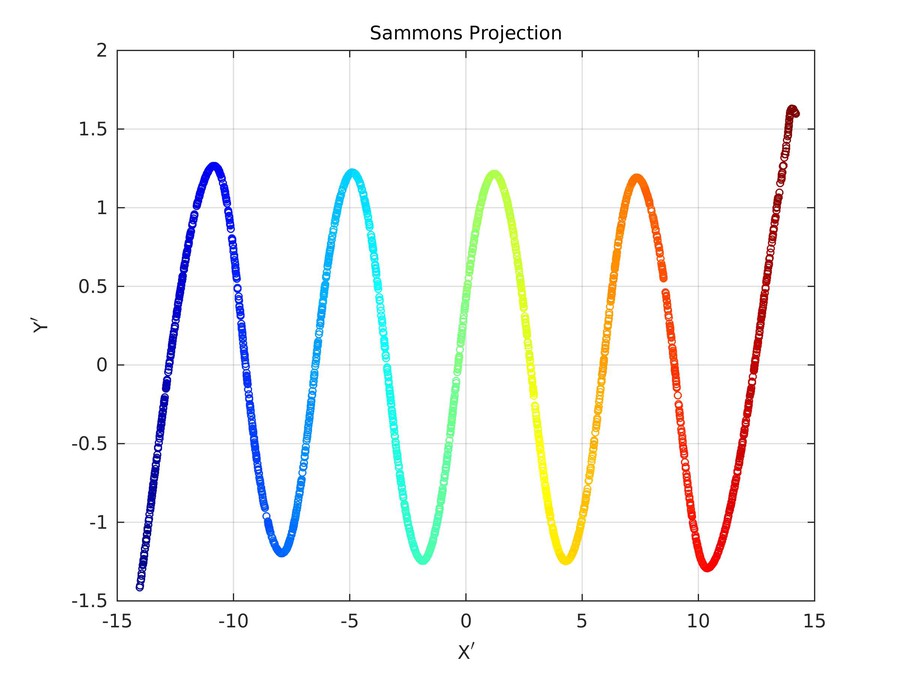}
			\caption{Sammon's Projection}
			\label{fig:helix sammons}
		\end{subfigure}%
		\\
		\begin{subfigure}{.33\textwidth}
			\centering
			\includegraphics[width=.8\linewidth]{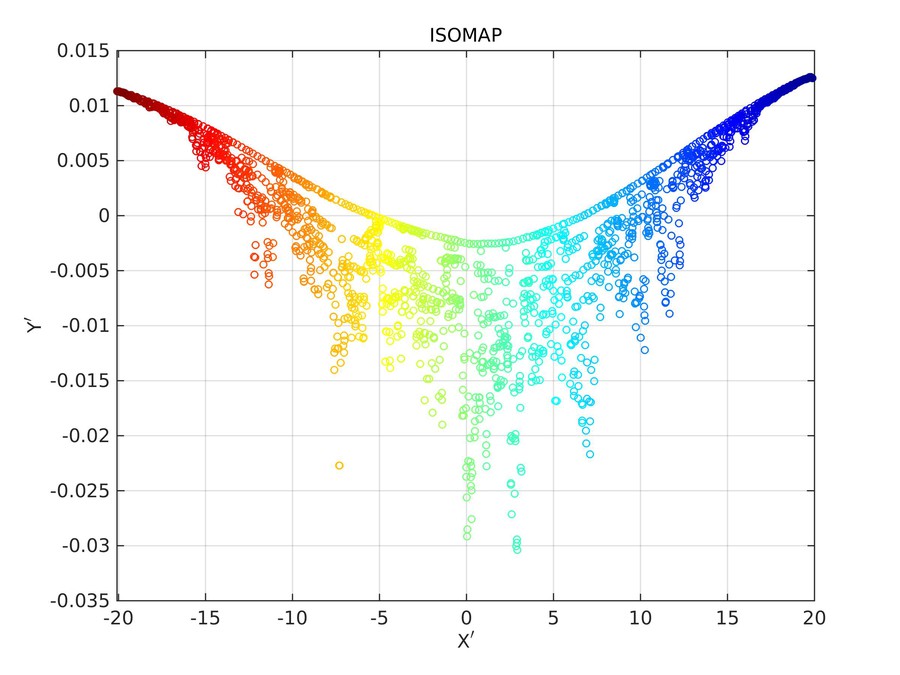}
			\caption{ISOMAP}
			\label{fig:helix ISOMAP}
		\end{subfigure}
		\begin{subfigure}{.33\textwidth}
			\centering
			\includegraphics[width=.8\linewidth]{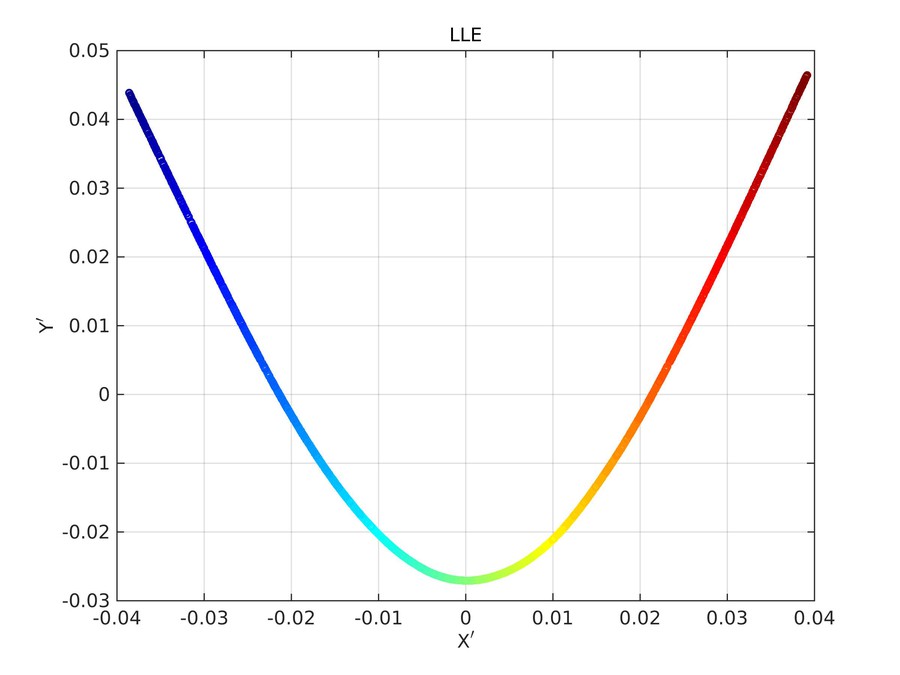}
			\caption{LLE}
			\label{fig:helix LLE}
		\end{subfigure}%
		\begin{subfigure}{.33\textwidth}
			\centering
			\includegraphics[width=.8\linewidth]{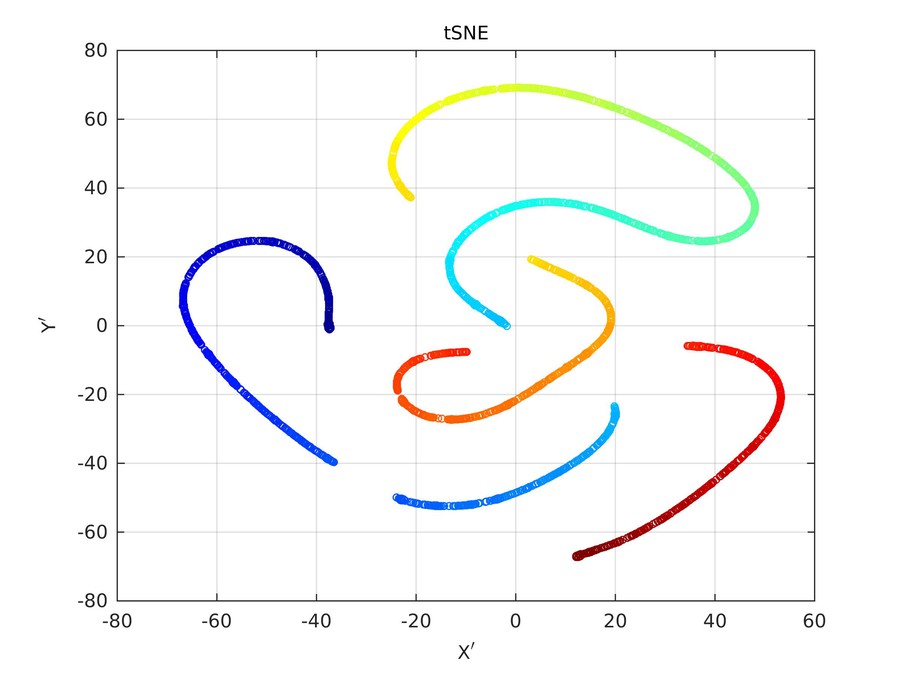}
			\caption{$t$-SNE}
			\label{fig:helix tSNE}
		\end{subfigure}
		\label{fig: helix results}
		\caption{Visualization of the Helix data with (a) original input space, (b) proposed method, (c) Sammon's projection, (d)ISOMAP, (e) LLE, and (f) $t$-SNE.}
	\end{figure*}

	For our experiments, we  consider three synthetic data sets and three real-world data sets. These data sets are used by others as benchmark data sets. The synthetic data are Swiss Roll, S Curve, and Helix as shown in Figs.~\ref{fig:swiss roll}, \ref{fig:s curve} (Figures in the Supplementary materials are numbered as S-$1$, S-$2$, $\cdots$ and so on), \ref{fig:helix}, respectively. They are all three dimensional data but contained completely within a two dimensional space. Each of these synthetic data sets consists of $2000$ points. The first two data sets are generated using `scikit-learn' package  \cite{scikit-learn} (version $0.19.2$) of python. Helix data set is generated by the equations: $z=\nicefrac{t}{\sqrt{2}}; x=\cos z; \text{ and } y=\sin z$.
	The variable $t$ is varied from $-20$ to $20$ with steps of $0.02$.
	\begin{figure*}[!tb]
		\centering
		\begin{subfigure}{.33\textwidth}
			\centering
			\includegraphics[width=.8\linewidth]{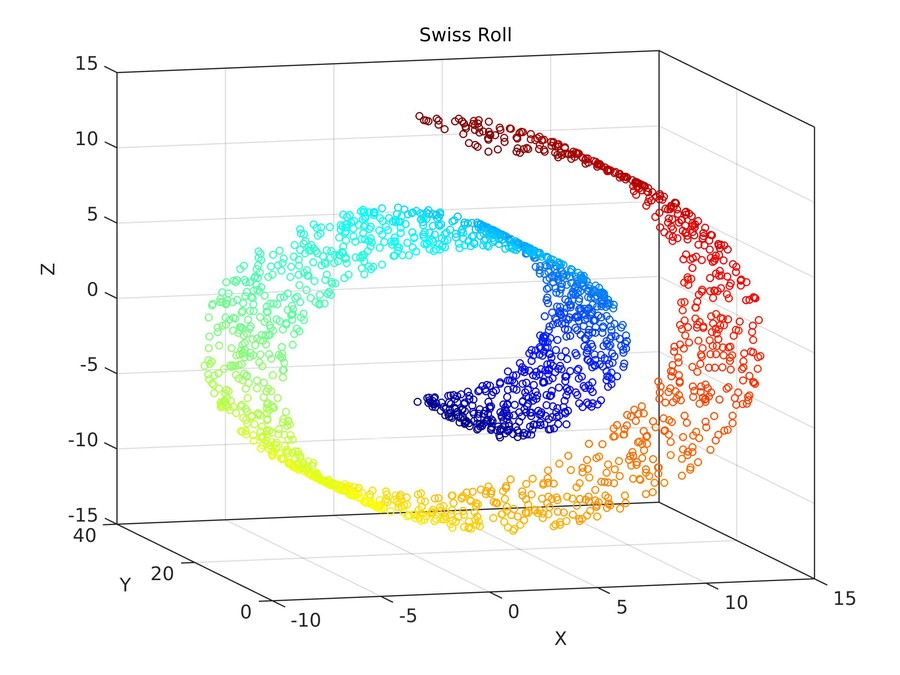}
			\caption{Original Data}
			\label{fig:swiss roll}
		\end{subfigure}%
		\begin{subfigure}{.33\textwidth}
			\centering
			\includegraphics[width=.8\linewidth]{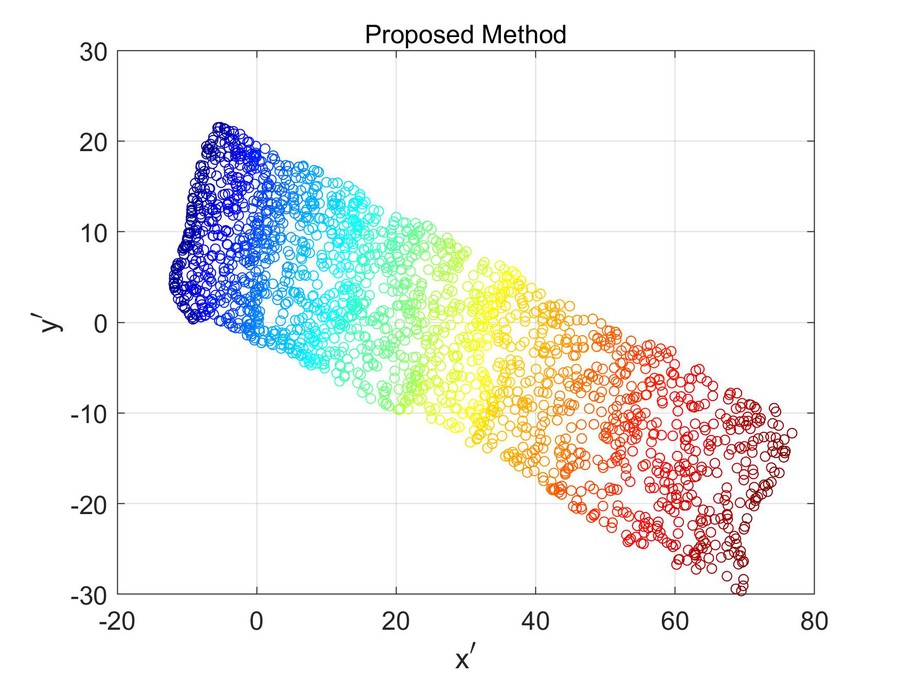}
			\caption{Proposed Method}
			\label{fig:swiss roll proposed}
		\end{subfigure}%
		\begin{subfigure}{.33\textwidth}
			\centering
			\includegraphics[width=.8\linewidth]{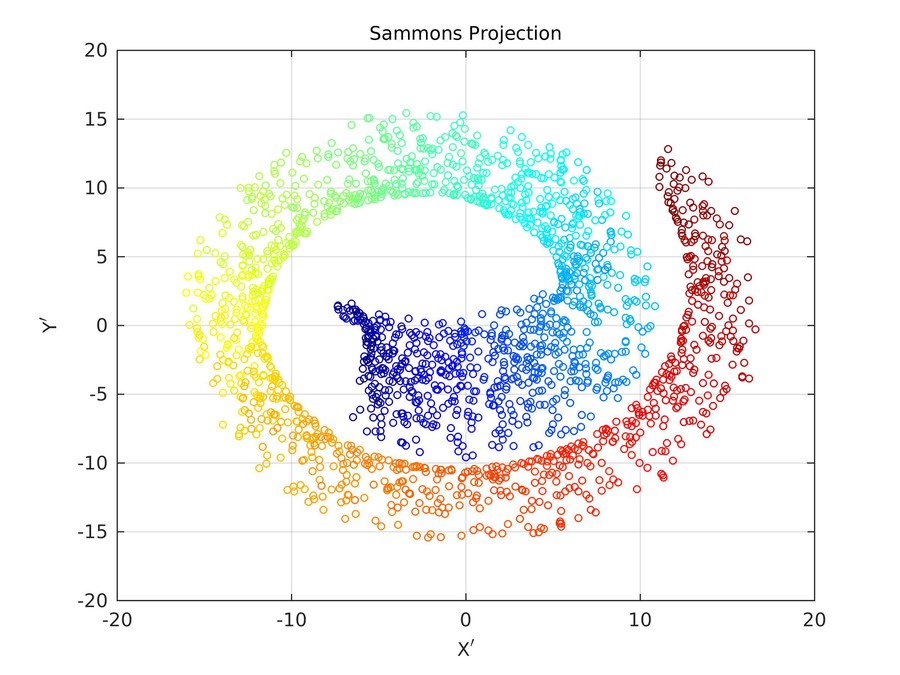}
			\caption{Sammon's Projection}
			\label{fig:swiss roll sammons}
		\end{subfigure}%
		\\
		\begin{subfigure}{.33\textwidth}
			\centering
			\includegraphics[width=.8\linewidth]{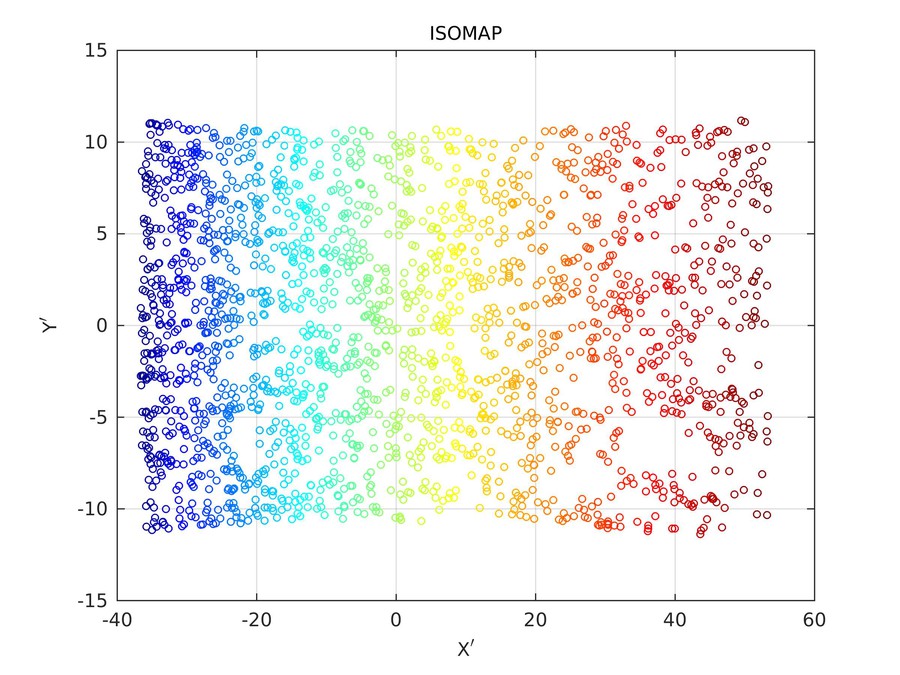}
			\caption{ISOMAP}
			\label{fig:swiss roll ISOMAP}
		\end{subfigure}
		\begin{subfigure}{.33\textwidth}
			\centering
			\includegraphics[width=.8\linewidth]{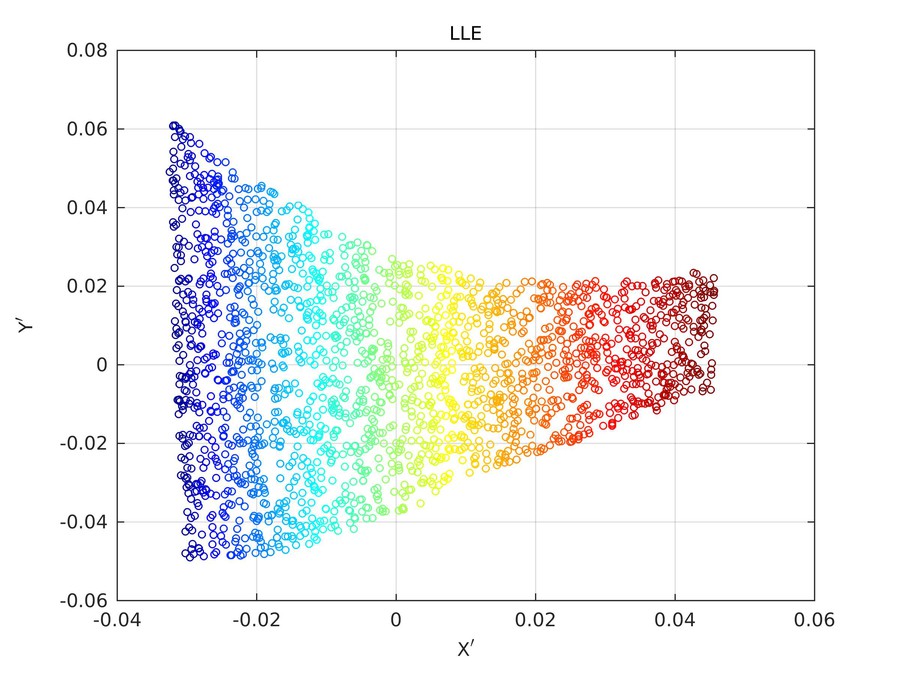}
			\caption{LLE}
			\label{fig:swiss roll LLE}
		\end{subfigure}%
		\begin{subfigure}{.33\textwidth}
			\centering
			\includegraphics[width=.8\linewidth]{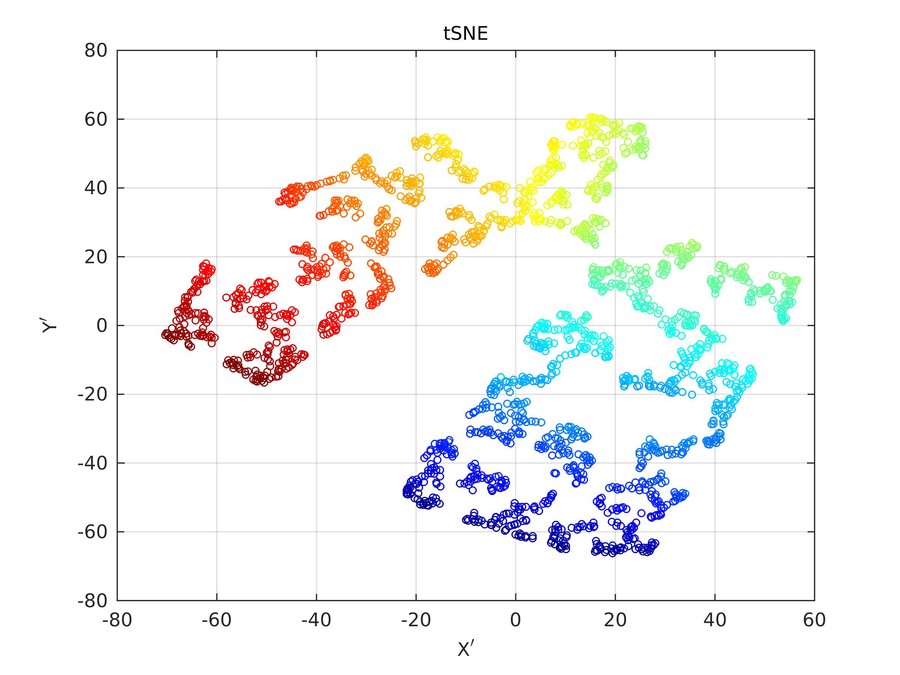}
			\caption{$t$-SNE}
			\label{fig:swiss roll tSNE}
		\end{subfigure}
		\label{fig:swiss roll results}
		\caption{
			Visualization of the Swiss Roll data with (a) original input space, (b) proposed method, (c) Sammon's projection, (d)ISOMAP, (e) LLE, and (f) $t$-SNE.}
	\end{figure*}
	The real-world data sets considered are Frey face \cite{frey_rawface}, COIL \cite{nene1996columbia}, and USPS handwritten digits \cite{frey_rawface}. Frey face consists of $1965$ gray scale images each of size $20 \times28$ pixels, i.e., the input dimension is $560$. These are of the face of a single person with different face orientations and facial emotions. The COIL data set consists of images of $20$ different objects viewed from $72$ equally spaced angels. We have considered the first object of the COIL data set. Each image is of resolution $32 \times 32$. Thus, the input dimension is $1024$ and the number of instances $=72$. USPS handwritten digits data set contains $1100$ images for each of the handwritten digits between $0$ and $9$. The images are of dimension $16 \times 16$, resulting in inputs of dimension $256$. We have considered the images of $0$
	only. We choose these data sets as they are commonly used in manifold learning studies \cite{sammon1969nonlinear,roweis2000nonlinear,maaten2008visualizing,qiao2012explicit}.
	
	\subsection{Experiment Settings \label{subsec:settings}}
	
	We compare the results obtained by the proposed method on the above mentioned data sets with four other data visualization methods: Sammon's projection, ISOMAP, LLE, and $t$-SNE. When testing the predictability of the system, the proposed method has to be compared with methods equipped with predictability. For that, we experiment on one synthetic data set (Swiss Roll) and one real-world data set (Frey face) using the proposed method and four other suitable methods, namely, out of sample extension of ISOMAP and LLE \cite{bengio2004out}, autoencoder \cite{hinton2006reducing}, and simplified NPPE (SNPPE) \cite{qiao2012explicit}. For all the synthetic data sets we apply the proposed method on the data set directly (without any processing). For image data sets, pixel values are divided by $255$ to have their values in $[0,1]$.  Following \cite{qiao2012explicit} we choose the number of nearest neighbors, $\epsilon=1\%$ of the training samples (rounded to the nearest integer). For the first object of the COIL data set (we shall refer to it as COIL), the number of instances is $72$. So we use $\epsilon=5$ instead of $\epsilon=1$ to construct a reasonable graph.  To choose the number of fuzzy rules, $n_{c}$ we perform an experiment. Details of the experiment and its results are included in the Section \ref{sec:number_of_rules} of the Supplementary Materials. From the experiment, we decide to set $n_{c}=1\%$ of the training samples (rounded to the nearest integer). Here also for the COIL data set to avoid under-fit, we choose $n_{c}=5$ instead of $n_{c}=1$. For synthetic data sets, we initialize, spreads of the Gaussian memberships, $\sigma$s in two different ways:  $0.2$ times and $0.3$ times the feature-specific range. For high dimensional real data sets, we need to choose higher values of $\sigma$s to avoid generating several zero rule-firing cases.  For $0$s of USPS data set (we shall refer to it as USPS) and Frey face we initialize $\sigma$s as $0.4$ times the feature-specific range. For COIL data set, $\sigma$s are initialized as $0.5$ times the feature-specific range. In all cases, consequent parameters are initialized with random values in ($-0.5$,$0.5$).

	For both synthetic and real-world data sets, we repeat the experiments five times for each choice of the spread. For the three synthetic data sets, there are two initial choices for spread. So  ten runs are conducted giving ten rule-based systems for each synthetic data set. On the other hand, for the three real-world data sets, we initialize spread in a single way. So a total of five runs are conducted giving five rule-based systems. Then based on the minimum value of the objective function in (\ref{eq: error}), we choose the best rule-based system for a data set.  We do not look at any test data and the best result can always be chosen in an automatic manner without any human intervention based on (\ref{eq: error}) as this gives the training error.   
	
	To optimize the objective function (\ref{eq: error})  we use an optimizer from a standard  machine learning framework, `TensorFlow' \cite{abadi2016TensorFlow}. We use the stochastic gradient descent with momentum to search an optimal solution of the given objective. We implement this with the help of `train.MomentumOptimizer' class of  `TensorFlow'. We set the values of the learning rate and the momentum to $0.1$ and $0.5$ respectively. For the four comparing data visualization methods, we apply the synthetic data sets directly. For real-world data sets, as in the case of the proposed method, we perform feature-wise zero-one normalization. For computing Sammon's projection, we use the MATLAB implementation of multi-dimensional scaling \cite{borg2003modern}, \textit{mdscale} with the parameter `Criterion' set to  `sammon'.
	The dimension  of the output is set to two. For other parameters, the default values are used. The other three methods are implemented using `scikit-learn'(sklearn) package  \cite{scikit-learn} (version $0.19.2$) of python. The classes `manifold.Isomap', `manifold.LocallyLinearEmbedding', `mani-fold.TSNE' of `sklearn'  are used to implement ISOMAP, LLE, and $t$-SNE, respectively. As mentioned in Section \ref{sec:literature} these three methods preserve the local structure for which they consider a neighborhood. For comparing the results we use the same neighborhood size for all three methods and the proposed method. For LLE and ISOMAP for every data set, we use the same neighborhood size (number of nearest neighbors) as used for the proposed method. For $t$-SNE, there is no provision of setting the neighborhood size directly. The `perplexity' \cite{maaten2008visualizing} parameter of $t$-SNE provides an indirect measure of the effective number of neighbors.
	So we set the `perplexity' parameter to the number of nearest neighbors used in the proposed method. For these three methods, all the other parameters are kept at their default values with respect to the routines we use to implement them. For predictability testing, except for SNPPE, the other three comparing methods are executed using Matlab Toolbox for Dimensionality Reduction (drtoolbox) by Laurens van der Maaten \cite{drtoolbox}. The results of SNPPE is generated by the code provided by the authors of SNPPE. In these cases also the neighborhood size, when required, is set as the value used in the proposed method. All the other parameters are kept at their default values.
	\subsection{Results and Comparisons \label{subsec:results}}
	From Figs. \ref{fig:swiss roll proposed}, \ref{fig:s curve proposed}, and \ref{fig:helix proposed} we can observe that the proposed method successfully unfolds the non-linear structures present respectively in the Swiss Roll, S Curve, and Helix data to  nearly linear structure. Employing the geodesic distance as the pairwise input distance in Sammon's stress the proposed method has resulted in a significant departure from the result  of Sammon's projection. For all three data sets, Sammon's projection can not unfold the original data to its intrinsic linear structure but it tries to preserve the global geometric structure. ISOMAP tries to preserve the pairwise distances over the manifold by using the geodesic distance in the classical multidimensional scaling error function \cite{tenenbaum2000global}. Preserving geodesic distance, ISOMAP is able to unfold the non-linear structure as shown in Figs. \ref{fig:swiss roll ISOMAP} and \ref{fig:s curve ISOMAP}. But its drawback is large distances play a stronger role compared to small distances in its objective. As a result, in the case of the Helix data set, as seen in Fig. \ref{fig:helix ISOMAP}, it can not unfold the data as desired. Both, LLE and t-SNE try to capture the local structure present in the original data, in the generated lower dimensional space. Figures \ref{fig:swiss roll LLE}, \ref{fig:s curve LLE}, and \ref{fig:helix LLE} reveal that the  LLE manages to unfold the underlying structure to some extent but the quality of unfolding is far from that by the proposed method and ISOMAP. Similarly, t-SNE although has the same aim, is not able to unfold the data sets properly as revealed by Figs. \ref{fig:swiss roll tSNE}, \ref{fig:s curve tSNE} and \ref{fig:helix tSNE}. In summary, of the five methods, only the fuzzy rule-based method is successful for all three data sets and the next best performing method is ISOMAP.
	
\begin{figure*}[!tb]
		\centering
		\centering
		\includegraphics[width=.8\linewidth]{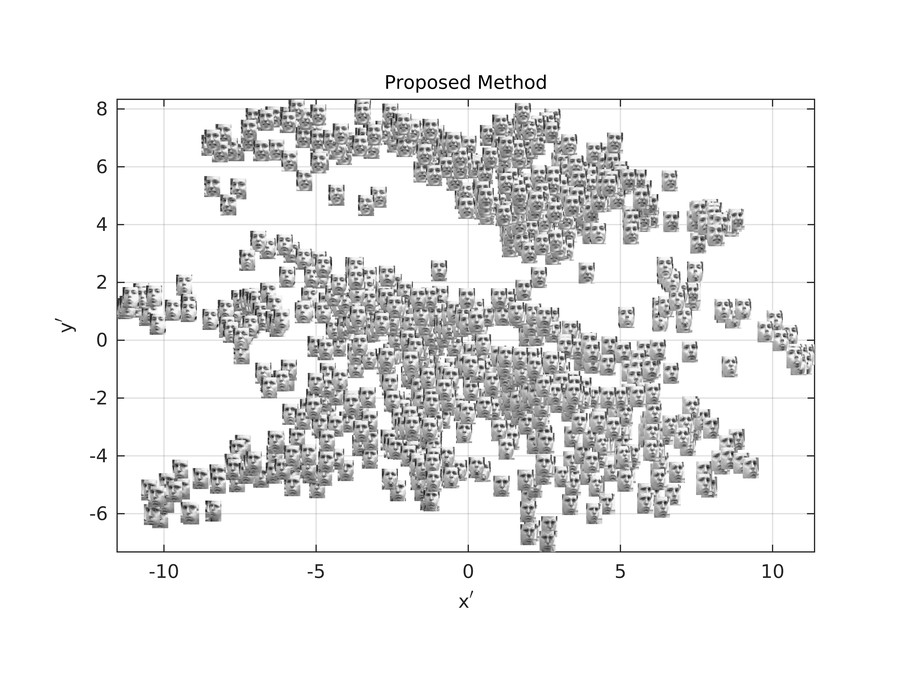}
		\caption{Visualization of the Frey face data set with the proposed method.}
		\label{fig:frey face proposed}
	\end{figure*}

	The proposed method can also faithfully project the real-world data sets in two dimensions. In Fig. \ref{fig:frey face proposed}, two dimensional embedding of the Frey-face data set is shown. 
	Figure \ref{fig:frey face proposed} shows that the gradual change in the face orientation from right to left is roughly mapped adjacently on the $2$D along with the increment of the extracted feature $x'$. Similarly, the change in expressed emotion varies roughly with the extracted feature $y'$. Smiling, neutral and annoyed faces are mapped in the top, middle and bottom regions, respectively. The results of the four comparing methods are placed in the Supplementary Materials. Sammon's projection being a Euclidean distance preserving method has placed the faces expressing different emotions in comparatively overlapping regions as seen in Fig. \ref{fig:frey face sammons}. Other neighborhood preserving methods have successfully unfolded the high dimensional manifold. Comparing the results in Figs. \ref{fig:frey face ISOMAP}, \ref{fig:frey face LLE}, and \ref{fig:frey face t-SNE} with the result in Fig. \ref{fig:frey face proposed} we can infer that the proposed method has mapped the variations of frey face data set comparatively in a more consistent and expected manner.  Figure \ref{fig: coil object1 results} shows results of the two dimensional embedding of instances of the COIL data set.  For this data set also, we got the expected outcome from the proposed method (Fig. \ref{fig:coil object1 proposed}) as well as from ISOMAP (Fig. \ref{fig:coil object1 ISOMAP}) and $t$-SNE (Fig. \ref{fig:coil object1 tSNE}). However, results obtained from Sammon's projection and LLE shown in Figs. \ref{fig:coil object1 sammons} and \ref{fig:coil object1 LLE} cannot recover the manifold faithfully on the two dimensional space. Fig. \ref{fig:usps_class_0_proposed} shows visualization of the handwritten character $0$s of the USPS data set. 

	In this case also, the proposed method successfully projects zeros with different orientations and line widths in different regions in a continuous manner. The projections by the other four methods as shown in Figs. \ref{fig:usps_sammons}, \ref{fig:usps_ISOMAP}, \ref{fig:usps_LLE}, \ref{fig:usps_t-SNE}  reveals that LLE (Fig. \ref{fig:usps_LLE}) and $t$-SNE (Fig. \ref{fig:usps_t-SNE})  clearly failed to represent the data in a way so that orientations and line widths of the character vary smoothly in the projected space.

	\subsection{Impact of Initial Rules \label{subsec:impact_initial_rule}}
	The proposed method initializes the rule antecedents using the centroids of the clusters obtained by Algorithm \ref{alg:GeoCMeans}. These centroids are used as centers of Gaussian membership functions of the fuzzy sets which compose the rule antecedents. To show the impact of the initial rules we have initialized the rule antecedents using uniformly distributed random values generated from the smallest hyperbox containing the training data.   The obtained projection for Swiss Roll data is shown in Fig. \ref{fig:swiss_roll_random_antecedent}. Figure \ref{fig:swiss_roll_cluster_center_antecedent} shows the result obtained by the proposed method with rule antecedents defined by the cluster centroids. From Figs. \ref{fig:swiss_roll_cluster_center_antecedent} and \ref{fig:swiss_roll_random_antecedent} it is evident that with or without judicious initialization of rule antecedents, the proposed method can unfold the non-linear structure satisfactorily. This observation is also true  in case of the Frey face data set. Fig. \ref{fig:frey_face_random_antecedent} shows the visualization of Frey face data set using the proposed method with random antecedents initialized as mentioned above. To investigate further, we have computed the average of final error values over the ten/five runs (performed as explained in sub-section \ref{subsec:settings}) for the two considered data sets with both types of rule antecedent initializations. The error referred here is the error function in equation (\ref{eq: error}). Table \ref{tab:average_error} shows that the averages of the final error values are less in case of initialization of the rule antecedents by cluster centroids compared to random rule antecedents for both data sets. This suggests that although in both cases the algorithm lands up in useful local minima, for the cluster-based initial rules, generally the minima are superior.
	
	\begin{table}[!tb]
		\caption{Average of the final error values for different initializations}\label{tab:average_error}
		\begin{center}
			\begin{tabular}{lcc}
				\hline
				{Data Set } & \multicolumn{2}{c}{Initial rule antecedents defined by}\\
				{ } &  {Cluster centroids} & { Random}\\
				\hline
				Swiss Roll & 0.04740	& 0.06490 \\
				Frey face & 0.50459 & 0.50500 \\
				\hline
			\end{tabular}
		\end{center}
	\end{table}
	 We have also done some experiments with rules which are initialized with random values in [0,1] hypercube. We have done 10 such experiments on the Swiss Roll data. The results are shown in Fig. \ref{fig:swiss_roll_initialization_random_zero_one}. Even in this case, the results are reasonably good, but not as good as the cases with judiciously initialized rules. In a few cases, the system could not unfold. Thus the system is quite robust with respect to the initial choice of rules, but certainly cluster-based rules help.
	\subsection{Validation of Predictability \label{subsec:predictability}}
	One of the key advantages of utilizing a fuzzy rule-based system is that it is parametric. The antecedent and consequent parameters of the TS model are learned from the given data. Consequently, we get an explicit mapping function that maps the high dimensional inputs on a lower dimensional one. So, after the system is trained it is possible to predict the lower dimensional embedding for new test points. As mentioned earlier, to compare the predictability performance of the proposed method we consider four methods: out of sample extension of ISOMAP and LLE \cite{bengio2004out} and two parametric methods: autoencoder\cite{hinton2006reducing} and SNPPE \cite{qiao2012explicit}. Note that SNPPE involves a polynomial mapping. Following \cite{qiao2012explicit} we use $2$nd order and $3$rd order polynomial based SNPPE.
	
\begin{figure*}[!tb]
		\centering
		\begin{subfigure}{.24\textwidth}
			\centering
			\includegraphics[width=.95\linewidth]{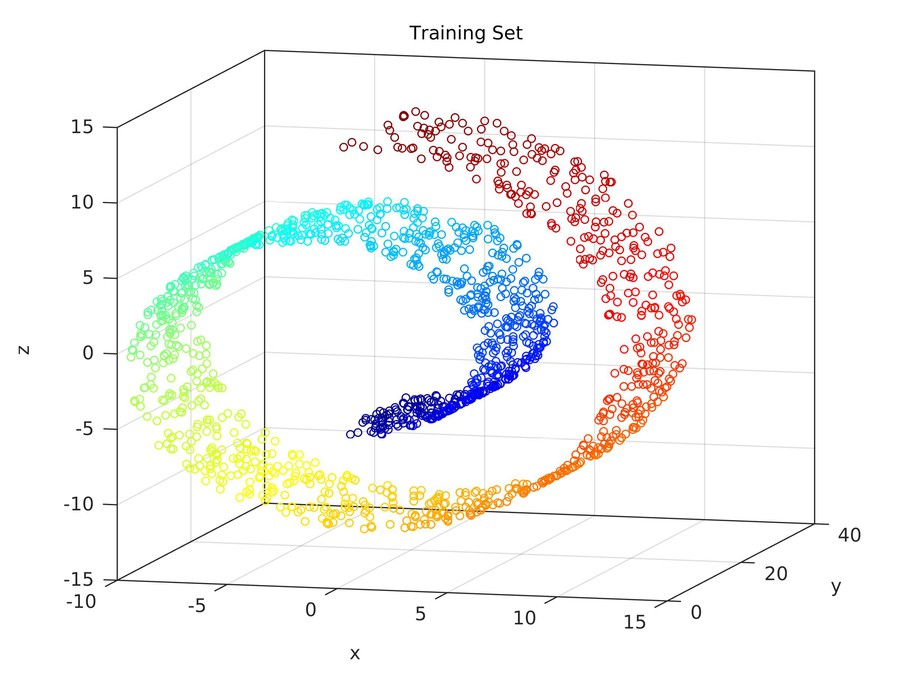}
			\caption{}
			\label{fig:swiss roll training set 1}
		\end{subfigure}%
		\begin{subfigure}{.24\textwidth}
			\centering
			\includegraphics[width=.95\linewidth]{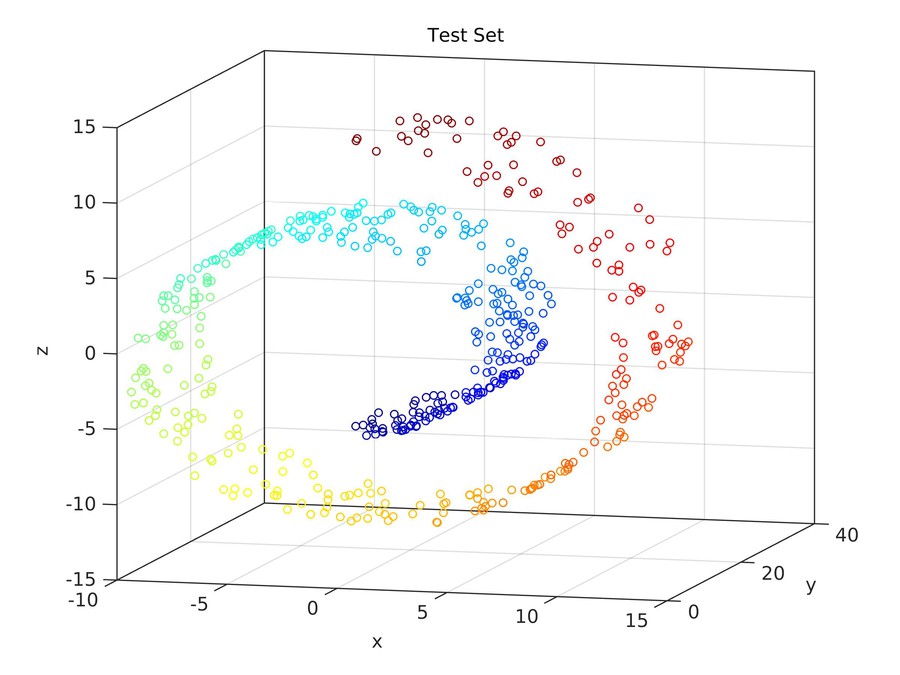}
			\caption{}
			\label{fig:swiss roll test set 1}
		\end{subfigure}%
		\begin{subfigure}{.24\textwidth}
			\centering
			\includegraphics[width=.95\linewidth]{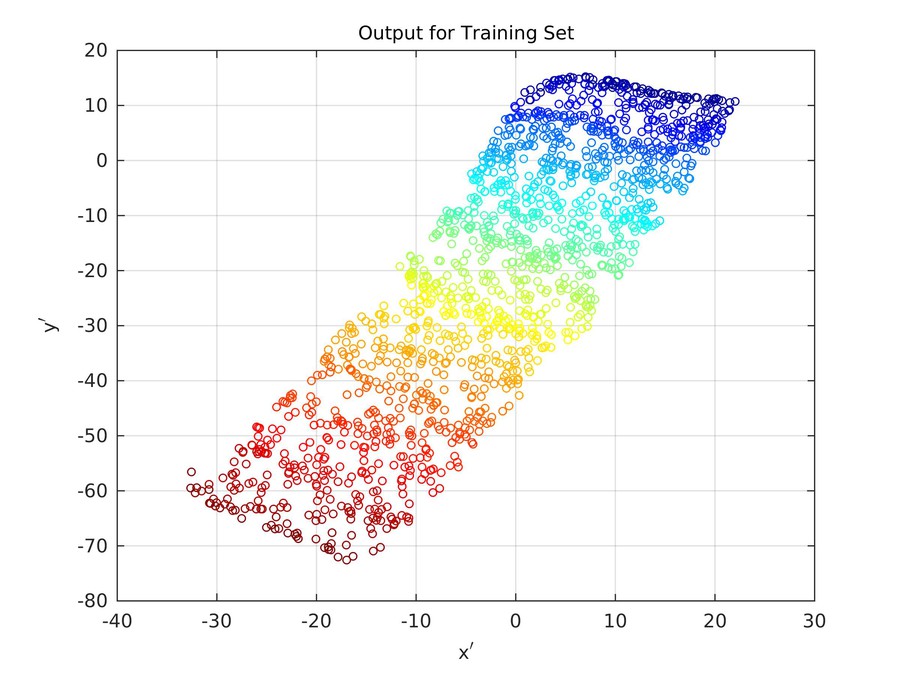}
			\caption{}
			\label{fig:swiss roll output for training set 1}
		\end{subfigure}%
		\begin{subfigure}{.24\textwidth}
			\centering
			\includegraphics[width=.95\linewidth]{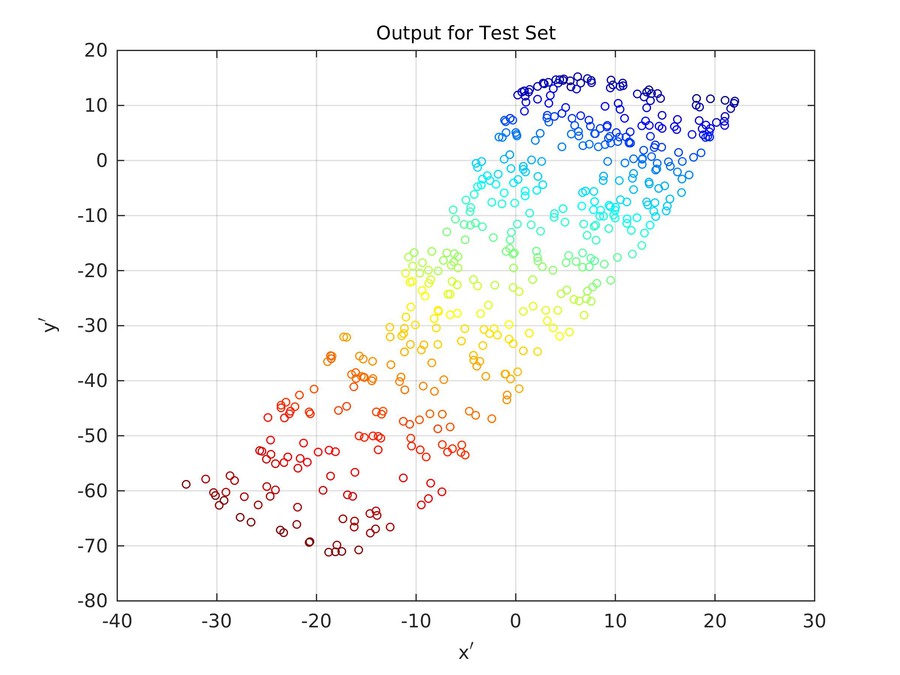}
			\caption{}
			\label{fig:swiss roll output for test set 1}
		\end{subfigure}
		\caption{For experiment 1 on validation of predictability with the Swiss Roll data : (a) Training Set, (b) Test Set, (c) Proposed method output for Training Set, and (d) Proposed method output for Test Set. }
		\label{fig:swiss roll results for training data set 1 and test data set 1}
	\end{figure*}

	To validate the predictability of the system, we have designed three tests:
	\begin{enumerate}
		\item Randomly dividing the data set into training and test sets.
		\item Using a contiguous portion of the manifold as the test set and rest as the training set.
		\item Leave one out validation.
	\end{enumerate}
	We have performed the first test on the Swiss Roll and Frey face data sets. We use $75\%$ of the Swiss Roll data set, i.e.,  $1500$ randomly selected points to from the training set and rest $25\%$,  i.e., $500$ points to form the test set. The training and test data sets are shown in Fig. \ref{fig:swiss roll training set 1} and Fig. \ref{fig:swiss roll test set 1} respectively.
 While partitioning the data set, the associated label vectors are also partitioned accordingly. Figure \ref{fig:swiss roll output for training set 1} and Fig. \ref{fig:swiss roll output for test set 1} are the results corresponding to the training and test data sets by the proposed method. From Fig. \ref{fig:swiss roll output for test set 1} it is evident the predicted locations determined by the proposed method for the test points are at their expected positions in the lower dimension. Figs. \ref{fig:ISOMAP_swiss_roll_out_of_sample}, \ref{fig:LLE_swiss_roll_out_of_sample}, \ref{fig:autoencoder_swiss_roll_out_of_sample}, \ref{fig:SNPPE2_swiss_roll_out_of_sample}, and \ref{fig:SNPPE3_swiss_roll_out_of_sample} show training and test set representation in lower dimension by ISOMAP, LLE, autoencoder, $2$nd order SNPPE, and $3$rd order SNPPE respectively. Except for the autoencoder, the other comparing methods successfully unfold the Swiss Roll structure in the training phase and appropriately predict the test points. From Fig. \ref{fig:autoencoder_swiss_roll_out_of_sample} we observe that the autoencoder based method is unable to preserve the geometry of the training data. In Figs. \ref{fig:swiss roll output for test set 1}, \ref{fig:ISOMAP_swiss_roll_ts1}, \ref{fig:LLE_swiss_roll_ts1}, \ref{fig:SNPPE2_swiss_roll_ts1}, and \ref{fig:SNPPE3_swiss_roll_ts1} color assignment to the projected test points by the label vector shows that the test data set is unfolded in the desired manner in all five cases. For the Frey face data set, we have selected 15 random instances as the test set and rest of the points as the training set. Figure \ref{fig:frey_face_training_random_test} illustrates the projected output by the proposed method for training and test  data together where the images with black rectangular borders correspond to the test instances. For all comparing methods on this data we follow the same representation of the training and test data. In Figs. \ref{fig:frey_face_training_random_test}, \ref{fig:frey_face_ISOMAP_out_of_sample}, \ref{fig:frey_face_LLE_out_of_sample} we can see that the test instances are projected to the regions where the training instances have the same/similar facial expression and orientation. Thus, for high dimensional image data sets also the proposed method, as well as LLE and ISOMAP, are successful in prediction. The autoencoder based lower dimensional projection, shown in fig. \ref{fig:frey_face_autoencoder} clustered different facial expressions and different face orientations in different regions. However, autoencoder based method is unable to unfold the intrinsic structure properly as in some cases it placed the opposite face orientations (left and right) closer to each other compared to the faces in straight orientations. A similar placement has occurred for opposite expressions (e.g. happy and annoyed) and neutral expressions. From Figs. \ref{fig:frey_face_SNPPE2}, \ref{fig:frey_face_SNPPE3} it is evident that SNPPE based methods neither unfolded the higher dimensional data nor predicted the positions of the test points in a nice way.  
	
\begin{figure*}[!tb]
		\centering
		\begin{subfigure}{.24\textwidth}
			\centering
			\includegraphics[width=.95\linewidth]{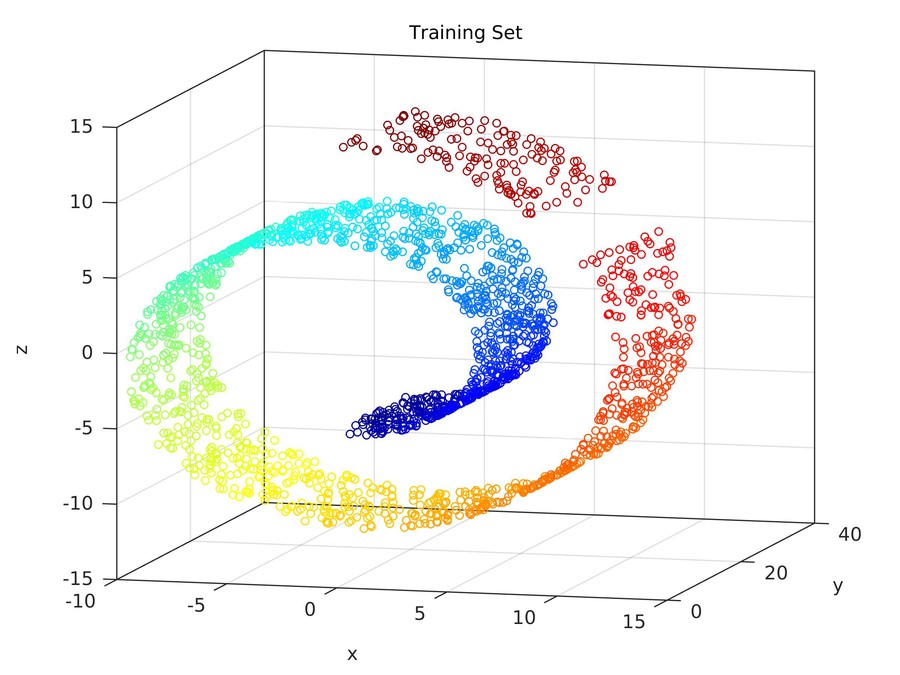}
			\caption{}
			\label{fig:swiss roll training set 2}
		\end{subfigure}%
		\begin{subfigure}{.24\textwidth}
			\centering
			\includegraphics[width=.95\linewidth]{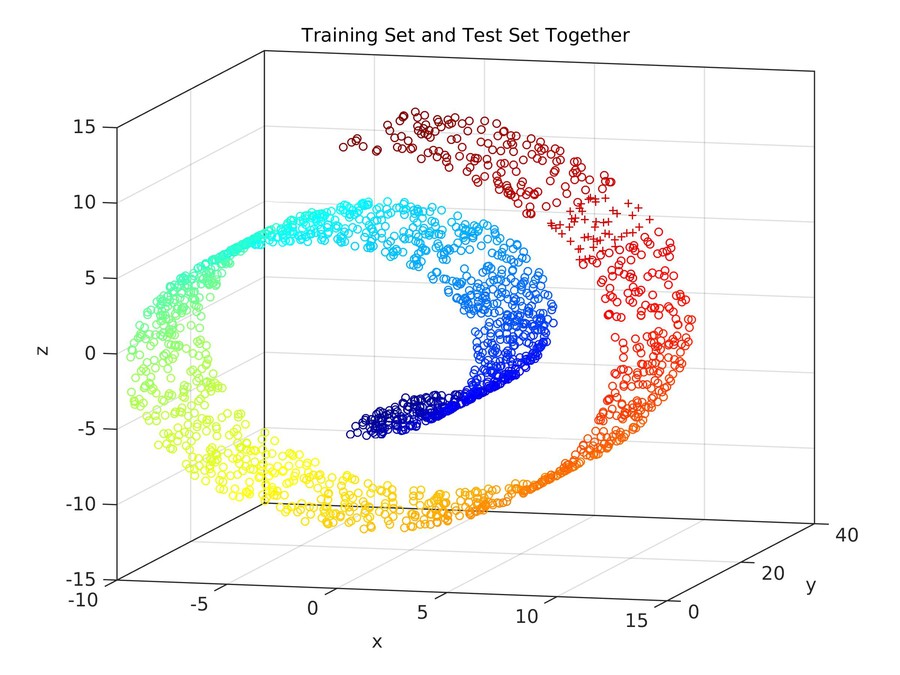}
			\caption{}
			\label{fig:swiss roll test set 2}
		\end{subfigure}
		\begin{subfigure}{.24\textwidth}
			\centering
			\includegraphics[width=.95\linewidth]{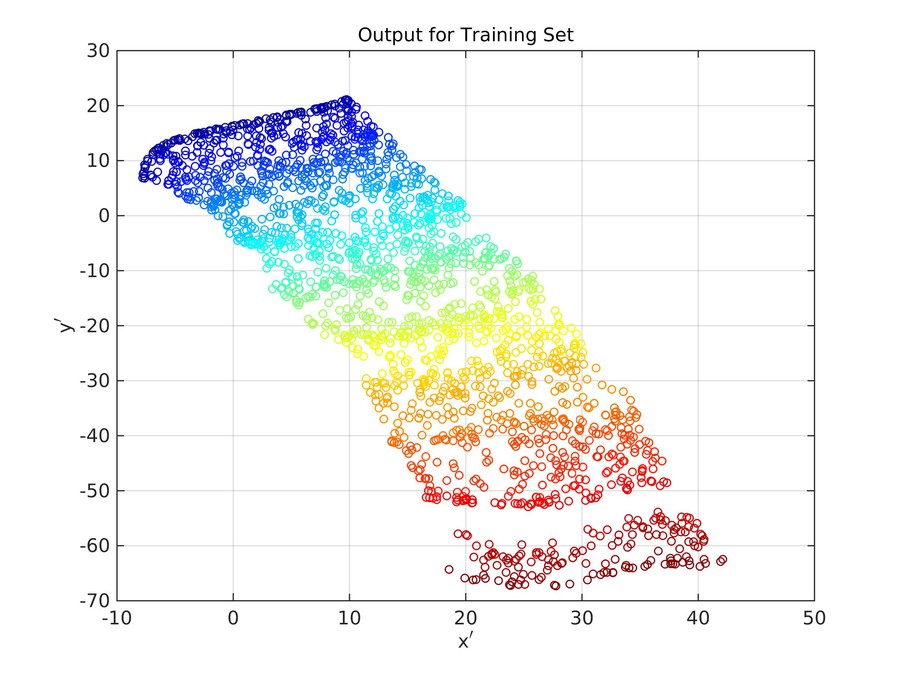}
			\caption{}
			\label{fig:swiss roll output for training set 2}
		\end{subfigure}%
		\begin{subfigure}{.24\textwidth}
			\centering
			\includegraphics[width=.95\linewidth]{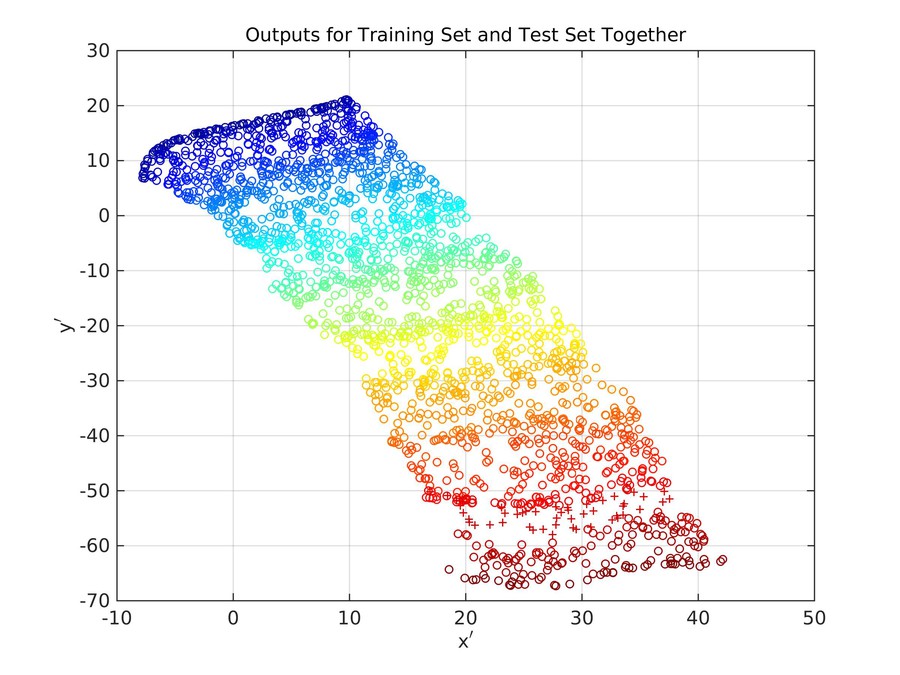}
			\caption{}
			\label{fig:swiss roll output for test set 2}
		\end{subfigure}
		\label{fig:swiss roll results for training data set 2 and test data set 2}
		\caption{For experiment 2 on validation of predictability with the Swiss Roll data : (a) Training Set, (b) Training Set and Test Set, (c) Proposed method output for Training Set, and (d) Proposed method outputs for Training Set and Test Set. }
	\end{figure*}

	The second test is also done with the Swiss Roll data set. A contiguous portion of the data set consisting of fifty points has been removed and the rest $1950$ points serve as the training data set as seen in Fig. \ref{fig:swiss roll training set 2}. The removed portion serves as the test set. Figure \ref{fig:swiss roll output for training set 2} displays the output of the proposed method corresponding to the training set. Figure \ref{fig:swiss roll output for test set 2} shows the test set outputs of the proposed method along with the training set. The test set points are demarcated with plus ($+$) symbol. For the comparing methods we represent the results in the same manner. Figures \ref{fig:ISOMAP_swiss_roll_tr5}, \ref{fig:LLE_swiss_roll_tr5}, \ref{fig:Autoencoder_swiss_roll_tr5}, \ref{fig:SNPPE2_swiss_roll_tr5}, and \ref{fig:SNPPE3_swiss_roll_tr5} show the training set projections in lower dimensional space for ISOMAP, LLE, autoencoder, $2$nd order SNPPE, and $3$rd order SNPPE respectively. Similarly, Figs. \ref{fig:ISOMAP_swiss_roll_trts5}, \ref{fig:LLE_swiss_roll_trts5}, \ref{fig:Autoencoder_swiss_roll_trts5}, \ref{fig:SNPPE2_swiss_roll_trts5}, and \ref{fig:SNPPE3_swiss_roll_trts5} show the test set outputs (demarcated with plus ($+$) symbol) along with the training set for ISOMAP, LLE, autoencoder, $2$nd order SNPPE, and $3$rd order SNPPE respectively. The results show that the proposed method, as well as the comparing methods except for the autoencoder, have appropriately interpolated the test points.

	The third test uses the leave one out validation on the COIL data set. Out of the $72$ instances, each one is considered as the test instance and rest $71$ instances are used as the training set. So, for $72$ such training set-test sets outputs are generated.
	Results of six such tests are displayed in the Supplementary Materials. By choosing every twelfth instance in the given sequence of $72$ instances as test instance, six partitions are generated. We name them Set 1, Set 2 and so on upto Set 6. Test instances used in Set 1 to Set 6 are instance number 1, 13, 25, 37, 49 and 51 respectively. For every set all the other data points except the corresponding test point form the training set. Results corresponding to these six test sets are shown in Figs. \ref{fig:coil_dataset_leave_one_out_result_set}. 
	The training points are demarcated in green and the test point is demarcated in red. Instead of showing the images of the object, the consecutive change in the viewing angle of the images are represented by consecutive numbers. In most of the cases, it is found that the test object is placed in the desired position, i.e., in between points corresponding to input images captured in plus minus five degrees of the viewing angle of the test point.

	\subsection{Rejection of Outputs}
	The proposed method is capable of identifying the test cases where the system may not produce reliable outputs. In a TS type fuzzy rule-based system, the output is computed by (\ref{eq:output}). The output is a convex sum of the individual rule outputs. Individual rule outputs are multiplied by the corresponding firing strengths. The rule having the maximum firing strength mainly characterizes the output. Essentially for test points far away from the training data, no fuzzy rule is expected to fire strongly. Consequently, for those points, the outputs obtained may not be reliable. Figure \ref{fig:swiss roll training set 3 and test set 3} shows a set of training points and test points. The training points are demonstrated with blue markers and test points with red markers. Test points are located away from the training points. The histogram of the maximum rule firing strengths is shown in Fig. \ref{fig: Histogram of max rule firing}. Here also training and test firing strengths are indicated with blue and red colours respectively. It is clear from Fig. \ref{fig: Histogram of max rule firing} that the maximum rule firing strengths of the test points are  much less compared to the maximum rule firings of the training points. This information can be used to decide when the system should reject its output. For example, if the maximum firing strength is less than 0.15, the system output can be rejected. Note that this threshold can be decided based on the training data only. In Fig. \ref{fig:undesired_output} we have depicted the projected outputs for the training and test data shown in Fig. \ref{fig:swiss roll training set 3 and test set 3} without exercising the rejection option. It is clear that the prediction by the proposed method for the test points (red circles) is inappropriate and should be discarded. It is worthy to mention that the other dimensionality reduction schemes which have predictability such as the methods mentioned in the previous sub-section do not have any provision for rejecting outputs.
	
	\subsection{Generalized Nature of the Proposed Model}
	The proposed model gives a general framework for learning an unsupervised fuzzy rule-based system for manifold learning or data projection. Instead of using the objective function in (\ref{eq: error}), we can use different objectives also. For example, the objective function of Sammon's projection (\ref{eq: sammon}) can be used. Figure \ref{fig:generalization of proposed model} shows the Swiss Roll data and its corresponding lower dimensional output using the proposed method with (\ref{eq: sammon}) as the objective function for learning.  Similarly, Figs. \ref{fig:SCurveProposedMethodGeneralization} and \ref{fig:HelixProposedMethodGeneralization} depict the output produced by the fuzzy rule-based system using (\ref{eq: sammon}) as the objective function for the S Curve and Helix data, respectively. In both cases the projection reveals what Sammon's method  is expected to do. For these experiments, the other settings are kept the same as described in subsection \ref{subsec:settings}. Note that, this system can predict the projections for test points and that is an advantage over the usual Sammon's method. We also note that unlike the fuzzy system in \cite{pal2002fuzzy}, the proposed system does not require any target output for learning.

	\section{Conclusion}\label{sec:conclusion}
	In this paper, we have proposed a framework for designing a Takagi Sugeno fuzzy rule-based system to reduce the dimensionality of data for visualization purposes. The proposed system can learn a manifold to make a lower dimensional representation. We have also proposed an appropriate scheme for learning the parameters of the system. Fuzzy rule-based systems are efficient for various machine learning tasks but unfortunately, almost unexplored for manifold learning or data visualization purposes.  Here, the rule base is designed in such a manner that the geodesic distance structure is preserved as the Euclidean distance in the projected space.   We have tested our system on three synthetic  and three high dimensional real-world data sets (image data sets) that are popular for bench-marking manifold learning algorithms. The results show that our approach can successfully unfold the high dimensional data sets on a low dimensional space. For both synthetic and real-world data sets, it provides good visual outputs. We have compared the performance of the proposed system with Sammon's method, ISOMAP, LLE, and $t$-SNE. The proposed fuzzy rule-based system is found to perform either better or comparable. The additional advantage of the proposed scheme is that it is a parametric system and consequently, an explicit mapping for projecting high dimensional data into a lower dimensional space is obtained. Using this mapping, output for new points can be predicted, i.e., the underlying system is equipped with predictability. The predictability of the proposed model is validated using three types of experiments involving both synthetic and real-world data sets.  If a test data point is located far from the training set then a machine learning system should not produce any output. Most machine learning systems do not have the capability of rejection when it should. Our proposed system can reject an output when it should. 
	
	One of the drawbacks of the proposed scheme like many other manifold learning schemes is that, it is sensitive to the number of neighbors that is used to construct the neighborhood graph. In our study the number of fuzzy rules has been chosen manually. The choice of the number of fuzzy rules is related to the number of clusters we choose for clustering the training data. We did not explore the possibilities of using cluster validity indices or other schemes to decide on the optimal number of clusters as this was not the primary objective of this study. Moreover, a cluster validity index helps to find the `optimal' number of clusters in the pattern recognition sense. But here even if the data do not have any cluster in the pattern recognition sense, we can use clustering to find the initial rule-base. We have restricted our work to data sets lying on a single manifold. Through neighborhood graph construction we can only approximate the geodesic distance over a single manifold or overlapping manifolds. But if the data lie in two (or more) widely separated manifolds, using a reasonable number of nearest neighbors we are not likely to get a connected neighborhood graph. We have not proposed any technique to approximate geodesic distance over a multi component neighborhood graph.
	
\bibliographystyle{IEEEtr}
\bibliography{IEEEabrv,ref}

\begin{thebibliography}{10}

\bibitem{alvarez2017kernel}
A.~M. {\'A}lvarez-Meza, J.~A. Lee, M.~Verleysen, and G.~Castellanos-Dominguez,
  ``Kernel-based dimensionality reduction using renyi's $\alpha$-entropy
  measures of similarity,'' {\em Neurocomputing}, vol.~222, pp.~36--46, 2017.

\bibitem{meng2008nonlinear}
D.~Meng, Y.~Leung, T.~Fung, and Z.~Xu, ``Nonlinear dimensionality reduction of
  data lying on the multicluster manifold,'' {\em IEEE Transactions on Systems,
  Man, and Cybernetics, Part B (Cybernetics)}, vol.~38, no.~4, pp.~1111--1122,
  2008.

\bibitem{talwalkar2008large}
A.~Talwalkar, S.~Kumar, and H.~Rowley, ``Large-scale manifold learning,'' in
  {\em Computer Vision and Pattern Recognition, 2008. CVPR 2008. IEEE
  Conference on}, pp.~1--8, IEEE, 2008.

\bibitem{van2009dimensionality}
L.~Van Der~Maaten, E.~Postma, and J.~Van~den Herik, ``Dimensionality reduction:
  a comparative,'' {\em J Mach Learn Res}, vol.~10, pp.~66--71, 2009.

\bibitem{wang2018perception}
Y.~Wang, K.~Feng, X.~Chu, J.~Zhang, C.-W. Fu, M.~Sedlmair, X.~Yu, and B.~Chen,
  ``A perception-driven approach to supervised dimensionality reduction for
  visualization,'' {\em IEEE transactions on visualization and computer
  graphics}, vol.~24, no.~5, pp.~1828--1840, 2018.

\bibitem{chernoff1973use}
H.~Chernoff, ``The use of faces to represent points in k-dimensional space
  graphically,'' {\em Journal of the American statistical Association},
  vol.~68, no.~342, pp.~361--368, 1973.

\bibitem{duda2001pattern}
R.~O. Duda, D.~G. Stork, and P.~E. Hart, ``Pattern classification,'' 2001.

\bibitem{borg2003modern}
I.~Borg and P.~Groenen, ``Modern multidimensional scaling: theory and
  applications,'' {\em Journal of Educational Measurement}, vol.~40, no.~3,
  pp.~277--280, 2003.

\bibitem{cayton2005algorithms}
L.~Cayton, ``Algorithms for manifold learning,'' {\em Univ. of California at
  San Diego Tech. Rep}, vol.~12, no.~1-17, p.~1, 2005.

\bibitem{hotelling1936relations}
H.~Hotelling, ``Relations between two sets of variates,'' {\em Biometrika},
  vol.~28, no.~3/4, pp.~321--377, 1936.

\bibitem{fisher1936use}
R.~A. Fisher, ``The use of multiple measurements in taxonomic problems,'' {\em
  Annals of eugenics}, vol.~7, no.~2, pp.~179--188, 1936.

\bibitem{akaike1987factor}
H.~Akaike, ``Factor analysis and aic,'' in {\em Selected Papers of Hirotugu
  Akaike}, pp.~371--386, Springer, 1987.

\bibitem{he2004locality}
X.~He and P.~Niyogi, ``Locality preserving projections,'' in {\em Advances in
  neural information processing systems}, pp.~153--160, 2004.

\bibitem{tuv2009feature}
E.~Tuv, A.~Borisov, G.~Runger, and K.~Torkkola, ``Feature selection with
  ensembles, artificial variables, and redundancy elimination,'' {\em Journal
  of Machine Learning Research}, vol.~10, no.~Jul, pp.~1341--1366, 2009.

\bibitem{chandrashekar2014survey}
G.~Chandrashekar and F.~Sahin, ``A survey on feature selection methods,'' {\em
  Computers \& Electrical Engineering}, vol.~40, no.~1, pp.~16--28, 2014.

\bibitem{nag2016multiobjective}
K.~Nag and N.~R. Pal, ``A multiobjective genetic programming-based ensemble for
  simultaneous feature selection and classification,'' {\em IEEE transactions
  on cybernetics}, vol.~46, no.~2, pp.~499--510, 2016.

\bibitem{chung2018feature}
I.-F. Chung, Y.-C. Chen, and N.~R. Pal, ``Feature selection with controlled
  redundancy in a fuzzy rule based framework,'' {\em IEEE Transactions on Fuzzy
  Systems}, vol.~26, no.~2, pp.~734--748, 2018.

\bibitem{sammon1969nonlinear}
J.~W. Sammon, ``A nonlinear mapping for data structure analysis,'' {\em IEEE
  Transactions on computers}, vol.~100, no.~5, pp.~401--409, 1969.

\bibitem{roweis2000nonlinear}
S.~T. Roweis and L.~K. Saul, ``Nonlinear dimensionality reduction by locally
  linear embedding,'' {\em science}, vol.~290, no.~5500, pp.~2323--2326, 2000.

\bibitem{tenenbaum2000global}
J.~B. Tenenbaum, V.~De~Silva, and J.~C. Langford, ``A global geometric
  framework for nonlinear dimensionality reduction,'' {\em science}, vol.~290,
  no.~5500, pp.~2319--2323, 2000.

\bibitem{donoho2003hessian}
D.~L. Donoho and C.~Grimes, ``Hessian eigenmaps: Locally linear embedding
  techniques for high-dimensional data,'' {\em Proceedings of the National
  Academy of Sciences}, vol.~100, no.~10, pp.~5591--5596, 2003.

\bibitem{belkin2003laplacian}
M.~Belkin and P.~Niyogi, ``Laplacian eigenmaps for dimensionality reduction and
  data representation,'' {\em Neural computation}, vol.~15, no.~6,
  pp.~1373--1396, 2003.

\bibitem{hinton2006reducing}
G.~E. Hinton and R.~R. Salakhutdinov, ``Reducing the dimensionality of data
  with neural networks,'' {\em science}, vol.~313, no.~5786, pp.~504--507,
  2006.

\bibitem{maaten2008visualizing}
L.~v.~d. Maaten and G.~Hinton, ``Visualizing data using t-sne,'' {\em Journal
  of machine learning research}, vol.~9, no.~Nov, pp.~2579--2605, 2008.

\bibitem{bengio2004out}
Y.~Bengio, J.-f. Paiement, P.~Vincent, O.~Delalleau, N.~L. Roux, and M.~Ouimet,
  ``Out-of-sample extensions for lle, isomap, mds, eigenmaps, and spectral
  clustering,'' in {\em Advances in neural information processing systems},
  pp.~177--184, 2004.

\bibitem{takagi1985fuzzy}
T.~Takagi and M.~Sugeno, ``Fuzzy identification of systems and its applications
  to modeling and control,'' {\em IEEE Transactions on Systems, Man, and
  Cybernetics}, no.~1, pp.~116--132, 1985.

\bibitem{ng1997fuzzy}
S.~K. Ng and H.~J. Chizeck, ``Fuzzy model identification for classification of
  gait events in paraplegics,'' {\em IEEE Transactions on Fuzzy Systems},
  vol.~5, no.~4, pp.~536--544, 1997.

\bibitem{bezdek1984fcm}
J.~C. Bezdek, R.~Ehrlich, and W.~Full, ``Fcm: The fuzzy c-means clustering
  algorithm,'' {\em Computers \& Geosciences}, vol.~10, no.~2-3, pp.~191--203,
  1984.

\bibitem{pal2002fuzzy}
N.~R. Pal, V.~K. Eluri, and G.~K. Mandal, ``Fuzzy logic approaches to structure
  preserving dimensionality reduction,'' {\em IEEE Transactions on Fuzzy
  Systems}, vol.~10, no.~3, pp.~277--286, 2002.

\bibitem{das2019unsupervised}
S.~Das and N.~R. Pal, ``An unsupervised fuzzy rule-based method for structure
  preserving dimensionality reduction with prediction ability,'' in {\em IFIP
  International Conference on Artificial Intelligence Applications and
  Innovations}, pp.~413--424, Springer, 2019.

\bibitem{cunningham2015linear}
J.~P. Cunningham and Z.~Ghahramani, ``Linear dimensionality reduction: Survey,
  insights, and generalizations,'' {\em The Journal of Machine Learning
  Research}, vol.~16, no.~1, pp.~2859--2900, 2015.

\bibitem{switzer1984Min}
P.~Switzer and A.~Green, ``Min/max autocorrelation factors for multivariate
  spatial imagery: Technical report 6,'' 01 1984.

\bibitem{larsen2002decomposition}
R.~Larsen, ``Decomposition using maximum autocorrelation factors,'' {\em
  Journal of Chemometrics: A Journal of the Chemometrics Society}, vol.~16,
  no.~8-10, pp.~427--435, 2002.

\bibitem{he2005neighborhood}
X.~He, D.~Cai, S.~Yan, and H.-J. Zhang, ``Neighborhood preserving embedding,''
  in {\em Computer Vision, 2005. ICCV 2005. Tenth IEEE International Conference
  on}, vol.~2, pp.~1208--1213, IEEE, 2005.

\bibitem{demartines1997curvilinear}
P.~Demartines and J.~H{\'e}rault, ``Curvilinear component analysis: A
  self-organizing neural network for nonlinear mapping of data sets,'' {\em
  IEEE Transactions on neural networks}, vol.~8, no.~1, pp.~148--154, 1997.

\bibitem{lee2000robust}
J.~A. Lee, A.~Lendasse, N.~Donckers, M.~Verleysen, {\em et~al.}, ``A robust
  non-linear projection method.,'' in {\em ESAnn}, pp.~13--20, 2000.

\bibitem{lee2001introduction}
J.~M. Lee, {\em Introduction to smooth manifolds}.
\newblock Springer, 2001.

\bibitem{scholkopf1997kernel}
B.~Sch{\"o}lkopf, A.~Smola, and K.-R. M{\"u}ller, ``Kernel principal component
  analysis,'' in {\em International Conference on Artificial Neural Networks},
  pp.~583--588, Springer, 1997.

\bibitem{weinberger2006introduction}
K.~Q. Weinberger and L.~K. Saul, ``An introduction to nonlinear dimensionality
  reduction by maximum variance unfolding,'' in {\em AAAI}, vol.~6,
  pp.~1683--1686, 2006.

\bibitem{coifman2006diffusion}
R.~R. Coifman and S.~Lafon, ``Diffusion maps,'' {\em Applied and computational
  harmonic analysis}, vol.~21, no.~1, pp.~5--30, 2006.

\bibitem{zhang2004principal}
Z.~Zhang and H.~Zha, ``Principal manifolds and nonlinear dimensionality
  reduction via tangent space alignment,'' {\em SIAM journal on scientific
  computing}, vol.~26, no.~1, pp.~313--338, 2004.

\bibitem{pal1998two}
N.~R. Pal and V.~K. Eluri, ``Two efficient connectionist schemes for structure
  preserving dimensionality reduction,'' {\em IEEE Transactions on Neural
  Networks}, vol.~9, no.~6, pp.~1142--1154, 1998.

\bibitem{jain1992artificial}
A.~K. Jain and J.~Mao, ``Artificial neural network for nonlinear projection of
  multivariate data,'' in {\em Neural Networks, 1992. IJCNN., International
  Joint Conference on}, vol.~3, pp.~335--340, IEEE, 1992.

\bibitem{mao1995artificial}
J.~Mao and A.~K. Jain, ``Artificial neural networks for feature extraction and
  multivariate data projection,'' {\em IEEE transactions on neural networks},
  vol.~6, no.~2, pp.~296--317, 1995.

\bibitem{demers1993non}
D.~DeMers and G.~W. Cottrell, ``Non-linear dimensionality reduction,'' in {\em
  Advances in neural information processing systems}, pp.~580--587, 1993.

\bibitem{wang2014generalized}
W.~Wang, Y.~Huang, Y.~Wang, and L.~Wang, ``Generalized autoencoder: A neural
  network framework for dimensionality reduction,'' in {\em Proceedings of the
  IEEE conference on computer vision and pattern recognition workshops},
  pp.~490--497, 2014.

\bibitem{mamdani1975experiment}
E.~H. Mamdani and S.~Assilian, ``An experiment in linguistic synthesis with a
  fuzzy logic controller,'' {\em International journal of man-machine studies},
  vol.~7, no.~1, pp.~1--13, 1975.

\bibitem{qiao2012explicit}
H.~Qiao, P.~Zhang, D.~Wang, and B.~Zhang, ``An explicit nonlinear mapping for
  manifold learning,'' {\em IEEE transactions on cybernetics}, vol.~43, no.~1,
  pp.~51--63, 2012.

\bibitem{yang2004sammon}
L.~Yang, ``Sammon's nonlinear mapping using geodesic distances,'' in {\em
  Pattern Recognition, 2004. ICPR 2004. Proceedings of the 17th International
  Conference on}, vol.~2, pp.~303--306, IEEE, 2004.

\bibitem{sugeno1993fuzzy}
M.~Sugeno and T.~Yasukawa, ``A fuzzy-logic-based approach to qualitative
  modeling,'' {\em IEEE Transactions on fuzzy systems}, vol.~1, no.~1, p.~7,
  1993.

\bibitem{wang1992generating}
L.-X. Wang and J.~M. Mendel, ``Generating fuzzy rules by learning from
  examples,'' {\em IEEE Transactions on systems, man, and cybernetics},
  vol.~22, no.~6, pp.~1414--1427, 1992.

\bibitem{cordon1999two}
O.~Cord{\'o}n and F.~Herrera, ``A two-stage evolutionary process for designing
  tsk fuzzy rule-based systems,'' {\em IEEE Transactions on Systems, Man, and
  Cybernetics, Part B (Cybernetics)}, vol.~29, no.~6, pp.~703--715, 1999.

\bibitem{mitra2000neuro}
S.~Mitra and Y.~Hayashi, ``Neuro-fuzzy rule generation: survey in soft
  computing framework,'' {\em IEEE transactions on neural networks}, vol.~11,
  no.~3, pp.~748--768, 2000.

\bibitem{pal2008simultaneous}
N.~R. Pal and S.~Saha, ``Simultaneous structure identification and fuzzy rule
  generation for takagi--sugeno models,'' {\em IEEE Transactions on Systems,
  Man, and Cybernetics, Part B (Cybernetics)}, vol.~38, no.~6, pp.~1626--1638,
  2008.

\bibitem{chen2012integrated}
Y.-C. Chen, N.~R. Pal, and I.-F. Chung, ``An integrated mechanism for feature
  selection and fuzzy rule extraction for classification,'' {\em IEEE
  Transactions on Fuzzy Systems}, vol.~20, no.~4, pp.~683--698, 2012.

\bibitem{asgharbeygi2008geodesic}
N.~Asgharbeygi and A.~Maleki, ``Geodesic k-means clustering,'' in {\em 2008
  19th International Conference on Pattern Recognition}, pp.~1--4, IEEE, 2008.

\bibitem{kim2007soft}
J.~Kim, K.-H. Shim, and S.~Choi, ``Soft geodesic kernel k-means,'' in {\em 2007
  IEEE International Conference on Acoustics, Speech and Signal
  Processing-ICASSP'07}, vol.~2, pp.~II--429, IEEE, 2007.

\bibitem{feil2007geodesic}
B.~Feil and J.~Abonyi, ``Geodesic distance based fuzzy clustering,'' in {\em
  Soft Computing in Industrial Applications}, pp.~50--59, Springer, 2007.

\bibitem{floyd1962algorithm}
R.~W. Floyd, ``Algorithm 97: shortest path,'' {\em Communications of the ACM},
  vol.~5, no.~6, p.~345, 1962.

\bibitem{pal2002complexity}
N.~R. Pal and J.~C. Bezdek, ``Complexity reduction for ``large image"
  processing,'' {\em IEEE Transactions on Systems, Man, and Cybernetics, Part B
  (Cybernetics)}, vol.~32, no.~5, pp.~598--611, 2002.

\bibitem{scikit-learn}
F.~Pedregosa, G.~Varoquaux, A.~Gramfort, V.~Michel, B.~Thirion, O.~Grisel,
  M.~Blondel, P.~Prettenhofer, R.~Weiss, V.~Dubourg, J.~Vanderplas, A.~Passos,
  D.~Cournapeau, M.~Brucher, M.~Perrot, and E.~Duchesnay, ``Scikit-learn:
  Machine learning in {P}ython,'' {\em Journal of Machine Learning Research},
  vol.~12, pp.~2825--2830, 2011.

\bibitem{frey_rawface}
S.~Roweis, ``Data for matlab hackers, faces, frey face,'' 2020.

\bibitem{nene1996columbia}
S.~A. Nene, S.~K. Nayar, H.~Murase, {\em et~al.}, ``Columbia object image
  library (coil-20),''

\bibitem{abadi2016TensorFlow}
M.~Abadi, P.~Barham, J.~Chen, Z.~Chen, A.~Davis, J.~Dean, M.~Devin,
  S.~Ghemawat, G.~Irving, M.~Isard, M.~Kudlur, J.~Levenberg, R.~Monga,
  S.~Moore, D.~G. Murray, B.~Steiner, P.~Tucker, V.~Vasudevan, P.~Warden,
  M.~Wicke, Y.~Yu, and X.~Zheng, ``Tensorflow: A system for large-scale machine
  learning,'' in {\em 12th USENIX Symposium on Operating Systems Design and
  Implementation (OSDI 16)}, pp.~265--283, 2016.

\bibitem{drtoolbox}
L.~van~der Maaten, ``Matlab toolbox for dimensionality reduction,'' 2020.

\end{thebibliography}
\makeatletter\@input{other.tex}\makeatother
\end{document}


\title{
		~\\\vspace{60 mm}
		\textsc{Supplementary Materials} \\ \vspace{30 mm}   Nonlinear Dimensionality Reduction for Data Visualization: An Unsupervised Fuzzy Rule Based Approach
	}
	\author{Suchismita~Das
		~and~Nikhil~R.~Pal,~\IEEEmembership{Fellow,~IEEE}}
	\maketitle
	\IEEEpeerreviewmaketitle

\newpage

\begin{algorithm*}[t] \SetAlgoRefName{S-1}
	\DontPrintSemicolon
	\SetAlgoLined
	\caption{Geodesic $c$-means algorithm (GCM)}\label{alg:GeoCMeans}
	Determine Geodesic distances $\mathbf{gd}$ between pairs of points in $\mathbf{X}$. \;
	Let the number of clusters $=n_c$. Initialize, cluster centroids $\left\lbrace \mathbf{v}_{1},\mathbf{v}_{2},\cdots, \mathbf{v}_{n_c}\right\rbrace $ with $n_c$ random input points i.e, $n_c$ random points from $\mathbf{X}=\left\lbrace \mathbf{x}_{1},\mathbf{x}_{2},\cdots,\mathbf{x}_{n} \right\rbrace $\;
	Initialize sets $S_{k}=\phi; \hspace{4pt} k=1,2,\cdots,n_c$.\;
	\While {$iteration < Maximum Iteration$} {
		\For {$i=1 \text{ to } n$}{
			$k^{*}=\arg_{k} \min  \hspace{4pt} \mathbf{gd}(\mathbf{x}_{i},\mathbf{v}_{k})$\\
			$S_{k^{*}}=S_{k^{*}} \cup \mathbf{x}_{i}$
		}	
		\For {$k=1 \text{ to } n_{c}$}{
			$\hat{\mathbf{v}}_{k}=\nicefrac{1}{|S_{k}|}\sum_{\mathbf{x}_{i}\in S_{k}}  \mathbf{x}_{i}$\;
			$i^{*}=\arg_{i} \min \hspace{4pt} \mathbf{ed}(\hat{\mathbf{v}}_{k},\mathbf{x}_{i}) $\;
			$\mathbf{v}_{k}=\mathbf{x}_{i^{*}}$\;
		}
	}
	\Return $\left\lbrace \mathbf{v}_{1},\mathbf{v}_{2},\cdots, \mathbf{v}_{n_c}\right\rbrace $ \;
	
\end{algorithm*}

\newpage
\section{Experiment to Choose Number of Rules and Initial Spread \label{sec:number_of_rules}}
\begin{figure*}[!hb]
	\centering
	\includegraphics[width=.55\linewidth]{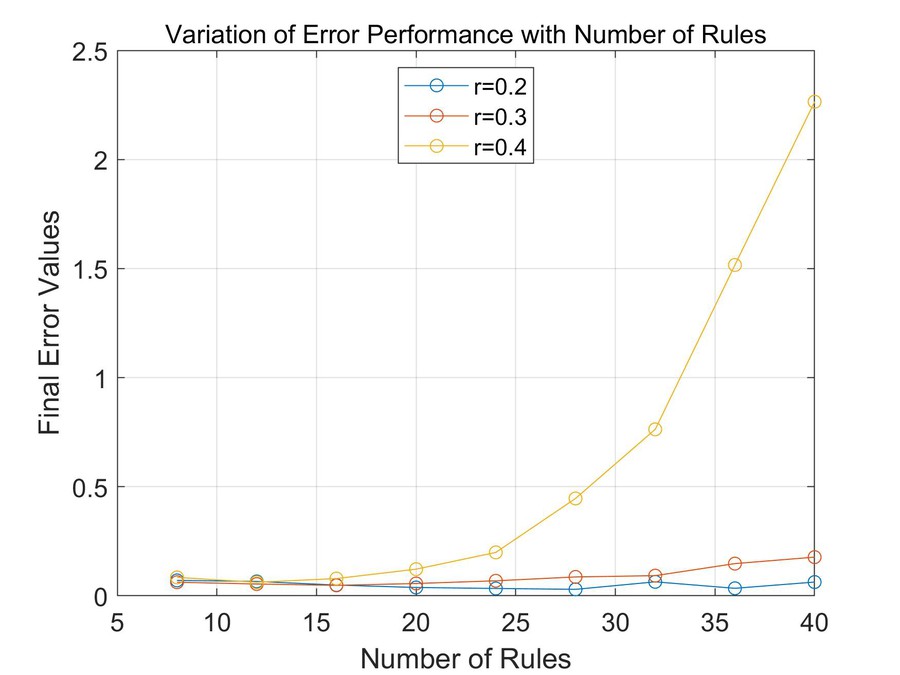}
	\caption{Variation of average final error values with number of rules and initial spreads.}
	\label{fig:nc_selection}
\end{figure*}

We run our method on the Swiss Roll dataset with number of rules$(n_c)= 8, 12, 16$ and so on till $40$ and repeat this experiment for three different spread initialization of the Gaussian membership functions: $\sigma=r*$feature-specific range, for $r=0.2, 0.3,$ and $0.4$. For each combination of $n_{c}$ and $\sigma$ five runs are performed and the average of the final error values (Eqn. (\ref{eq: error})) over the five runs is computed. Fig. \ref{fig:nc_selection} shows the plot of average final error values vs. the number of rules. We observe that for all the three spread initializations, when the number of rules=20, we get an acceptable low average error value. But again we should consider that to model a data set with a higher number of points satisfactorily, more rules may be necessary. In case of the Swiss Roll dataset, the number of input points is $2000$. Since $20$ is $1\%$ of $2000$, $n_{c}$ is set as exactly or nearly $1\%$ of the number of input instances. With $r=0.2$ or $0.3$ low average final error values are obtained for all the values of $n_{c}$. So, along with Swiss Roll for all the three synthetic data sets we conduct experiments with these two choices of $\sigma$. However, for real data sets we have to choose higher values of $\sigma$ to avoid many cases with zero rule-firing strength.   

	\begin{figure*}[!tb]
			\centering
			\begin{subfigure}{.33\textwidth}
				\centering
				\includegraphics[width=.98\linewidth]{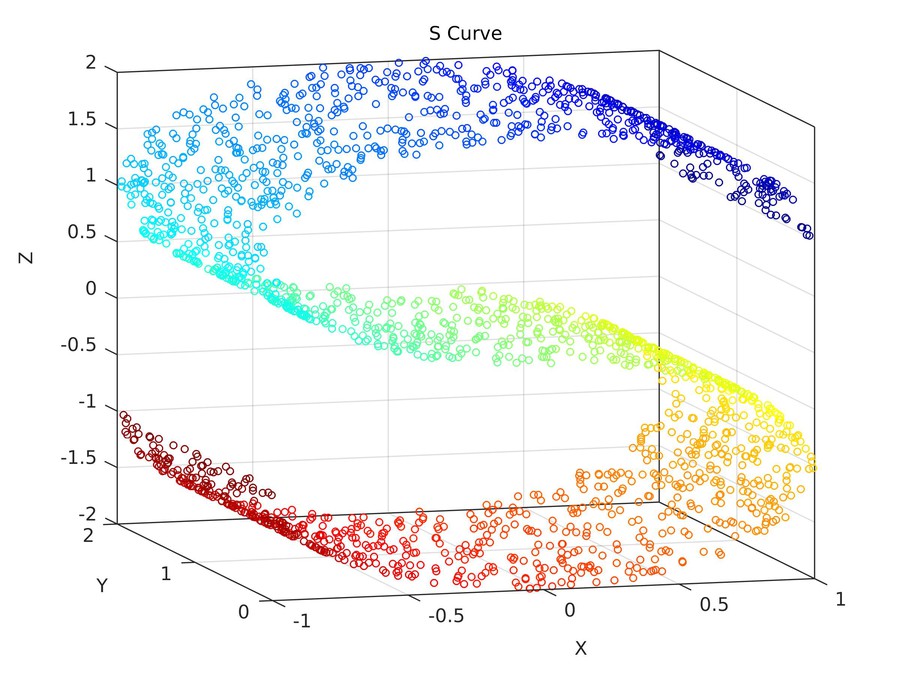}%
				\caption{Original Data}
				\label{fig:s curve}
			\end{subfigure}%
			\begin{subfigure}{.33\textwidth}
				\centering
				\includegraphics[width=.98\linewidth]{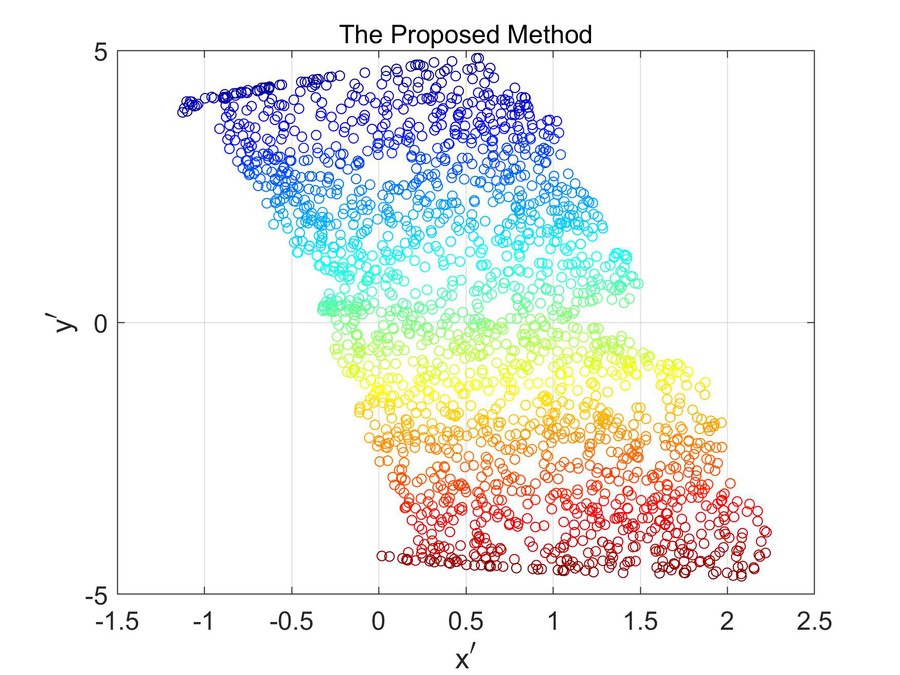}
				\caption{Proposed Method}
				\label{fig:s curve proposed}
			\end{subfigure}%
			\begin{subfigure}{.33\textwidth}
				\centering
				\includegraphics[width=.98\linewidth]{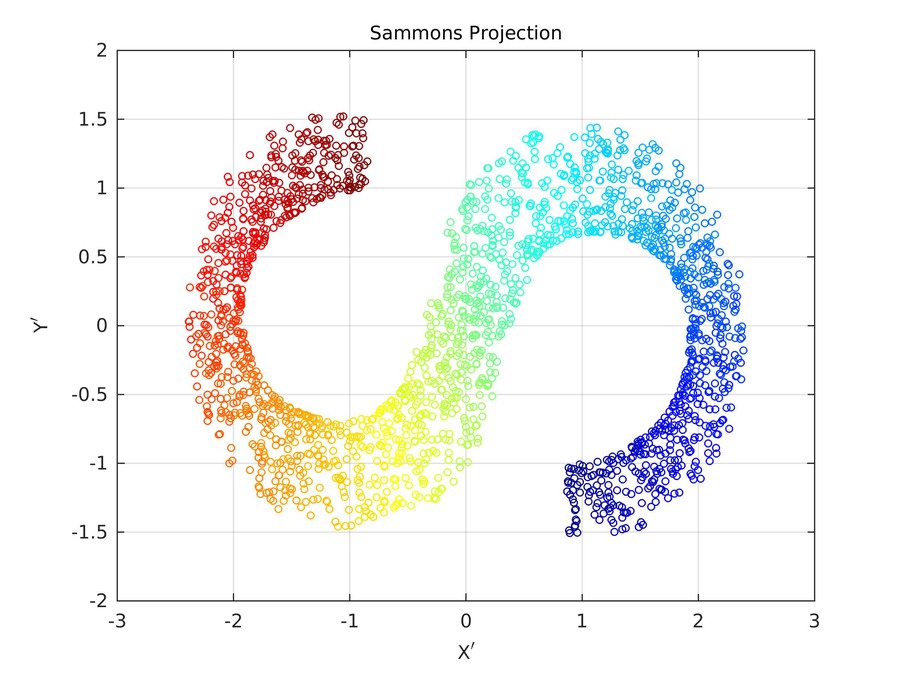}
				\caption{Sammon's Projection}
				\label{fig:s curve sammons}
			\end{subfigure}%
			\\
			\begin{subfigure}{.33\textwidth}
				\centering
				\includegraphics[width=.98\linewidth]{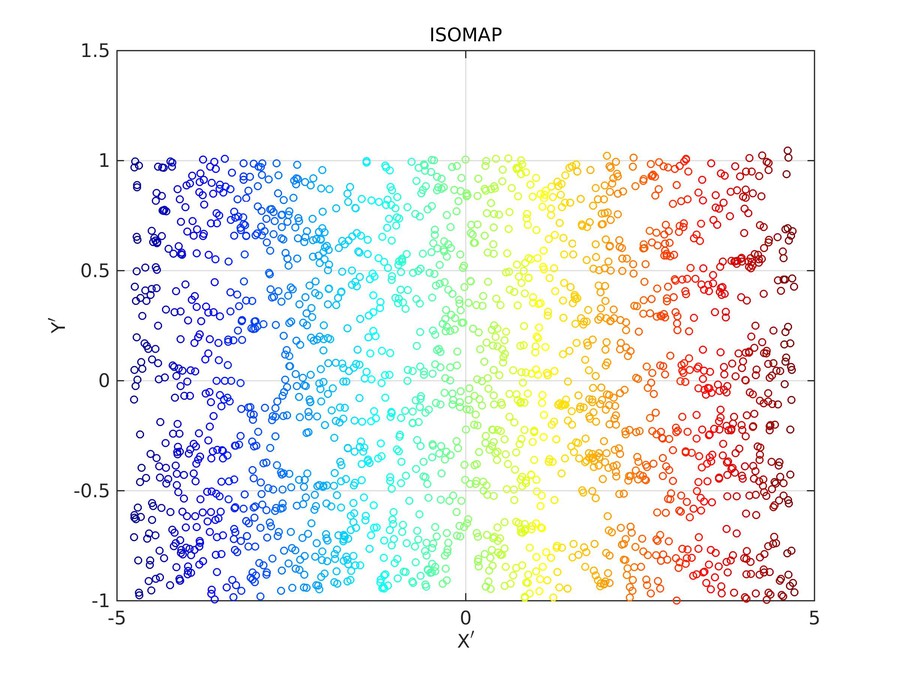}
				\caption{ISOMAP}
				\label{fig:s curve ISOMAP}
			\end{subfigure}
			\begin{subfigure}{.33\textwidth}
				\centering
				\includegraphics[width=.98\linewidth]{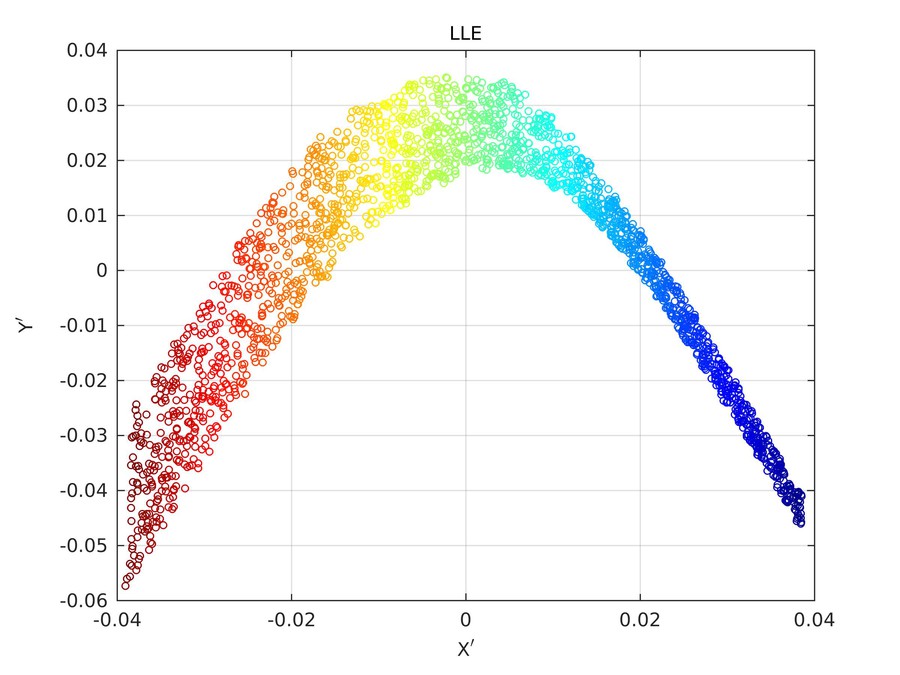}
				\caption{LLE}
				\label{fig:s curve LLE}
			\end{subfigure}%
			\begin{subfigure}{.33\textwidth}
				\centering
				\includegraphics[width=.98\linewidth]{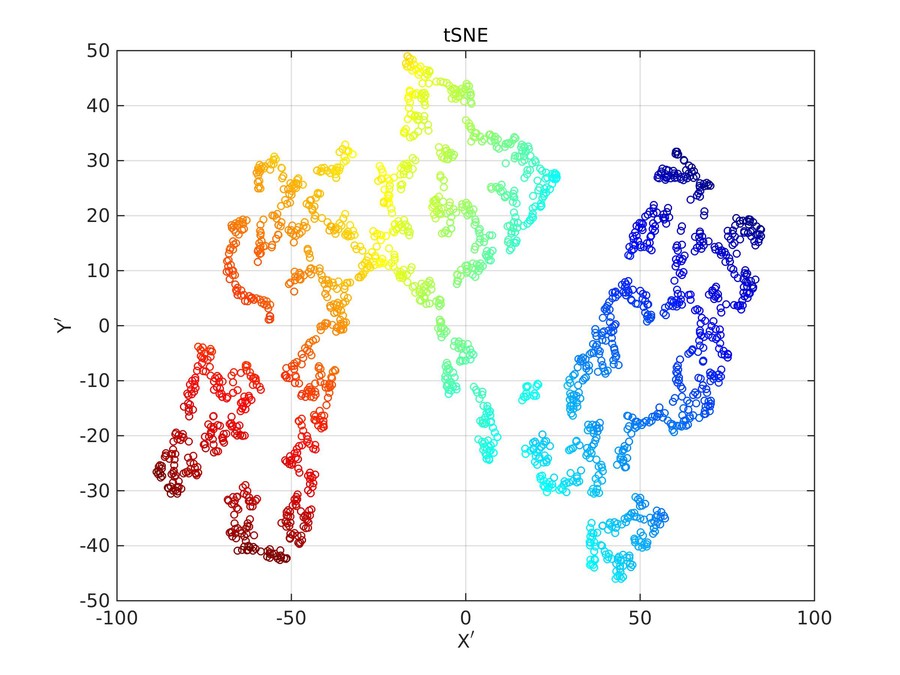}
				\caption{$t$-SNE }
				\label{fig:s curve tSNE}
			\end{subfigure}
			\label{fig:s curve results}
			\caption{Visualization of the S Curve data (a) original input space, (b) proposed method, (c) Sammon's projection, (d)ISOMAP, (e) LLE, and (f) $t$-SNE.}
		\end{figure*}

	\begin{figure*}[!tb]
		\centering
		\includegraphics[width=.98\linewidth]{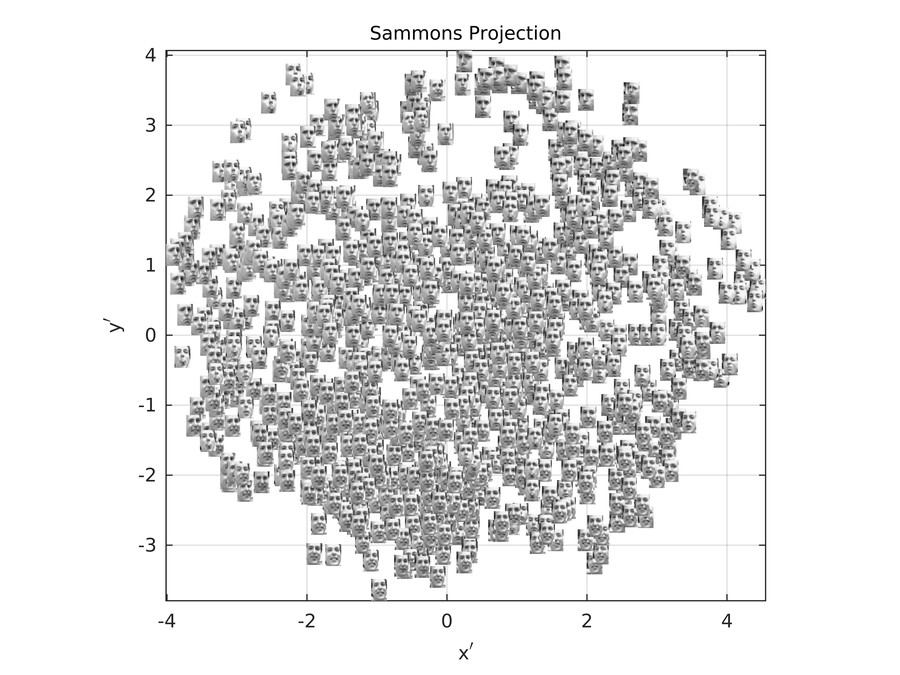}
		\caption{Visualization of the Frey-face data set with Sammon's Projection. }
		\label{fig:frey face sammons}
	\end{figure*}
	\begin{figure*}[!tb]
		\centering
		\includegraphics[width=.98\linewidth]{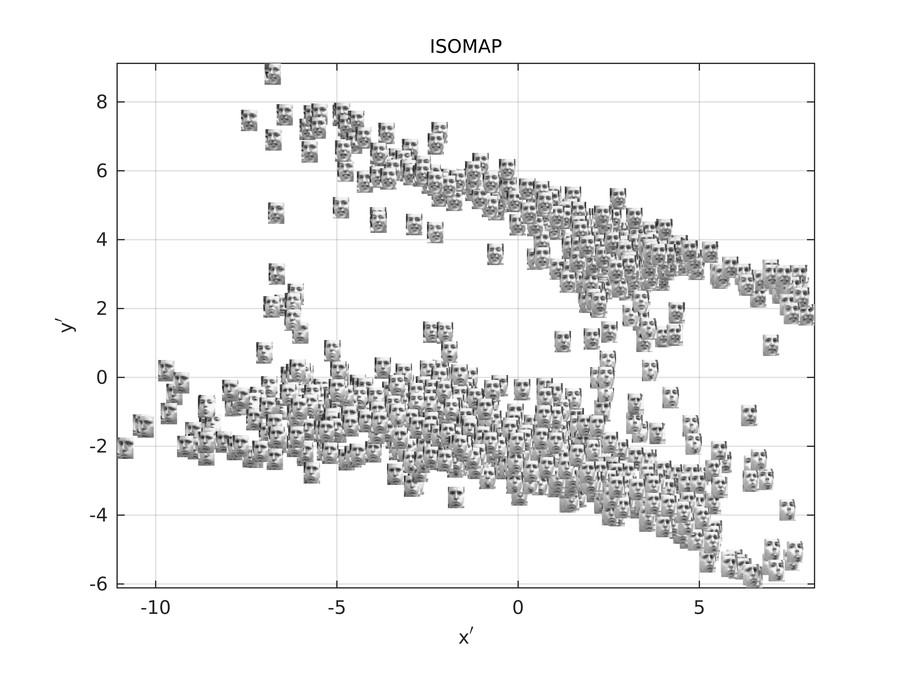}
		\caption{Visualization of the Frey-face data set with ISOMAP}
		\label{fig:frey face ISOMAP}
	\end{figure*}
	\begin{figure*}[!tb]
		\centering
		\includegraphics[width=.98\linewidth]{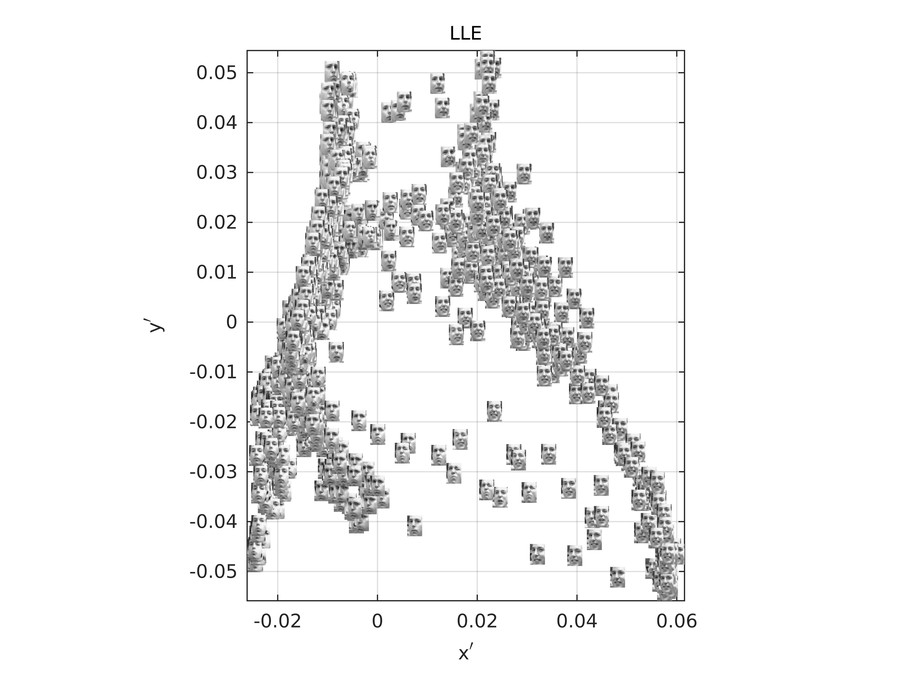}
		\caption{Visualization of the Frey-face data set with LLE.}
		\label{fig:frey face LLE}
	\end{figure*}
	\begin{figure*}[!tb]
		\centering
		\includegraphics[width=.98\linewidth]{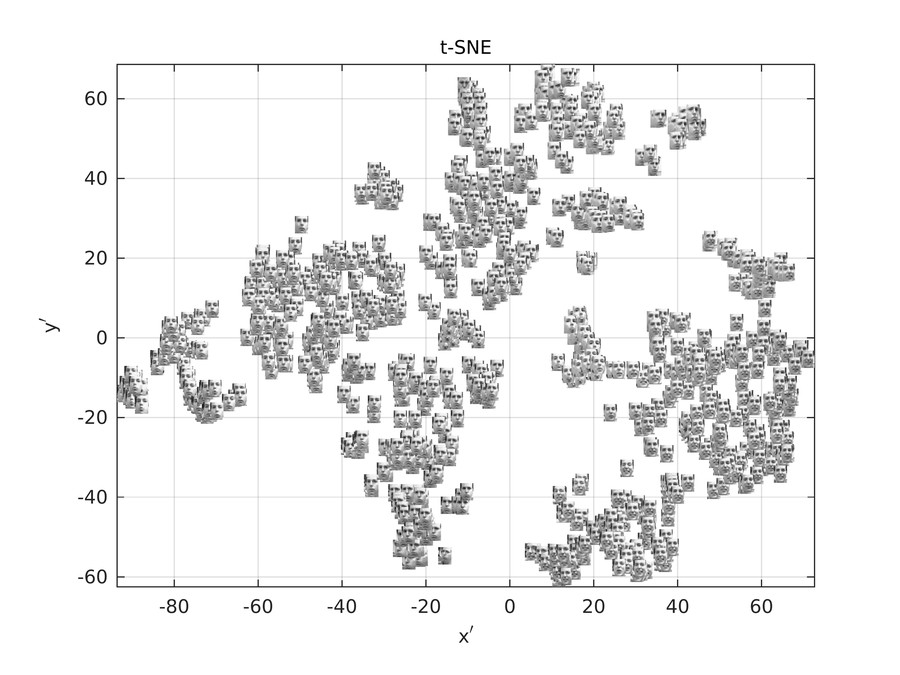}
		\caption{Visualization of the Frey-face data set with $t$-SNE.}
		\label{fig:frey face t-SNE}
	\end{figure*}
	
	\begin{figure*}[!tb]
		\centering
		\begin{subfigure}{.33\textwidth}
			\centering
			\includegraphics[width=.98\linewidth]{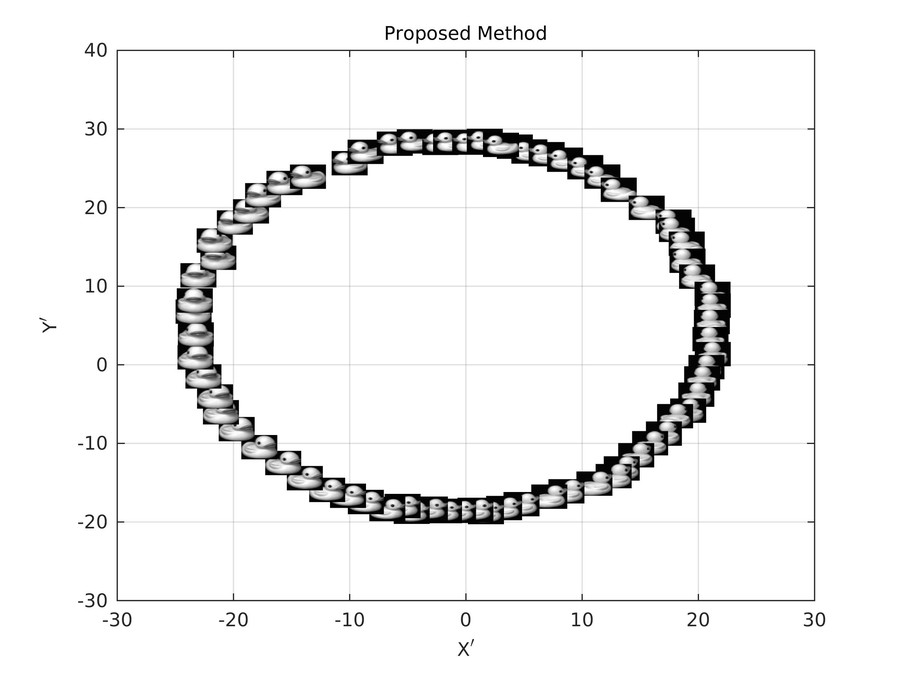}%
			\caption{Proposed Method}
			\label{fig:coil object1 proposed}
		\end{subfigure}
		\begin{subfigure}{.33\textwidth}
			\centering
			\includegraphics[width=.98\linewidth]{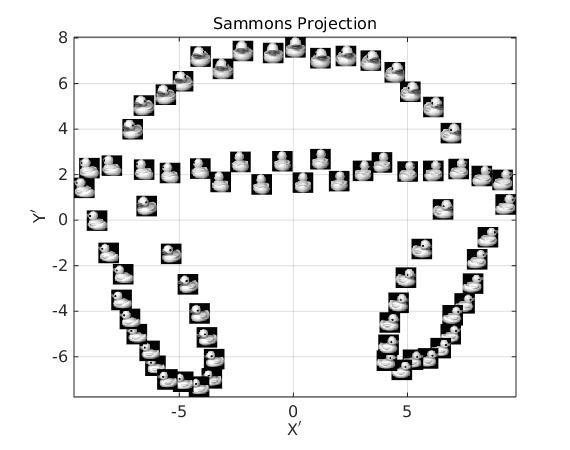}
			\caption{Sammon's Projection}
			\label{fig:coil object1 sammons}
		\end{subfigure}%
		\begin{subfigure}{.33\textwidth}
			\centering
			\includegraphics[width=.98\linewidth]{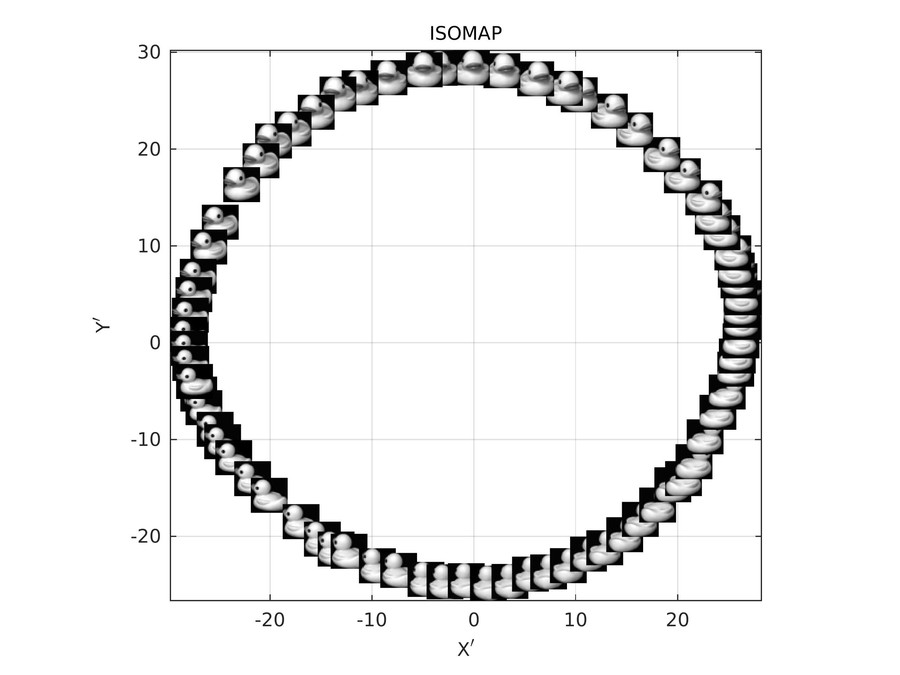}
			\caption{ISOMAP}
			\label{fig:coil object1 ISOMAP}
		\end{subfigure}
		\\
		\begin{subfigure}{.33\textwidth}
			\centering
			\includegraphics[width=.98\linewidth]{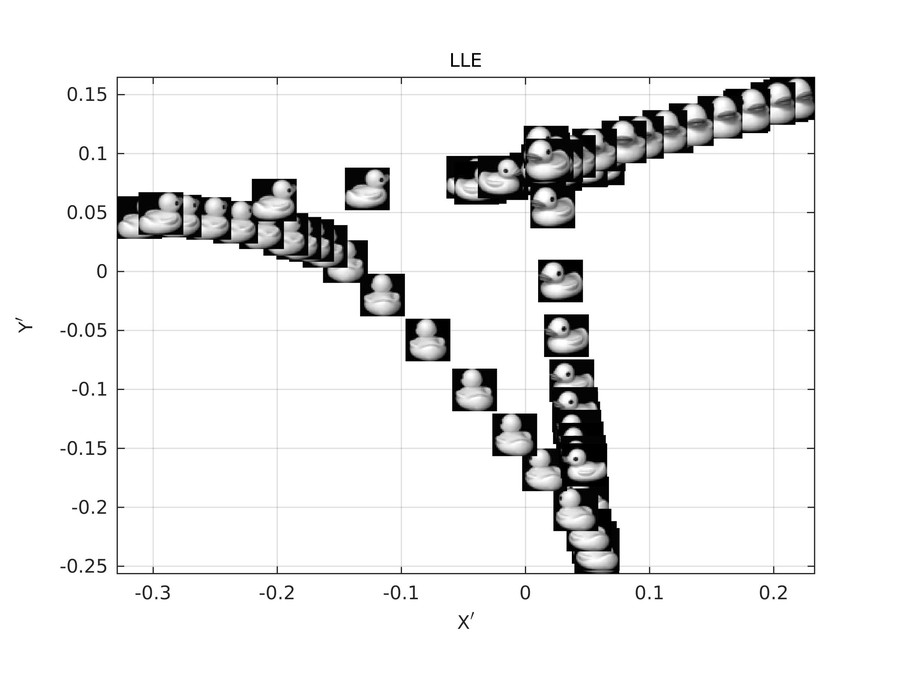}
			\caption{LLE}
			\label{fig:coil object1 LLE}
		\end{subfigure}%
		\begin{subfigure}{.33\textwidth}
			\centering
			\includegraphics[width=.98\linewidth]{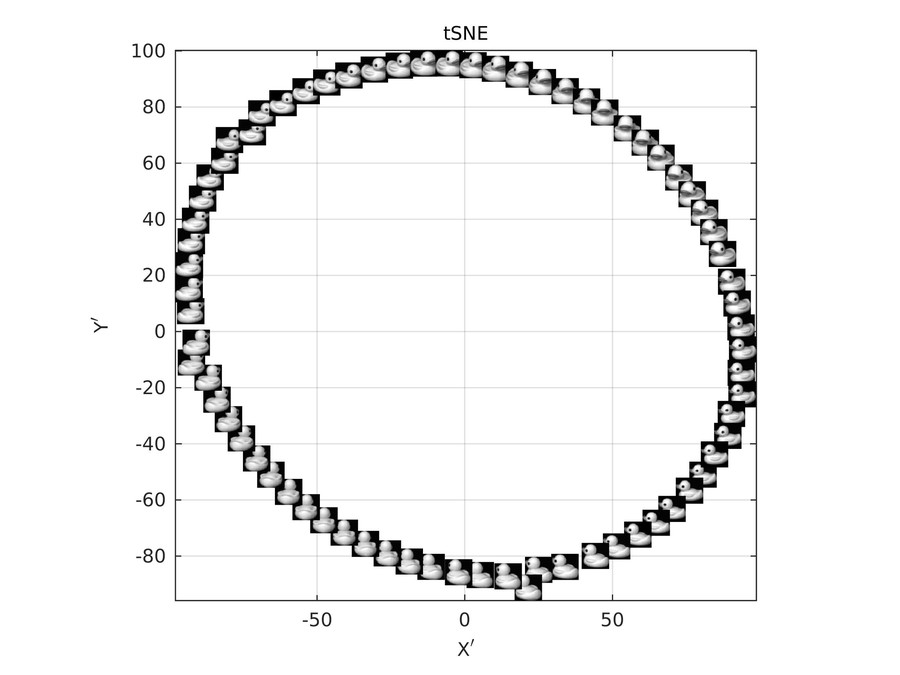}
			\caption{$t$-SNE}
			\label{fig:coil object1 tSNE}
		\end{subfigure}
		
		\caption{Visualization of the first object of the COIL Data Set with (a) proposed method, (b) Sammon's projection, (c)ISOMAP, (d) LLE, and (e) $t$-SNE.}
		\label{fig: coil object1 results}
	\end{figure*}

	\begin{figure*}[!tb]
		\centering
		\centering
		\includegraphics[width=.75\linewidth]{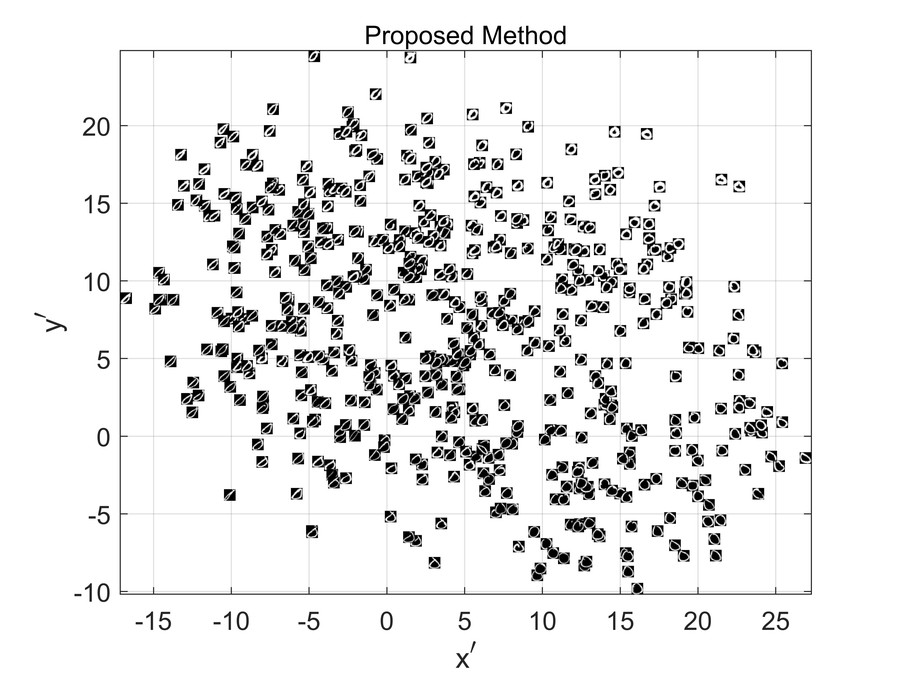}
		\caption{Visualization of the USPS data set with the proposed method.}
		\label{fig:usps_class_0_proposed}
	\end{figure*}
    
    	\begin{figure*}[!tb]
    	\centering
    	\includegraphics[width=.98\linewidth]{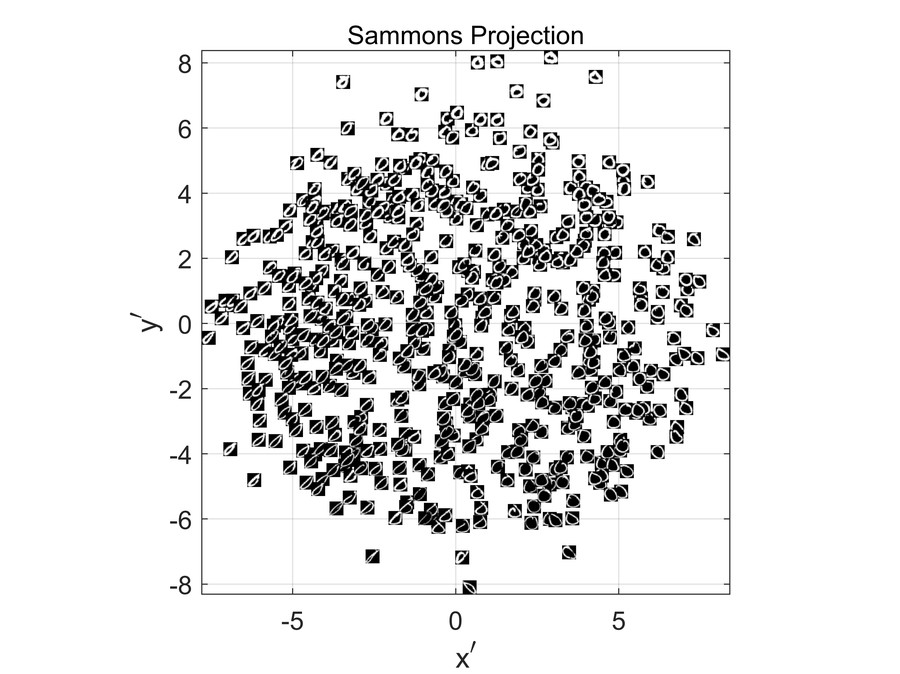}
    	\caption{Visualization of the USPS data set with Sammon's Projection. }
    	\label{fig:usps_sammons}
    \end{figure*}
    \begin{figure*}[!tb]
    	\centering
    	\includegraphics[width=.98\linewidth]{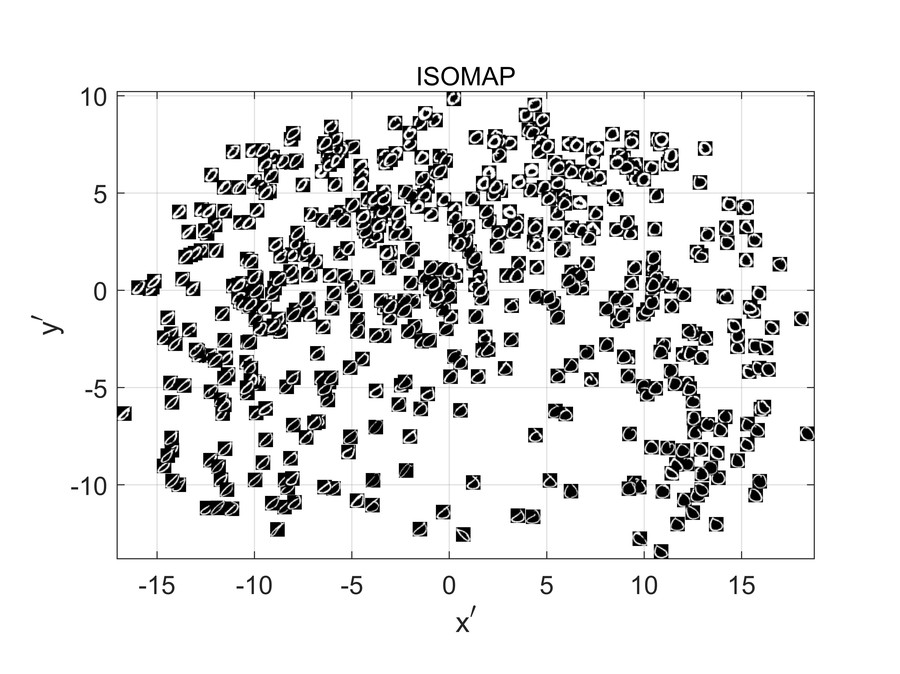}
    	\caption{Visualization of the USPS data set with ISOMAP}
    	\label{fig:usps_ISOMAP}
    \end{figure*}
    \begin{figure*}[!tb]
    	\centering
    	\includegraphics[width=.98\linewidth]{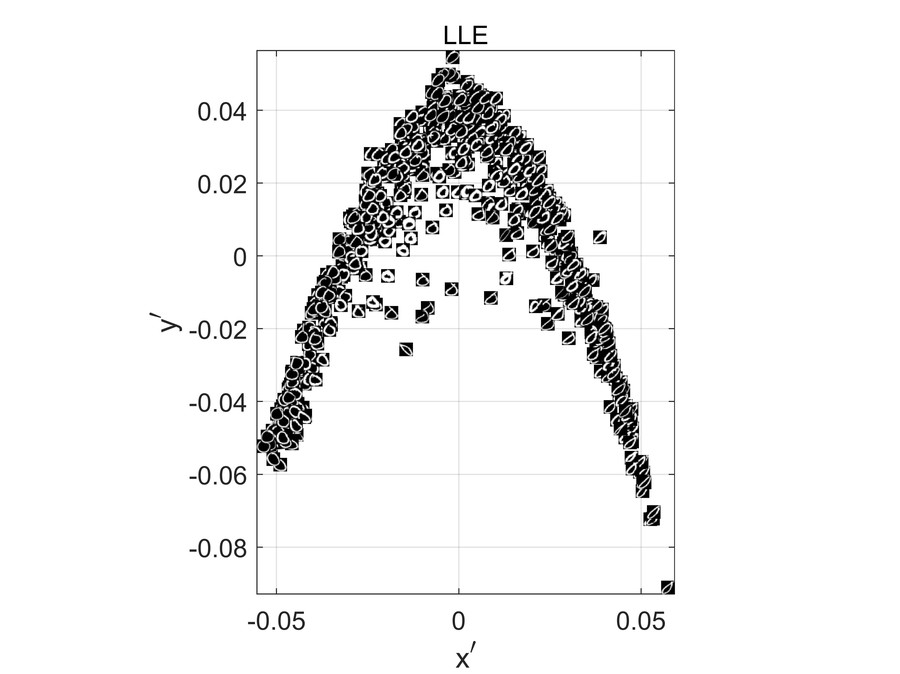}
    	\caption{Visualization of the USPS data set with LLE.}
    	\label{fig:usps_LLE}
    \end{figure*}
    \begin{figure*}[!tb]
    	\centering
    	\includegraphics[width=.98\linewidth]{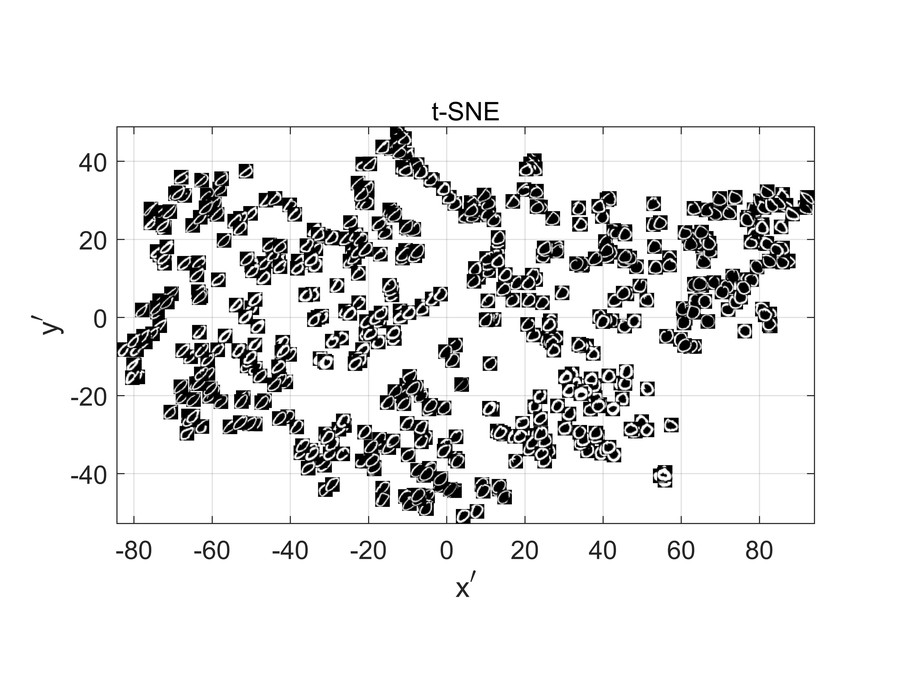}
    	\caption{Visualization of the USPS data set with $t$-SNE.}
    	\label{fig:usps_t-SNE}
    \end{figure*}
   	
   	\begin{figure*}[!tb]
   	\centering
   	\begin{subfigure}{.48\textwidth}
   		\centering
   		\includegraphics[width=.98\linewidth]{raw_swiss_roll2000_proposed_neighbors_20}%
   		\caption{ Rule antecedents defined by cluster centers}
   		\label{fig:swiss_roll_cluster_center_antecedent}
   	\end{subfigure}%
   	\begin{subfigure}{.48\textwidth}
   		\centering
   		\includegraphics[width=.98\linewidth]{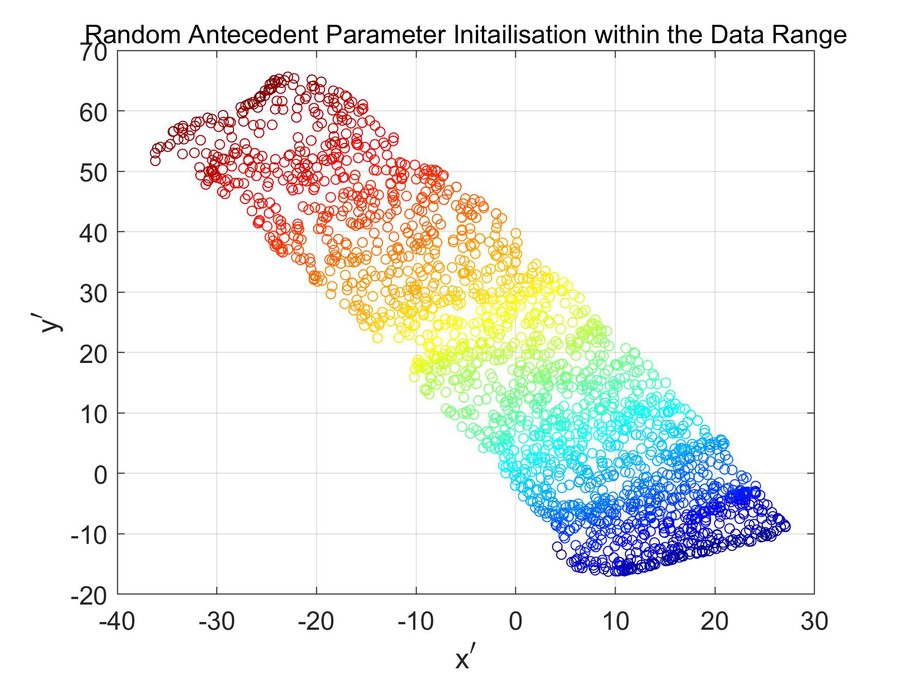}%
   		\caption{Antecedents defined by random values from the smallest hyper-box containing the training data}
   		\label{fig:swiss_roll_random_antecedent}
   	\end{subfigure}%
   	\caption{Visualization of the Swiss Roll data using proposed method with rule antecedents defined by (a) cluster centroids, (b)random values from the smallest hyperbox containing the training data. }
   	\label{fig:swiss_roll_impact_of_antecedent_initialization}
    \end{figure*}
\begin{figure*}[!thb]
	\centering
	\includegraphics[width=.95\linewidth, height=0.6\linewidth]{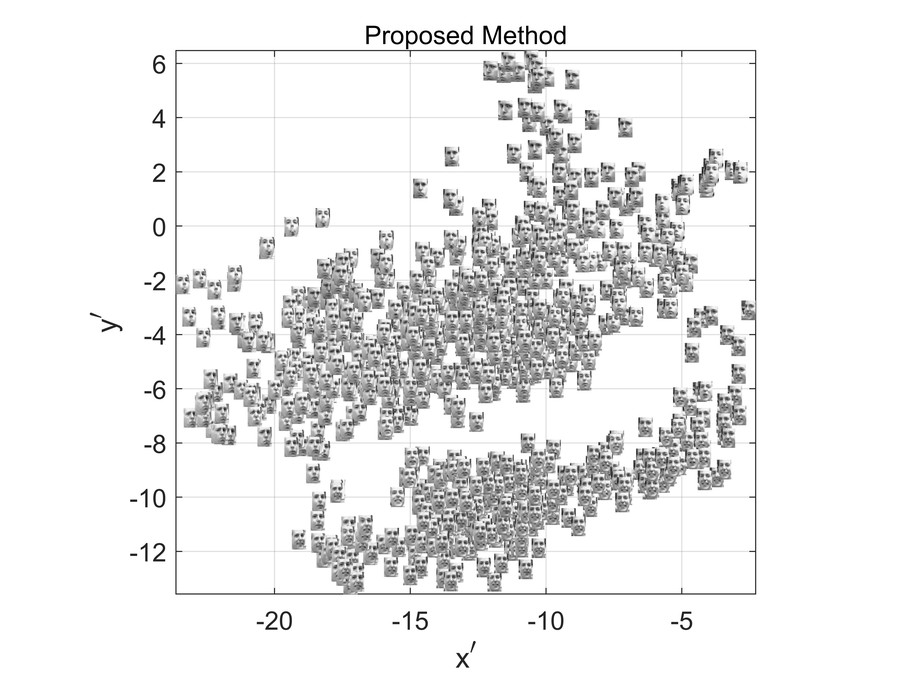}
	\caption{Visualization of the Frey face data set using the proposed method with random rule antecedents.}
	\label{fig:frey_face_random_antecedent}
\end{figure*}

	\begin{figure*}[!tb]
		\centering
		\begin{subfigure}{.33\textwidth}
			\centering
			\includegraphics[width=.98\linewidth]{raw_swiss_roll_2000_ver1}%
			\caption{Original data}
			\label{fig:swiss_roll_sup}
		\end{subfigure}
		\begin{subfigure}{.33\textwidth}
			\centering
			\includegraphics[width=.98\linewidth]{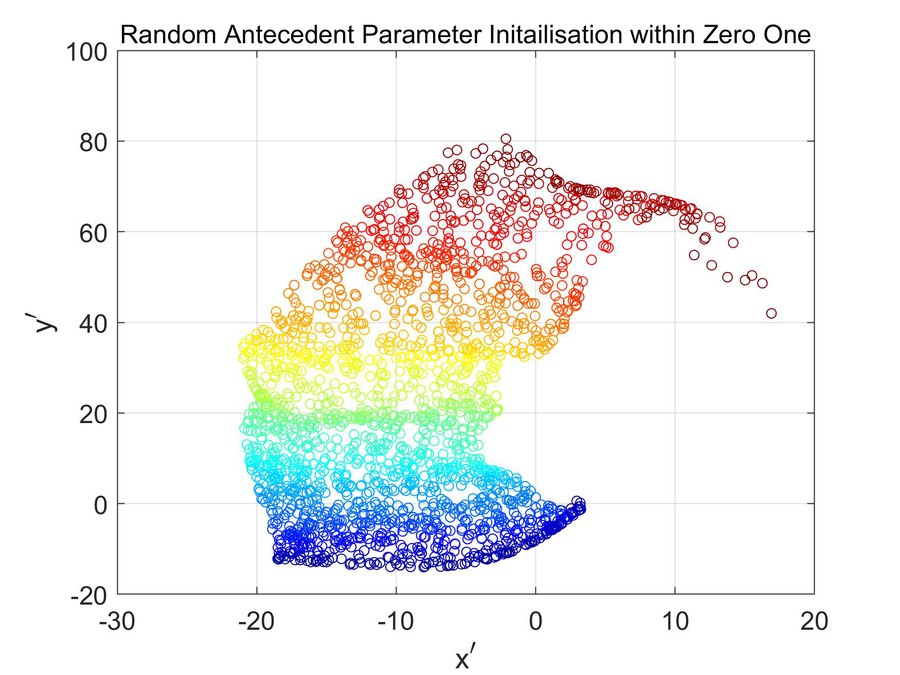}
			\caption{Antecedents defined by random values within zero and one}
			\label{fig:swiss_roll_initialization_random_zero_one_best}
		\end{subfigure}%
		\begin{subfigure}{.33\textwidth}
			\centering
			\includegraphics[width=.98\linewidth]{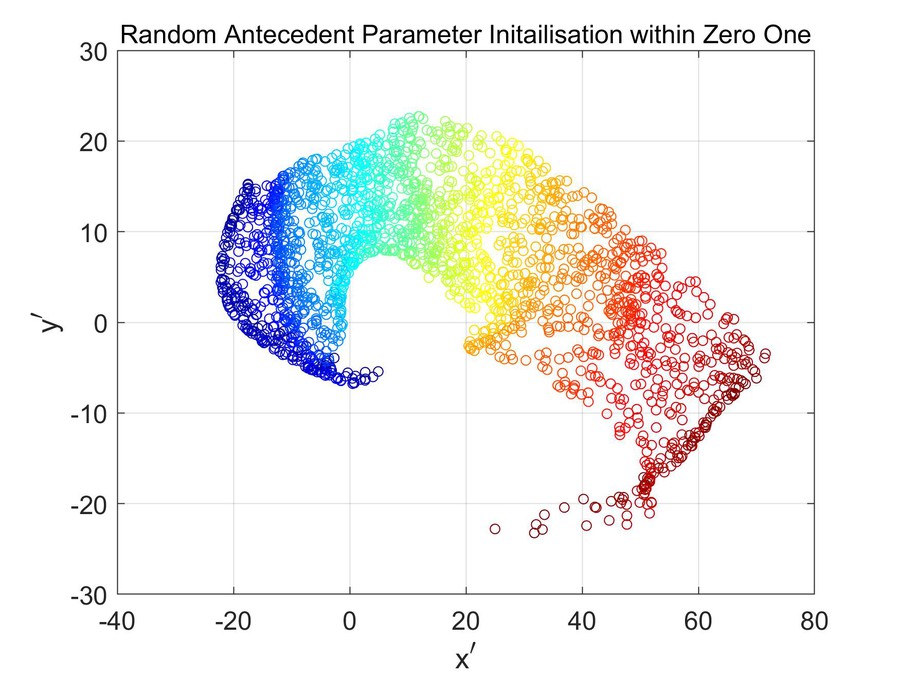}
			\caption{Antecedents defined by random values within zero and one}
			\label{fig:swiss_roll_initialization_random_zero_one_worst}
		\end{subfigure}
		\caption{Visualization of the Swiss Roll data set (a) in original space; by proposed method with rule antecedents defined by random values within zero and one (b) best run (c) worst run.}
		\label{fig:swiss_roll_initialization_random_zero_one}
	\end{figure*}
	\begin{figure*}[!thb]
		\centering
		\begin{subfigure}{.48\textwidth}
			\centering
			\includegraphics[width=.98\linewidth]{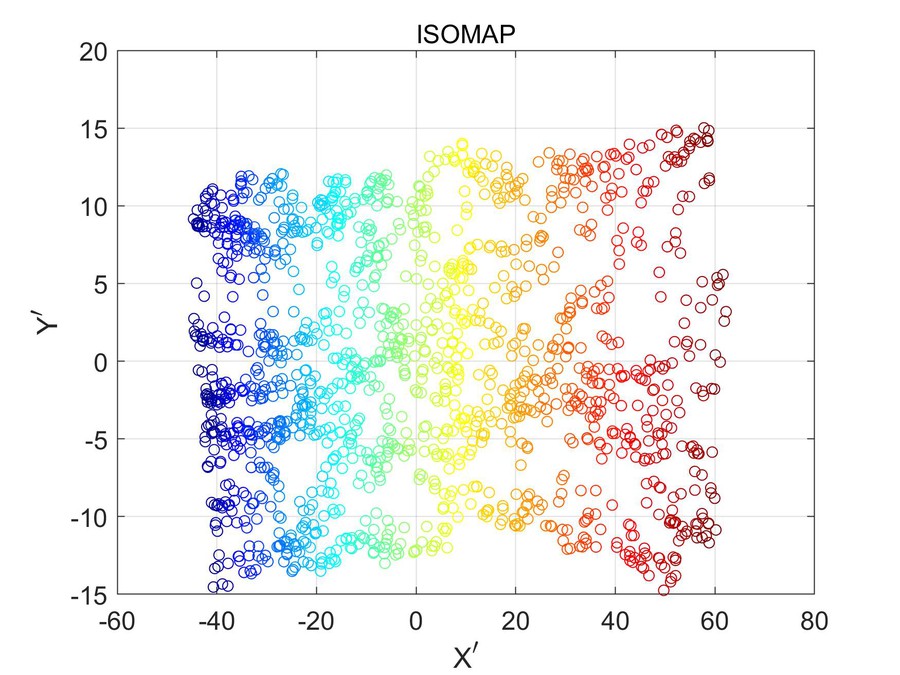}
			\caption{ ISOMAP output for Training Set}
			\label{fig:ISOMAP_swiss_roll_tr1}
		\end{subfigure}%
		\begin{subfigure}{.48\textwidth}
			\centering
			\includegraphics[width=.98\linewidth]{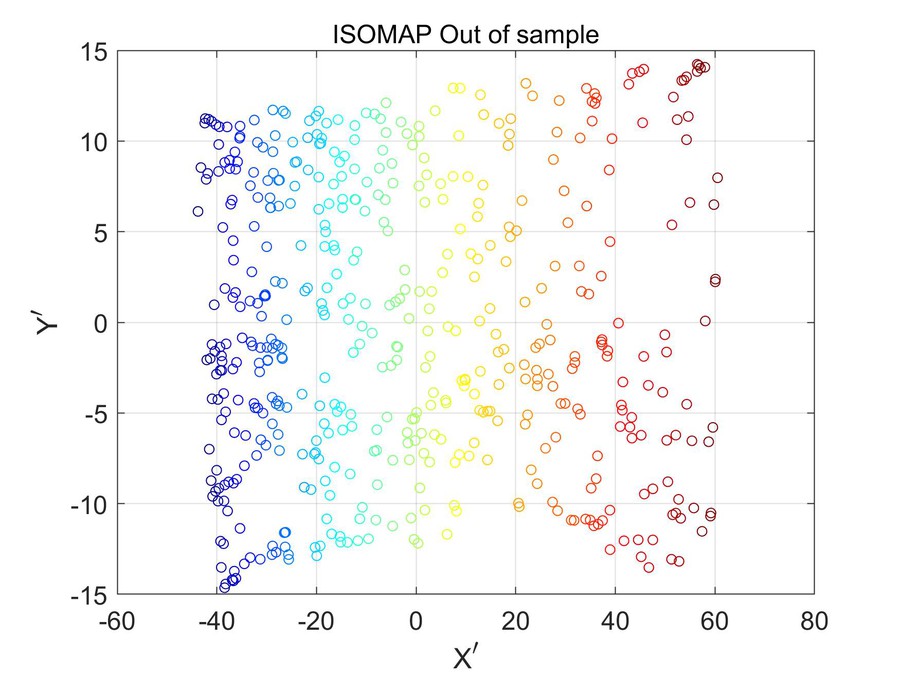}
			\caption{Out of sample extension of ISOMAP for Test Set}
			\label{fig:ISOMAP_swiss_roll_ts1}
		\end{subfigure}%
		\caption{For experiment 1 on validation of predictability with the Swiss Roll data : (a) ISOMAP output for Training Set,  and (b) Out of sample extension of ISOMAP for Test Set.}
		\label{fig:ISOMAP_swiss_roll_out_of_sample}
	\end{figure*}
	\begin{figure*}[!thb]
		\centering
		\begin{subfigure}{.48\textwidth}
			\centering
			\includegraphics[width=.98\linewidth]{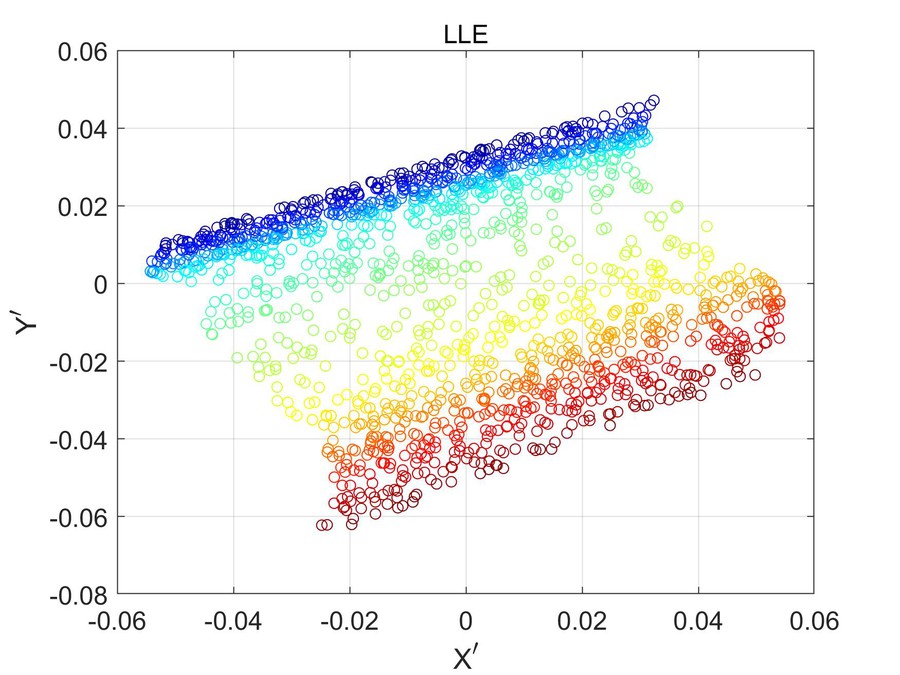}
			\caption{ LLE output for Training Set}
			\label{fig:LLE_swiss_roll_tr1}
		\end{subfigure}%
		\begin{subfigure}{.48\textwidth}
			\centering
			\includegraphics[width=.98\linewidth]{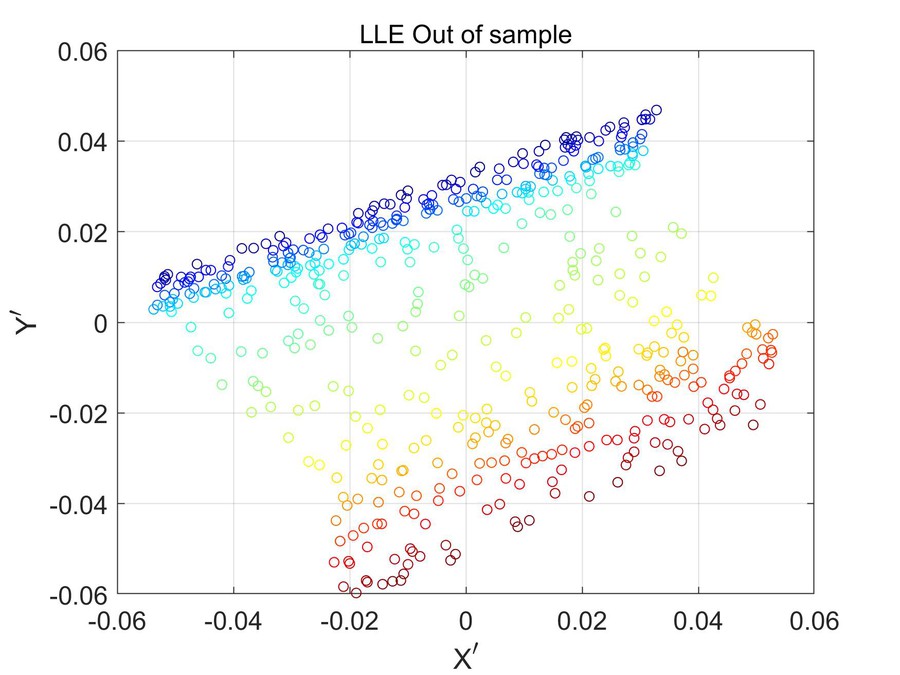}
			\caption{Out of sample extension of LLE for Test Set}
			\label{fig:LLE_swiss_roll_ts1}
		\end{subfigure}%
		\caption{For experiment 1 on validation of predictability with the Swiss Roll data : (a) LLE output for Training Set,  and (b) Out of sample extension of LLE for Test Set.}
		\label{fig:LLE_swiss_roll_out_of_sample}
	\end{figure*}

	\begin{figure*}[!thb]
		\centering
		\begin{subfigure}{.48\textwidth}
			\centering
			\includegraphics[width=.98\linewidth]{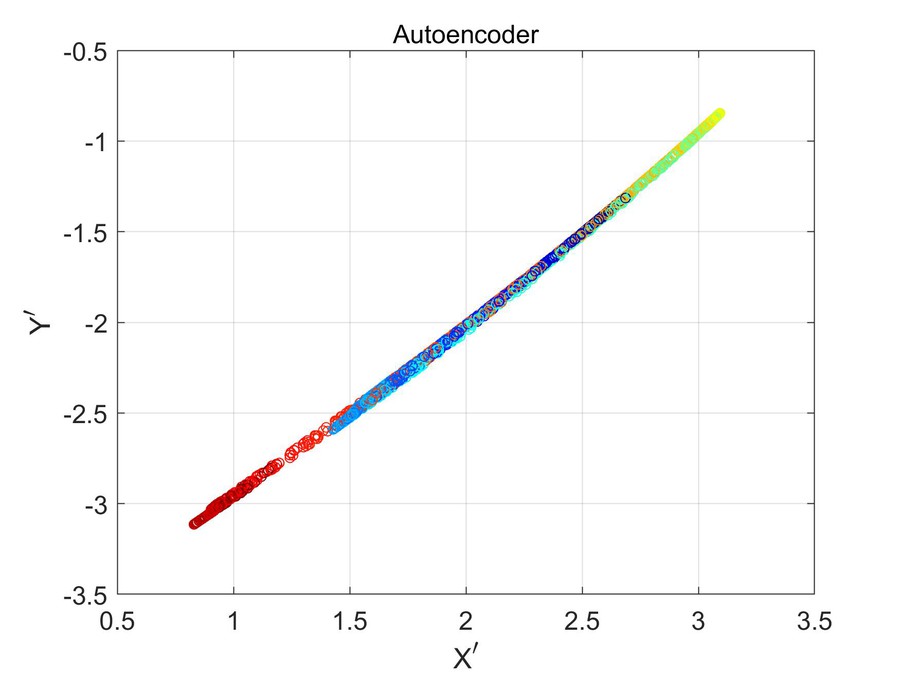}
			\caption{ Autoencoder encoded output for Training Set}
			\label{fig:autoencoder_swiss_roll_tr1}
		\end{subfigure}%
		\begin{subfigure}{.48\textwidth}
			\centering
			\includegraphics[width=.98\linewidth]{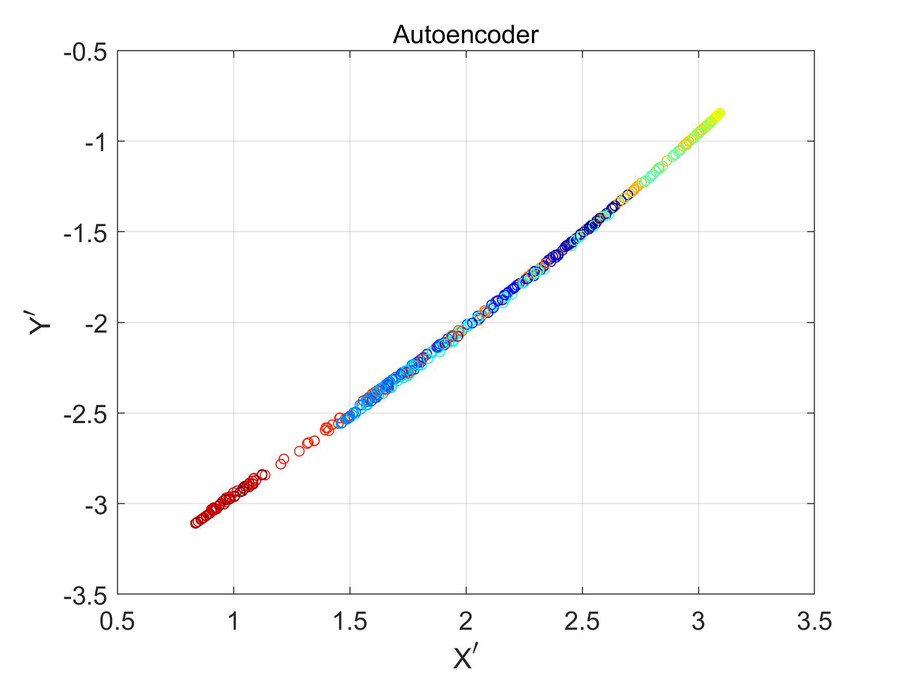}
			\caption{Autoencoder encoded output for Test Set}
			\label{fig:autoencoder_swiss_roll_ts1}
		\end{subfigure}%
		\caption{For experiment 1 on validation of predictability with the Swiss Roll data : (a) Autoencoder encoded output for Training Set,  and (b) Autoencoder encoded output for Test Set.}
		\label{fig:autoencoder_swiss_roll_out_of_sample}
	\end{figure*}

	\begin{figure*}[!thb]
		\centering
		\begin{subfigure}{.48\textwidth}
			\centering
			\includegraphics[width=.98\linewidth]{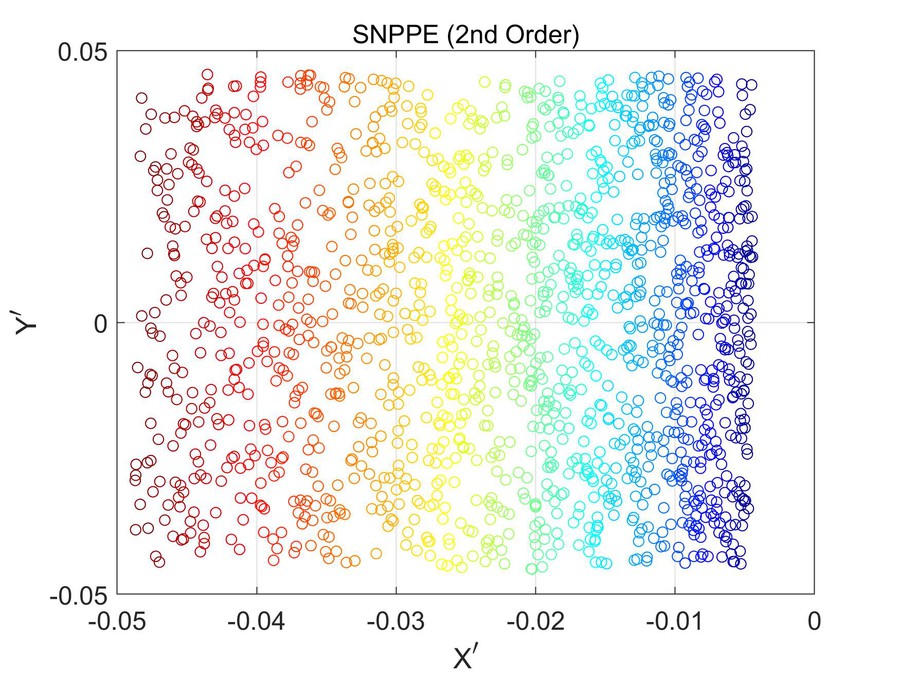}
			\caption{ 2nd Order SNPPE output for Training Set}
			\label{fig:SNPPE2_swiss_roll_tr1}
		\end{subfigure}
		\begin{subfigure}{.48\textwidth}
			\centering
			\includegraphics[width=.98\linewidth]{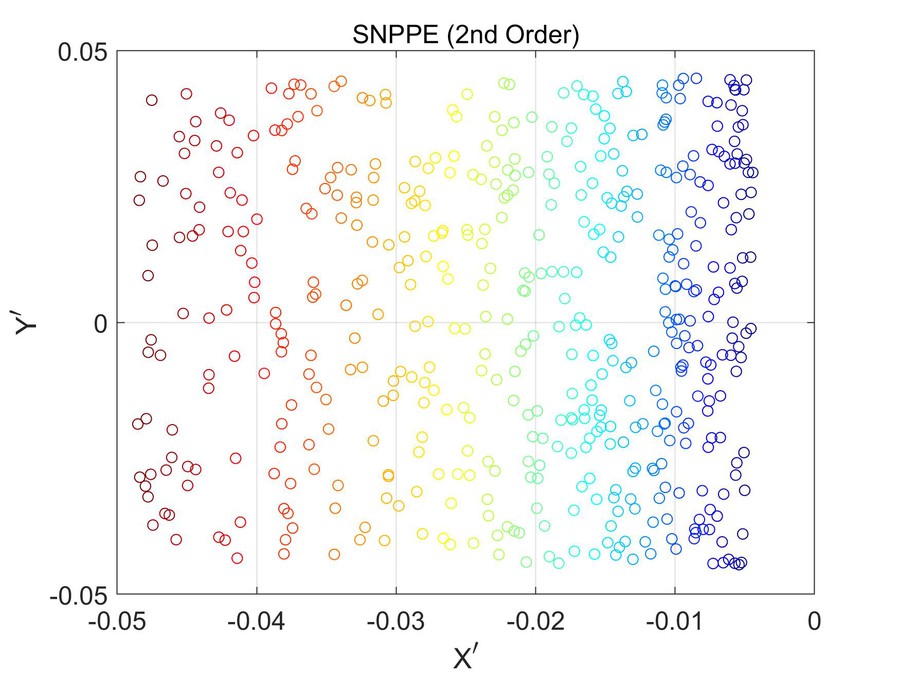}
			\caption{ 2nd Order SNPPE output for Test Set}
			\label{fig:SNPPE2_swiss_roll_ts1}
		\end{subfigure}%
		
		\caption{For experiment 1 on validation of predictability with the Swiss Roll data : (a) 2nd Order SNPPE output for Training Set,   (b) 2nd Order SNPPE output for Test Set.}
		\label{fig:SNPPE2_swiss_roll_out_of_sample}
	\end{figure*}

\begin{figure*}[!thb]
	\centering
	\begin{subfigure}{.48\textwidth}
		\centering
		\includegraphics[width=.98\linewidth]{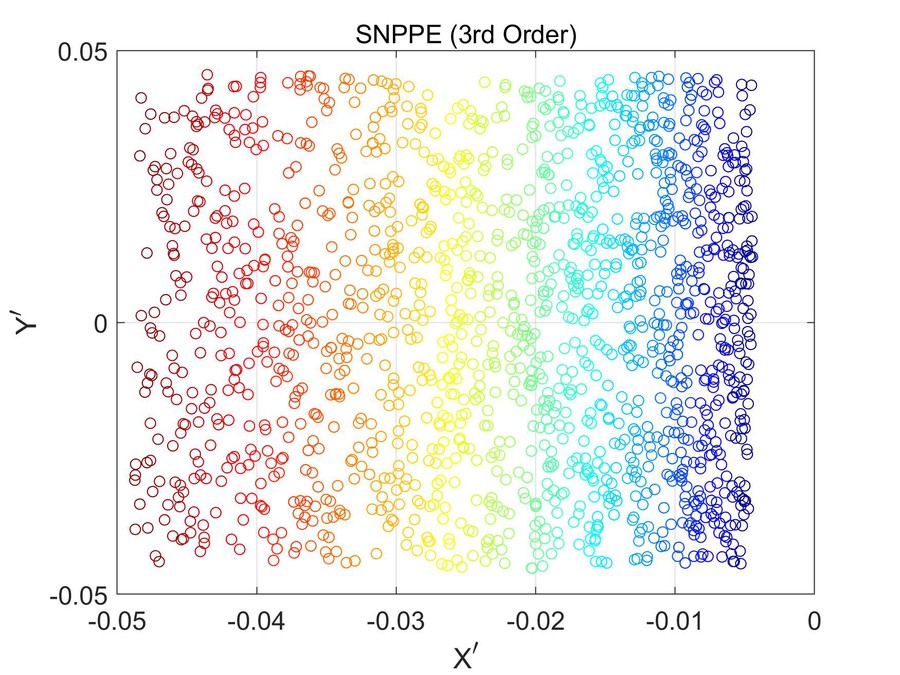}
		\caption{ 3rd Order SNPPE output for Training Set}
		\label{fig:SNPPE3_swiss_roll_tr1}
	\end{subfigure}%
	\begin{subfigure}{.48\textwidth}
		\centering
		\includegraphics[width=.98\linewidth]{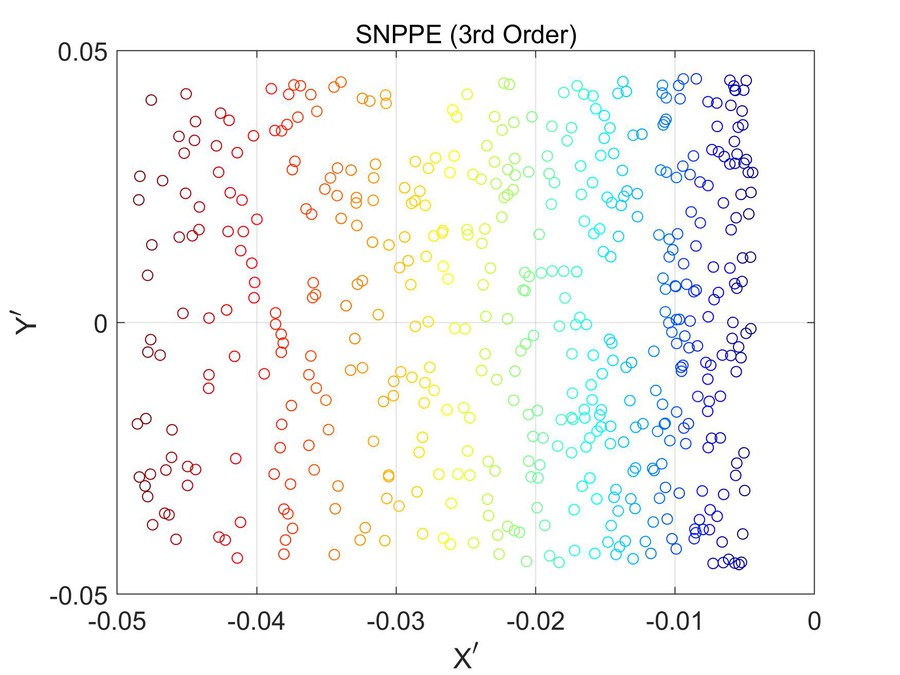}
		\caption{3rd Order SNPPE output for Test Set}
		\label{fig:SNPPE3_swiss_roll_ts1}
	\end{subfigure}
	\caption{For experiment 1 on validation of predictability with the Swiss Roll data : (a) 3rd Order SNPPE output for Training Set, (b) 3rd Order SNPPE output for Test Set.}
	\label{fig:SNPPE3_swiss_roll_out_of_sample}
\end{figure*}

	\begin{figure*}[!tb]
		\centering
		\includegraphics[width=0.95\linewidth]{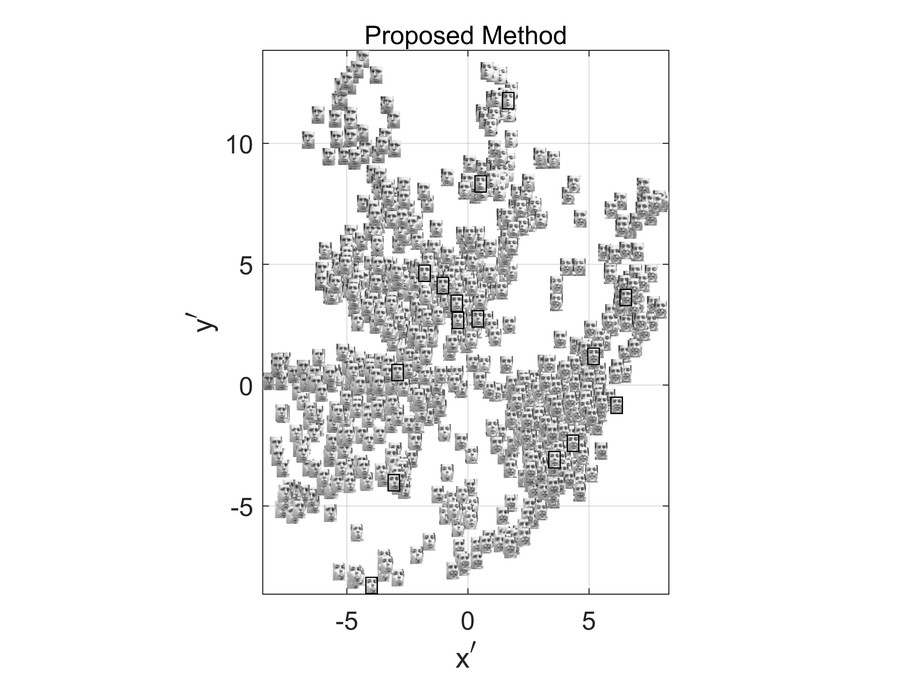}
		\caption{Visualization of the training and test set (black rectangle borders) for Frey face data set using the proposed method}
		\label{fig:frey_face_training_random_test}
	\end{figure*}

	\begin{figure*}[!tb]
		\centering
		\includegraphics[width=0.85\linewidth]{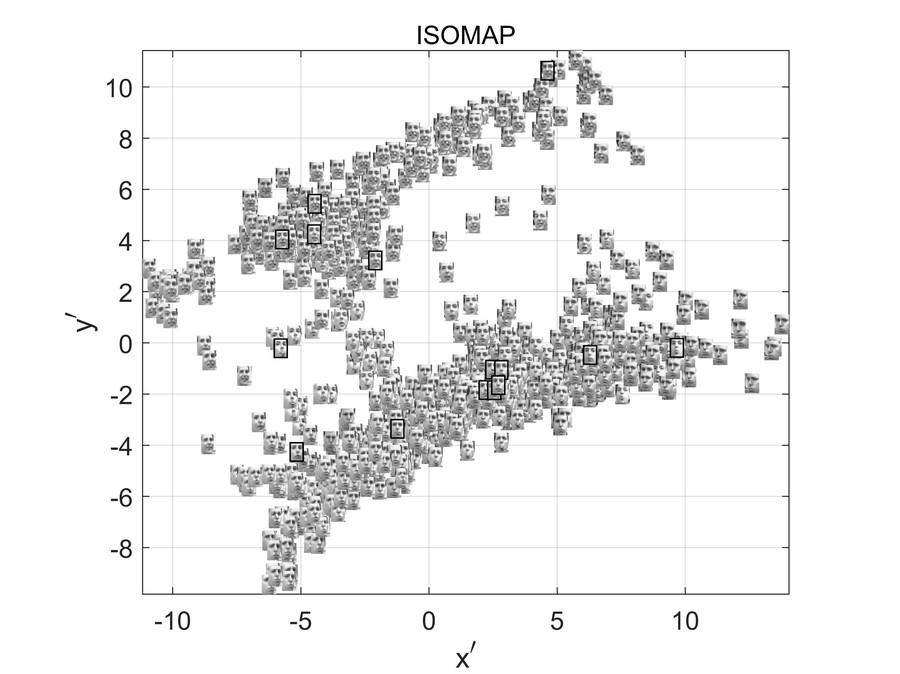}
		\caption{Visualization of the training set using ISOMAP and test set (black rectangle borders) with its Out of sample extension for Frey face data set}
		\label{fig:frey_face_ISOMAP_out_of_sample}
	\end{figure*}

	\begin{figure*}[!tb]
		\centering
		\includegraphics[width=0.95\linewidth]{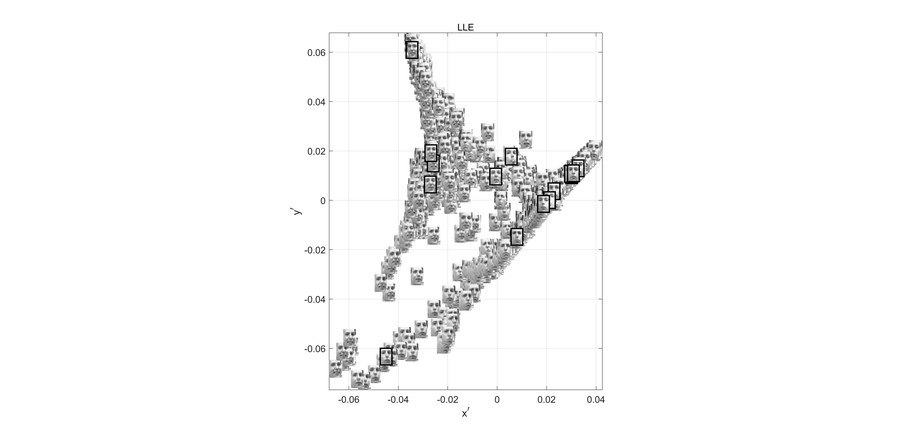}
		\caption{Visualization of the training set using LLE and test set (black rectangle borders) with its Out of sample extension for Frey face data set}
		\label{fig:frey_face_LLE_out_of_sample}
	\end{figure*}

	\begin{figure*}[!tb]
		\centering
		\includegraphics[width=0.95\linewidth]{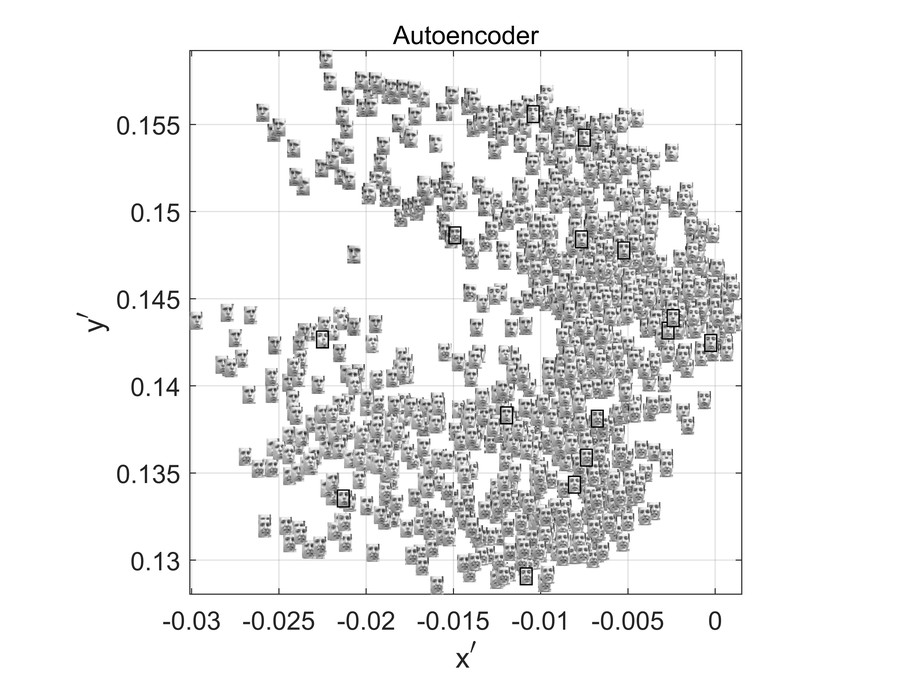}
		\caption{Visualization of the training set using Autoencoder encoded output and test set (black rectangle borders) for Frey face data set}
		\label{fig:frey_face_autoencoder}
	\end{figure*}

	\begin{figure*}[!tb]
		\centering
		\includegraphics[width=0.95\linewidth]{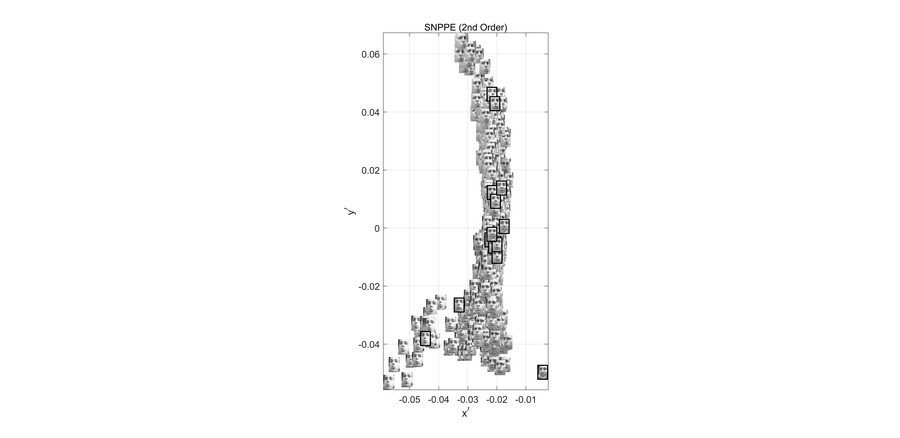}
		\caption{Visualization of the training set using 2nd Order SNPPE output and test set (black rectangle borders) for Frey face data set}
		\label{fig:frey_face_SNPPE2}
	\end{figure*}

	\begin{figure*}[!tb]
		\centering
		\includegraphics[width=0.95\linewidth]{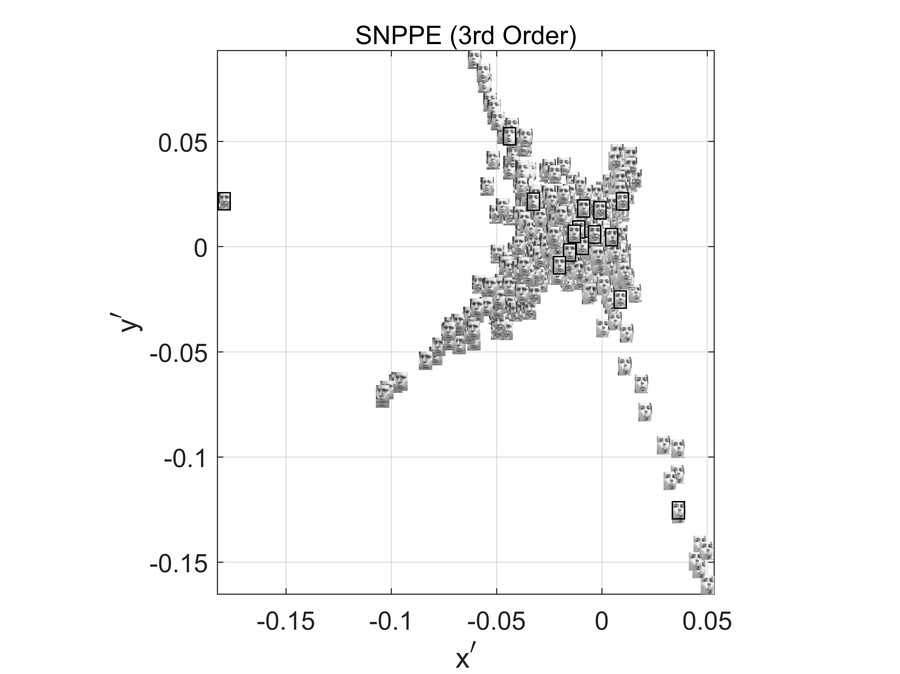}
		\caption{Visualization of the training set using 3rd Order SNPPE output and test set (black rectangle borders) for Frey face data set}
		\label{fig:frey_face_SNPPE3}
	\end{figure*}

	\begin{figure*}[!thb]
		\centering
		\begin{subfigure}{.48\textwidth}
			\centering
			\includegraphics[width=.98\linewidth]{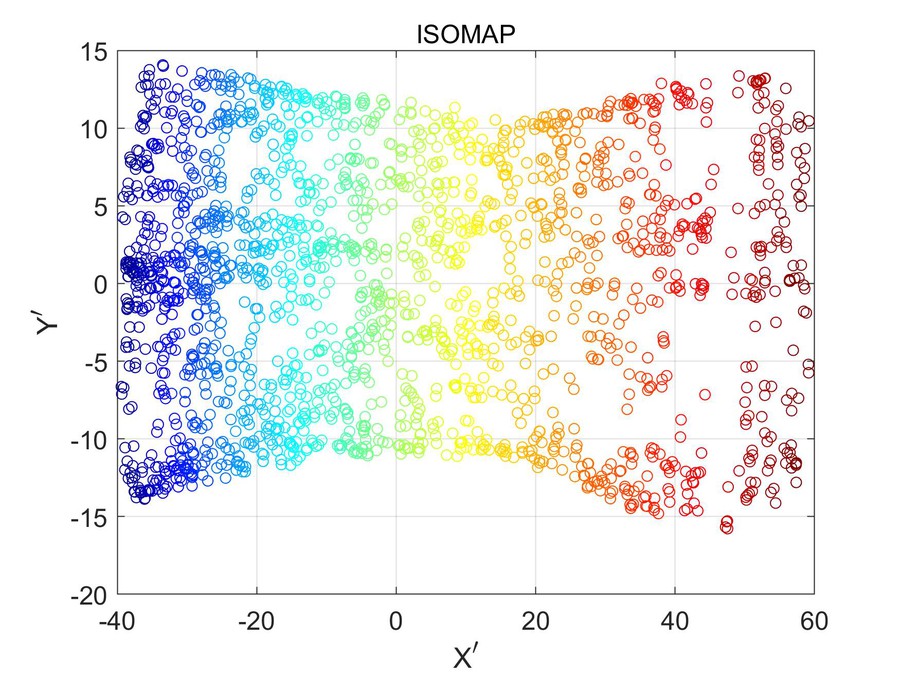}
			\caption{ ISOMAP output for Training Set}
			\label{fig:ISOMAP_swiss_roll_tr5}
		\end{subfigure}%
		\begin{subfigure}{.48\textwidth}
			\centering
			\includegraphics[width=.98\linewidth]{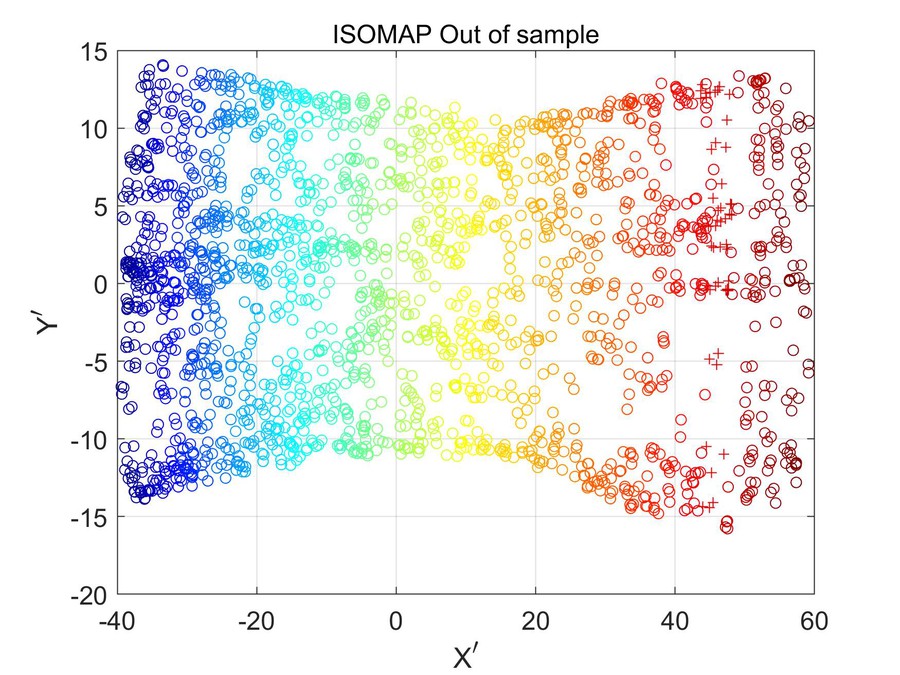}
			\caption{Out of sample extension of ISOMAP for Test Set}
			\label{fig:ISOMAP_swiss_roll_trts5}
		\end{subfigure}%
		\caption{For experiment 2 on validation of predictability with the Swiss Roll data : (a) ISOMAP output for Training Set,  and (b) Out of sample extension of ISOMAP for Training Set and Test Set.}
		\label{fig:ISOMAP_swiss_roll_out_of_sample1}
	\end{figure*}

	\begin{figure*}[!thb]
		\centering
		\begin{subfigure}{.48\textwidth}
			\centering
			\includegraphics[width=.98\linewidth]{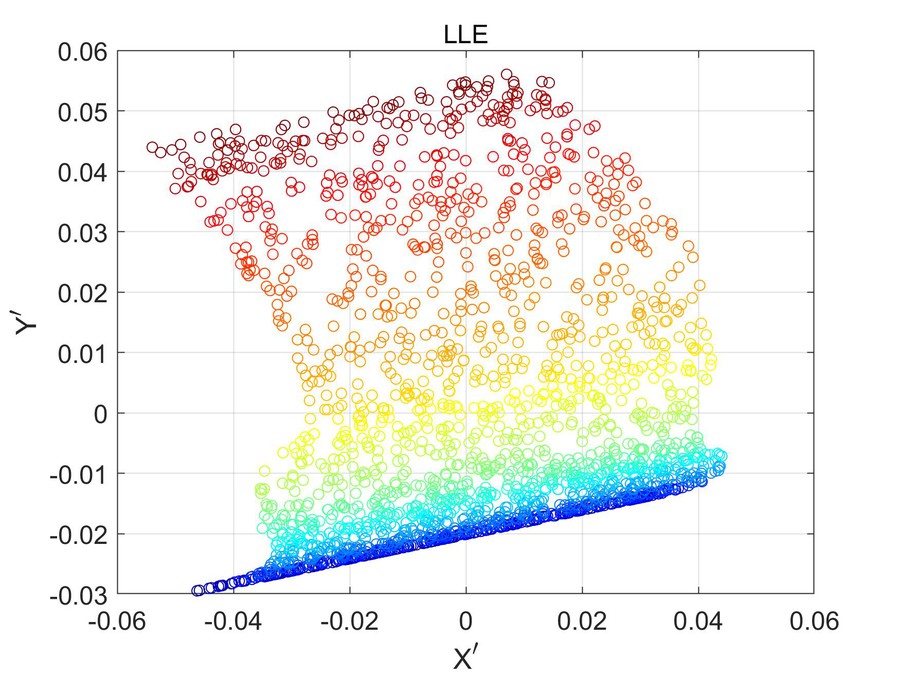}
			\caption{ LLE output for Training Set}
			\label{fig:LLE_swiss_roll_tr5}
		\end{subfigure}%
		\begin{subfigure}{.48\textwidth}
			\centering
			\includegraphics[width=.98\linewidth]{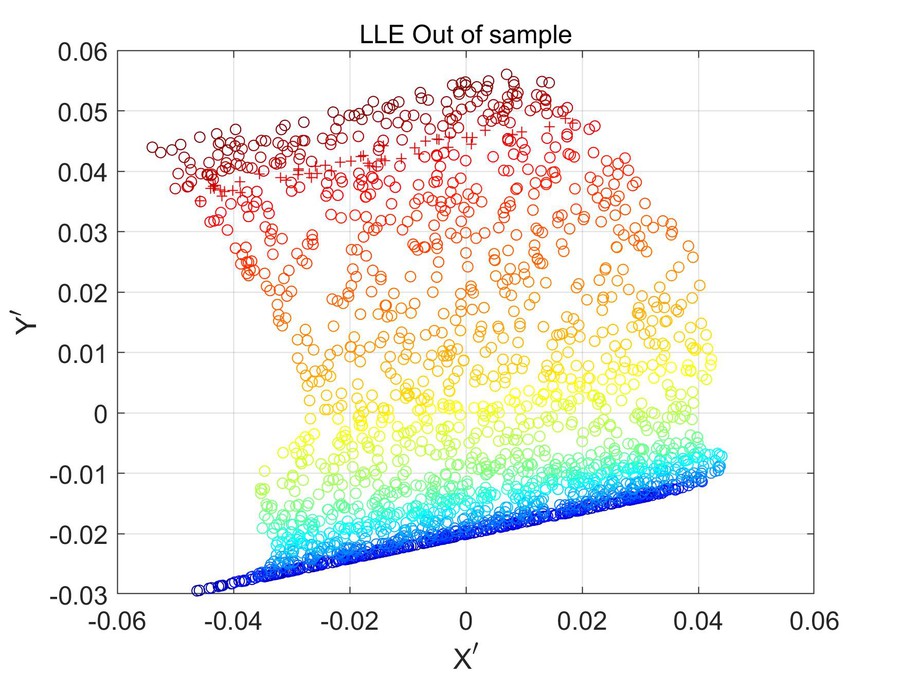}
			\caption{Out of sample extension of LLE for Test Set}
			\label{fig:LLE_swiss_roll_trts5}
		\end{subfigure}%
		\caption{For experiment 2 on validation of predictability with the Swiss Roll data : (a) LLE output for Training Set,  and (b) Out of sample extension of LLE for Training Set and Test Set.}
		\label{fig:LLE_swiss_roll_out_of_sample1}
	\end{figure*}

	\begin{figure*}[!thb]
		\centering
		\begin{subfigure}{.48\textwidth}
			\centering
			\includegraphics[width=.98\linewidth]{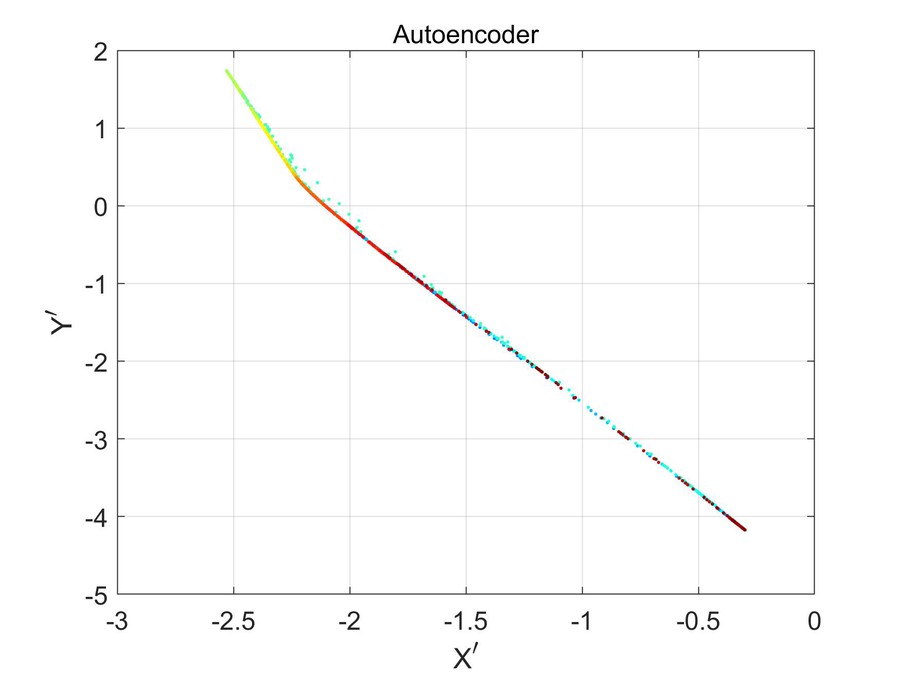}
			\caption{ Autoencoder encoded output for Training Set}
			\label{fig:Autoencoder_swiss_roll_tr5}
		\end{subfigure}%
		\begin{subfigure}{.48\textwidth}
			\centering
			\includegraphics[width=.98\linewidth]{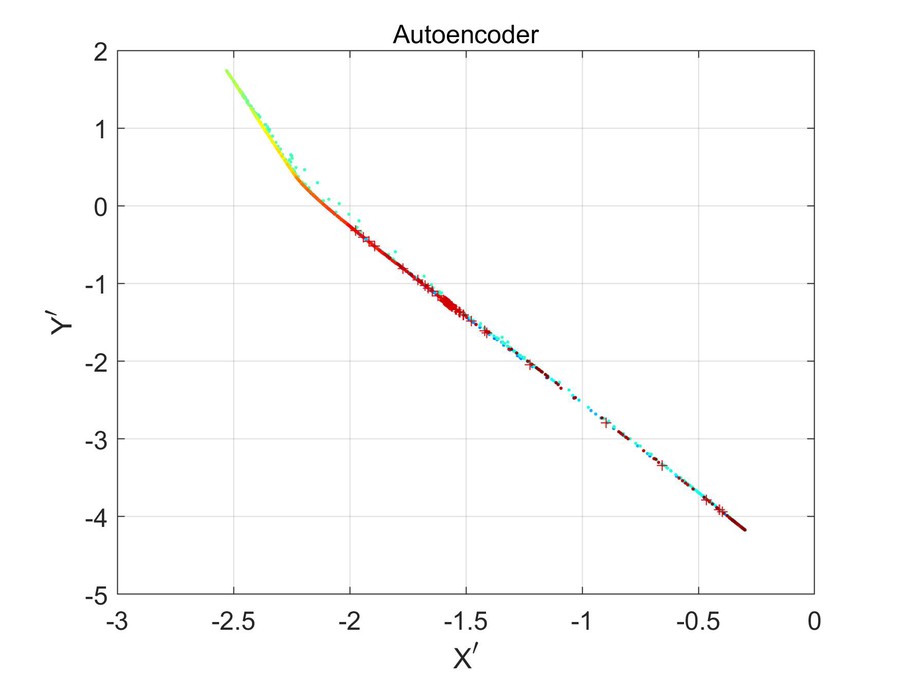}
			\caption{ Autoencoder encoded output for Test Set}
			\label{fig:Autoencoder_swiss_roll_trts5}
		\end{subfigure}%
		\caption{For experiment 2 on validation of predictability with the Swiss Roll data : (a) ISOMAP output for Training Set,  and (b) Out of sample extension of ISOMAP for Training Set and Test Set.}
		\label{fig:Autoencoder_swiss_roll_out_of_sample1}
	\end{figure*}

	\begin{figure*}[!thb]
		\centering
		\begin{subfigure}{.48\textwidth}
			\centering
			\includegraphics[width=.98\linewidth]{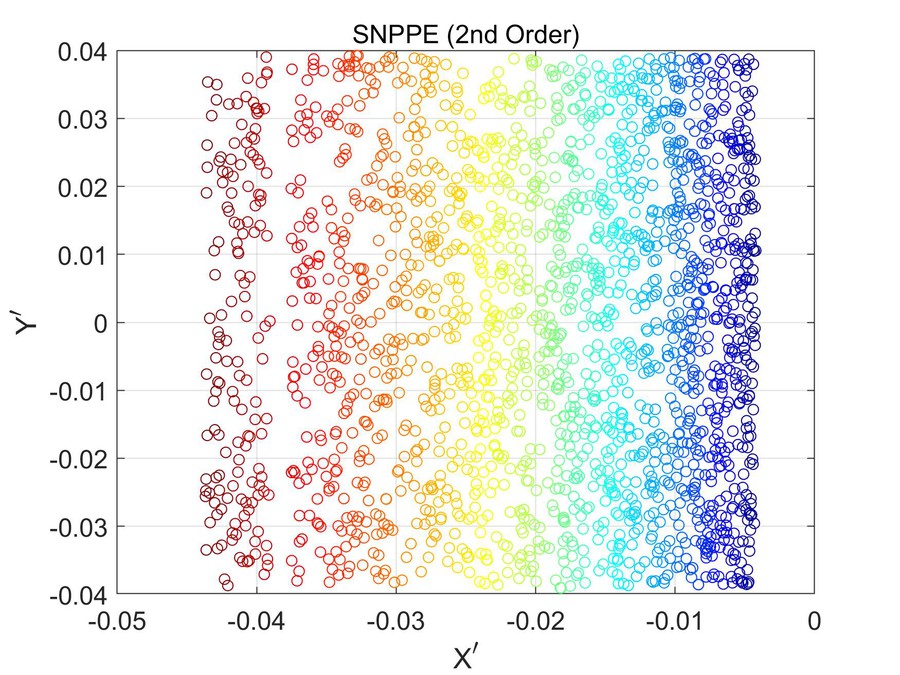}
			\caption{ 2nd Order SNPPE output for Training Set}
			\label{fig:SNPPE2_swiss_roll_tr5}
		\end{subfigure}%
		\begin{subfigure}{.48\textwidth}
			\centering
			\includegraphics[width=.98\linewidth]{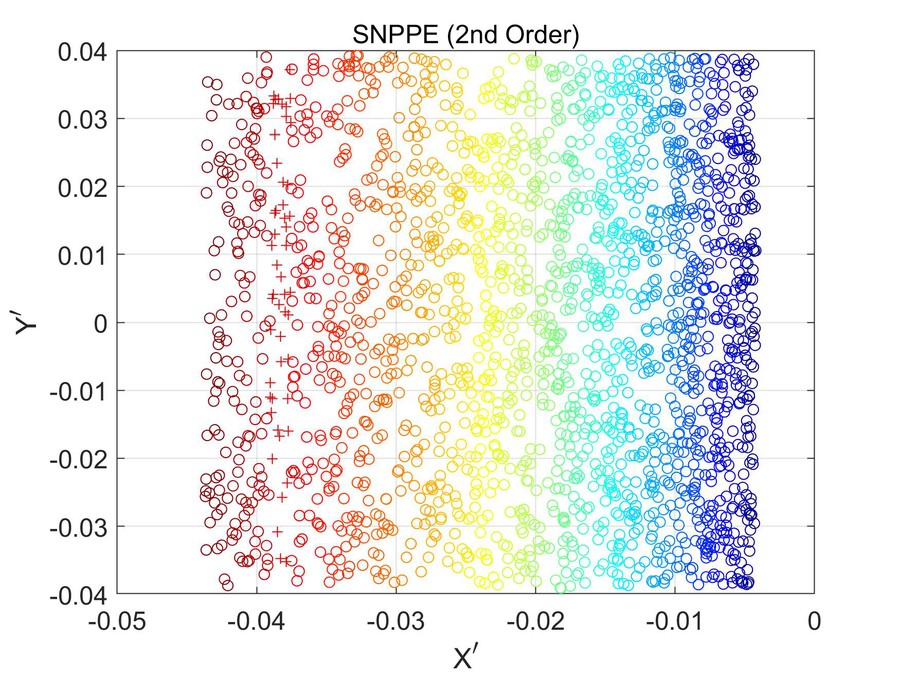}
			\caption{  2nd Order SNPPE encoded output for Test Set}
			\label{fig:SNPPE2_swiss_roll_trts5}
		\end{subfigure}%
		\caption{For experiment 2 on validation of predictability with the Frey face data : (a) 2nd Order SNPPE output for Training Set,  and (b) 2nd Order SNPPE for Training Set and Test Set.}
		\label{fig:SNPPE2_swiss_roll_out_of_sample1}
	\end{figure*}

	\begin{figure*}[!thb]
		\centering
		\begin{subfigure}{.48\textwidth}
			\centering
			\includegraphics[width=.98\linewidth]{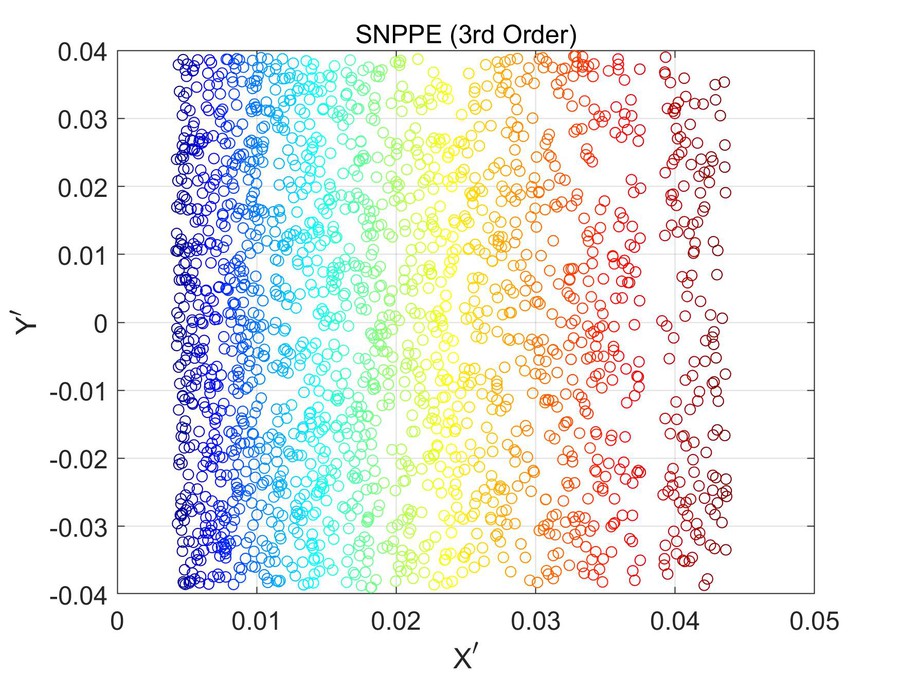}
			\caption{ 3rd Order SNPPE output for Training Set}
			\label{fig:SNPPE3_swiss_roll_tr5}
		\end{subfigure}%
		\begin{subfigure}{.48\textwidth}
			\centering
			\includegraphics[width=.98\linewidth]{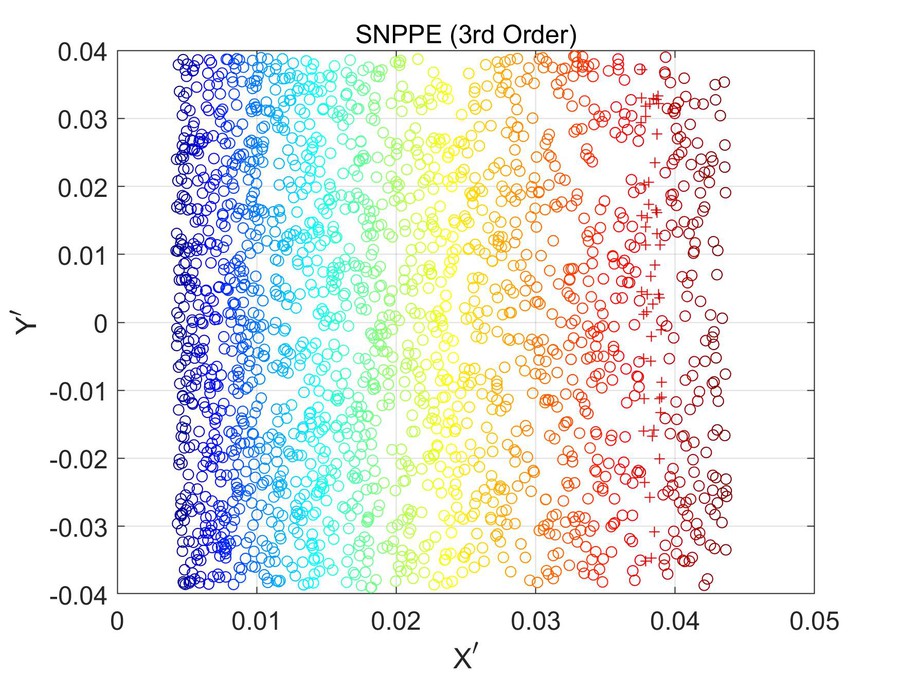}
			\caption{  3rd Order SNPPE encoded output for Test Set}
			\label{fig:SNPPE3_swiss_roll_trts5}
		\end{subfigure}%
		\caption{For experiment 2 on validation of predictability with the Frey face data : (a) 3rd Order SNPPE output for Training Set,  and (b) 3rd Order SNPPE for Training Set and Test Set.}
		\label{fig:SNPPE3_swiss_roll_out_of_sample1}
	\end{figure*}
\begin{figure*}[!tb]
	\centering
	\begin{subfigure}{.48\textwidth}
		\centering
		\includegraphics[width=.98\linewidth]{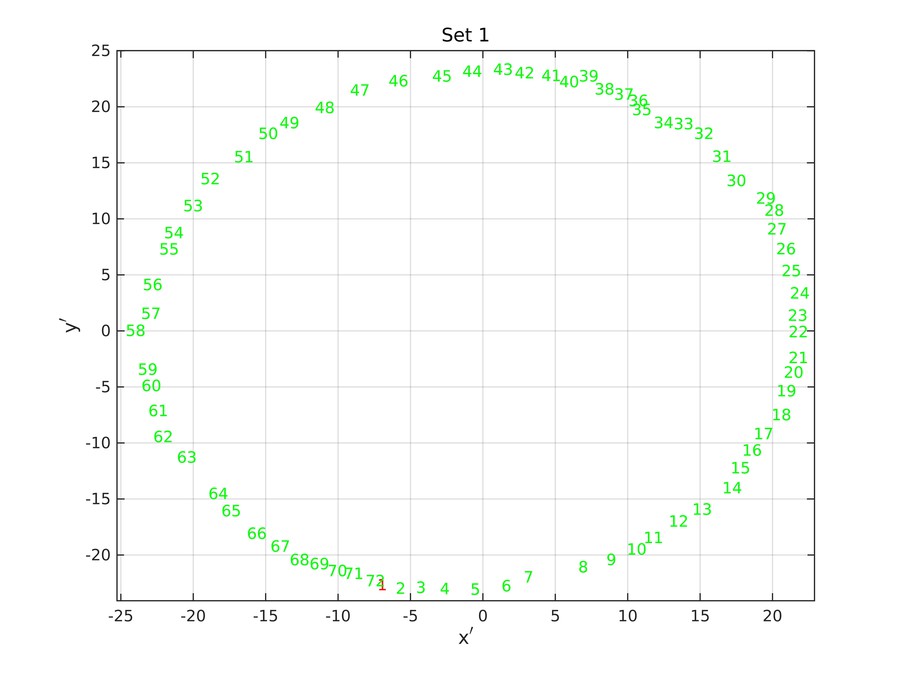}%
		\caption{ Proposed Method Output for Set 1}
		\label{fig: Set 1}
	\end{subfigure}%
	\begin{subfigure}{.48\textwidth}
		\centering
		\includegraphics[width=.98\linewidth]{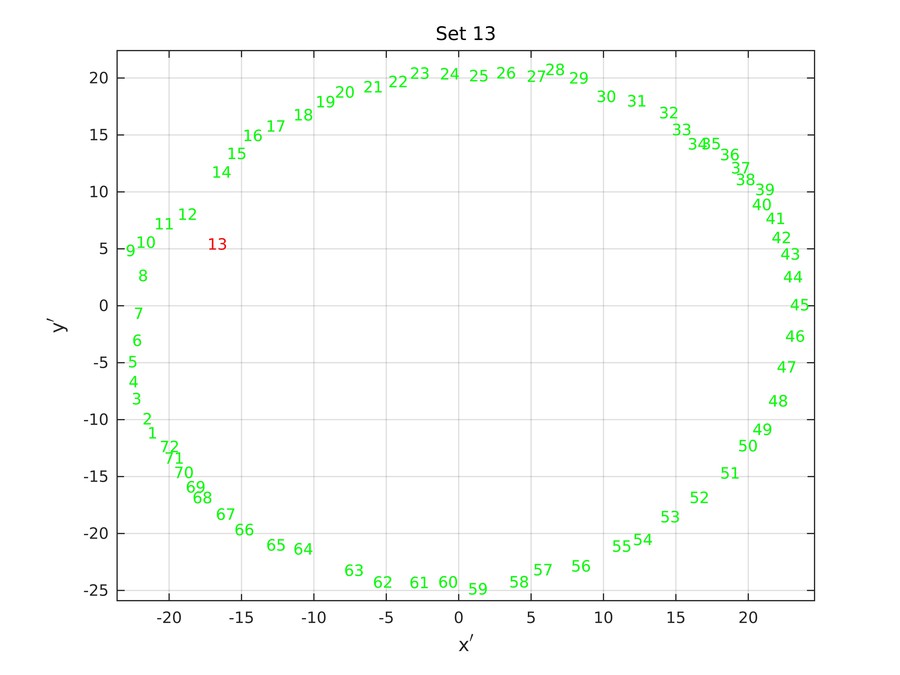}
		\caption{Proposed Method Output for Set 2}
		\label{fig: Set 2}
	\end{subfigure}%
	\\
	\begin{subfigure}{.48\textwidth}
		\centering
		\includegraphics[width=.98\linewidth]{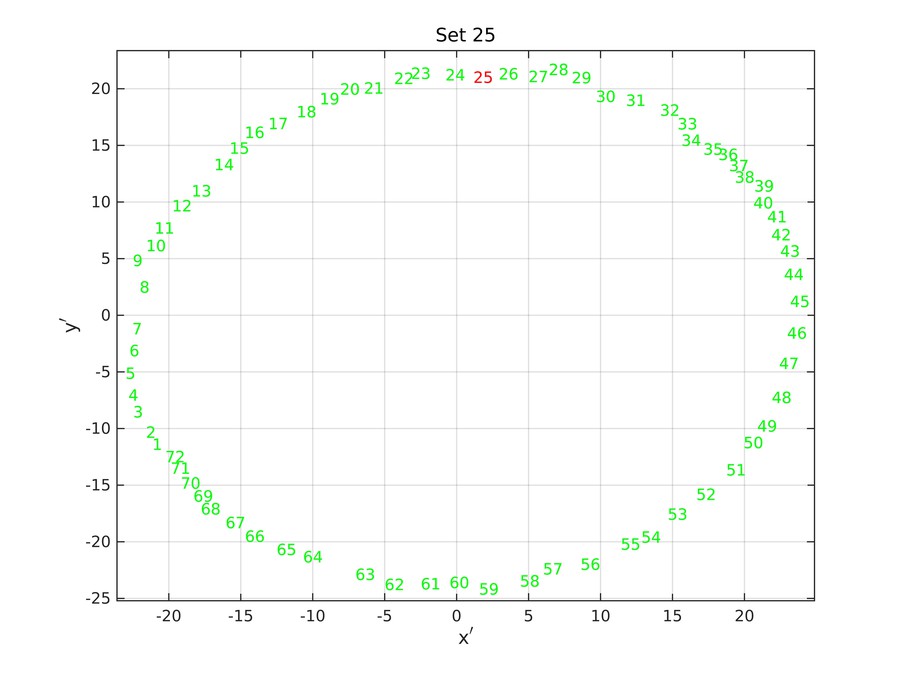}
		\caption{Proposed Method Output for Set 3}
		\label{fig: Set 3}
	\end{subfigure}%
	\begin{subfigure}{.48\textwidth}
		\centering
		\includegraphics[width=.98\linewidth]{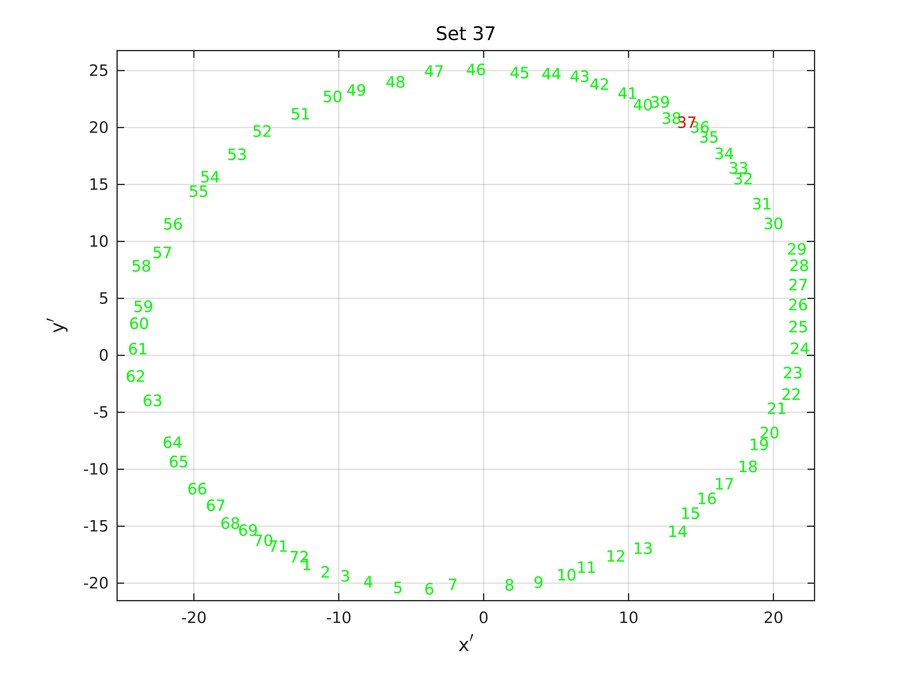}
		\caption{Proposed Method Output for Set 4}
		\label{fig: Set 4}
	\end{subfigure}
	\\
	\begin{subfigure}{.48\textwidth}
		\centering
		\includegraphics[width=.98\linewidth]{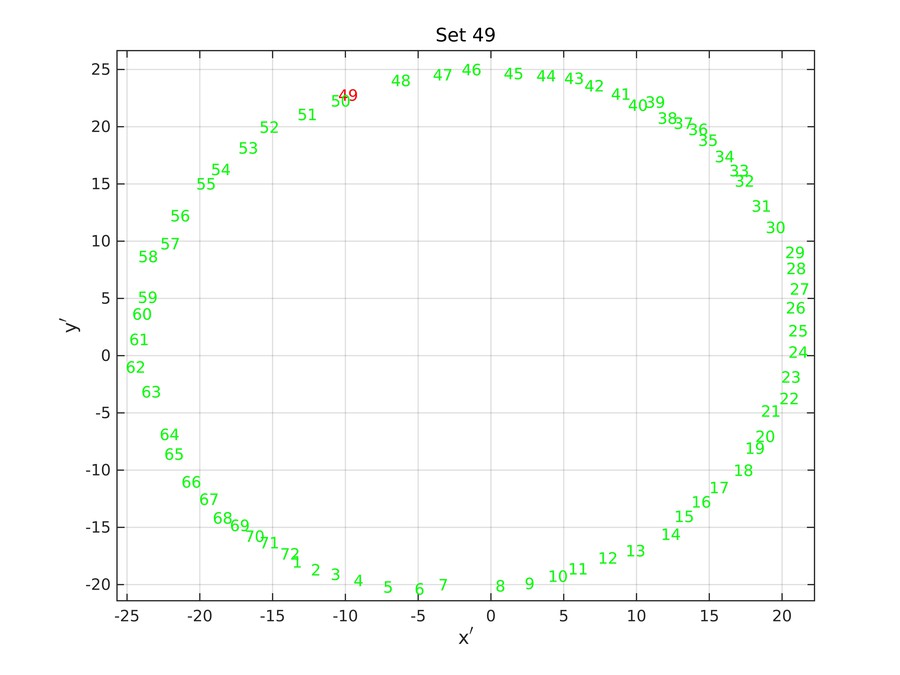}
		\caption{Proposed Method Output for Set 5}
		\label{fig: Set 5}
	\end{subfigure}%
	\begin{subfigure}{.48\textwidth}
		\centering
		\includegraphics[width=.98\linewidth]{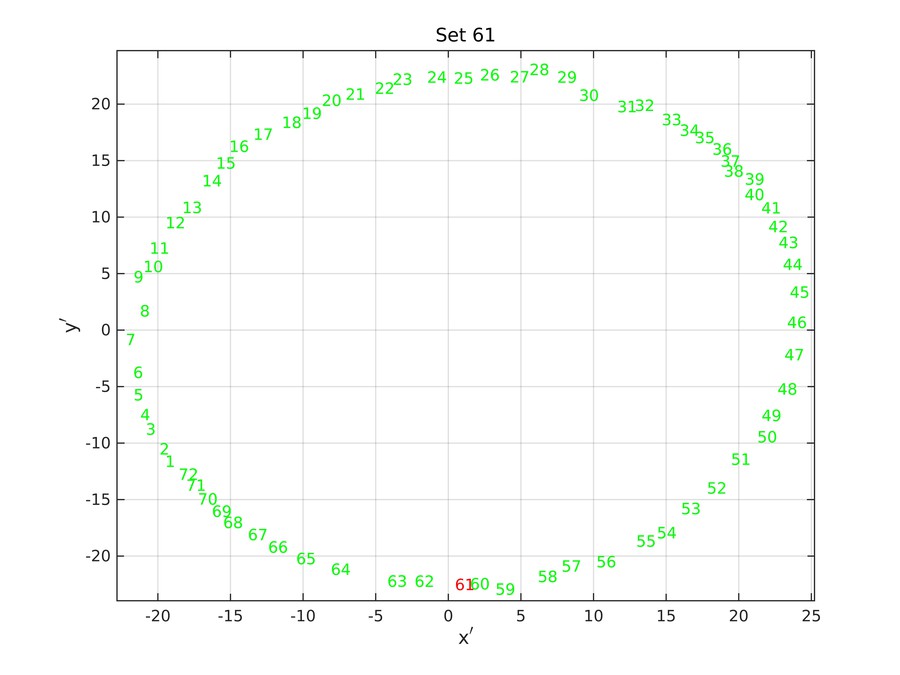}
		\caption{Proposed Method Output for Set 6 }
		\label{fig: Set 6}
	\end{subfigure}
	\caption{ For experiment 3 on validation of predictability with leave one out sets composed of the first object of the COIL data set, proposed method outputs corresponding to : (a) Set 1, (b) Set 2, (c) Set 3, (d) Set 4, (e) Set 5, and (f) Set 6 . }
	\label{fig:coil_dataset_leave_one_out_result_set}
\end{figure*}

\begin{figure*}[!tbh]
	\centering
	\begin{subfigure}{.289\textwidth}
		\centering
		\includegraphics[width=.9\linewidth]{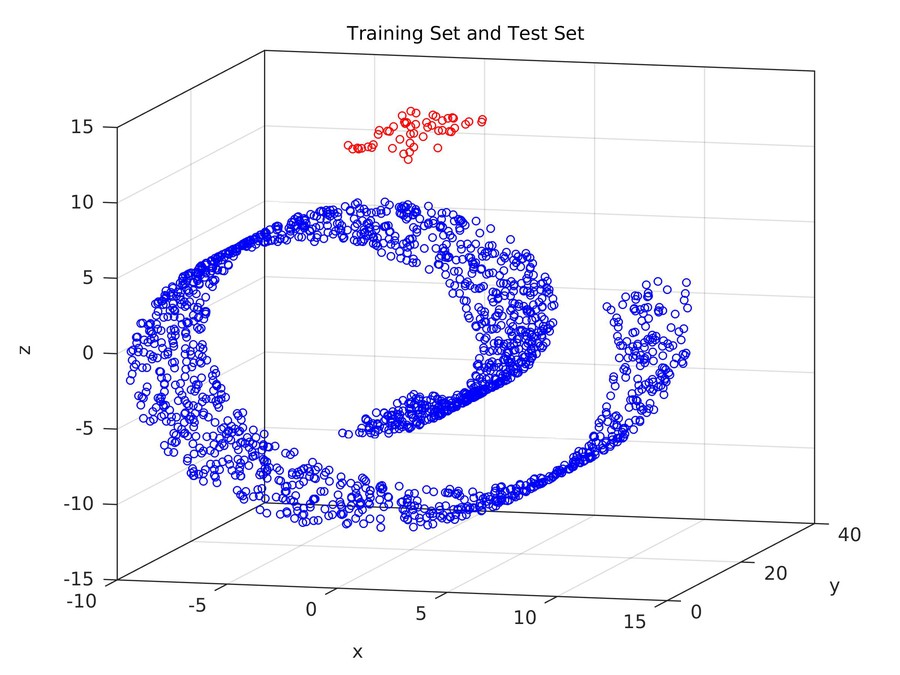}%
		\caption{ Training Set (blue circles) and Test Set (red circles) }
		\label{fig:swiss roll training set 3 and test set 3}
	\end{subfigure}
	\begin{subfigure}{.289\textwidth}
		\centering
		\includegraphics[width=.9\linewidth]{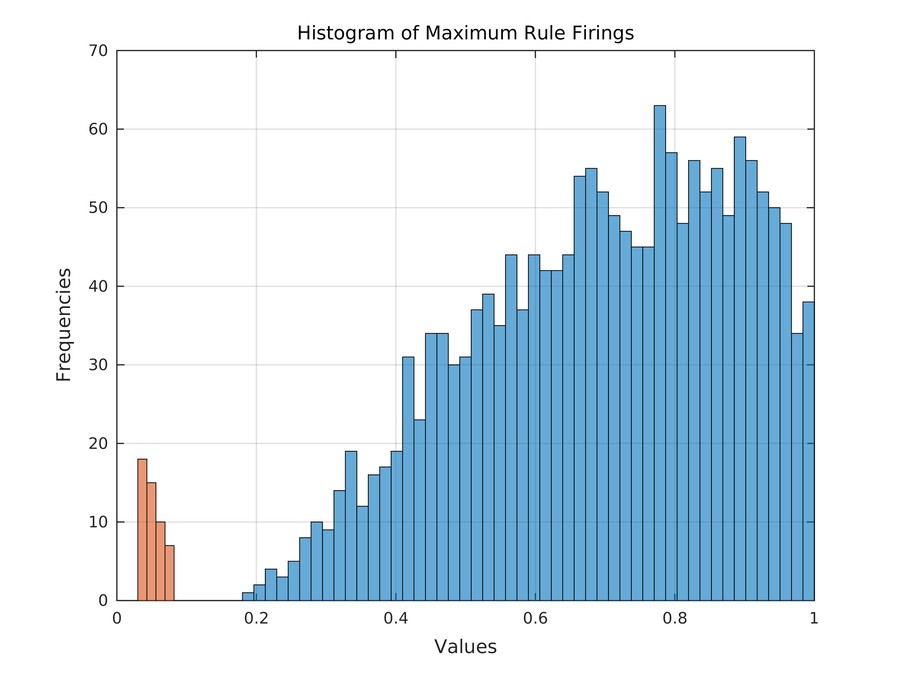}%
		\caption{ Histogram of maximum rule firings for Training Set (blue region) and Test Set (red region) }
		\label{fig: Histogram of max rule firing}
	\end{subfigure}
	\begin{subfigure}{.289\textwidth}
		\centering
		\includegraphics[width=.9\linewidth]{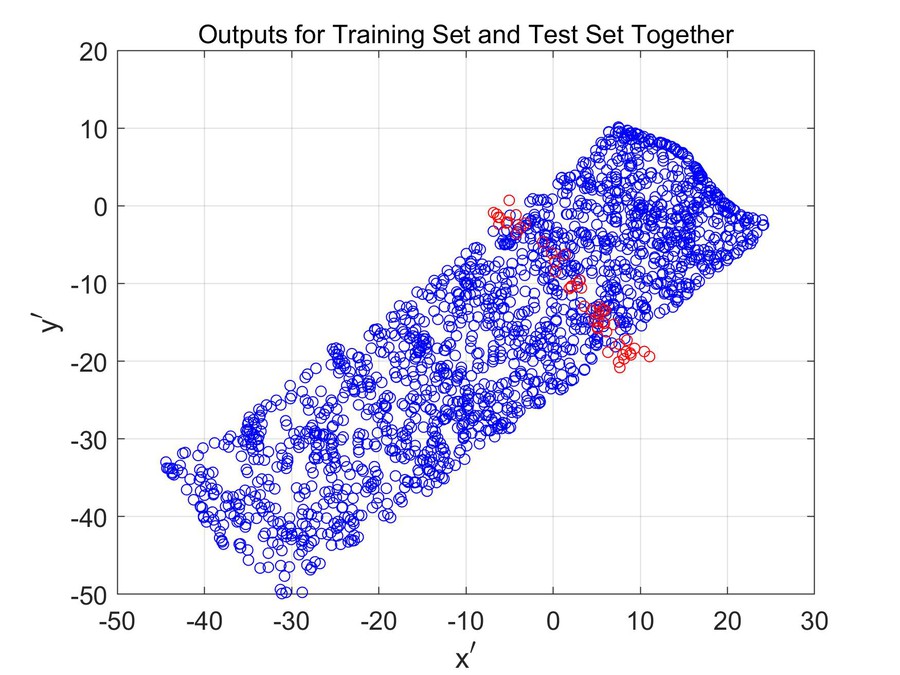}%
		\caption{ Proposed Method Outputs for Training Set (blue circles) and Test Set (red circles) without exercising rejection option}
		\label{fig:undesired_output}
	\end{subfigure}
	\caption{ (a) Training Set (blue circles) and Test Set (red circles), (b) Histogram of maximum rule firings for Training Set (blue region) and Test Set (red region) and (c) Proposed Method Outputs for Training Set (blue circles) and Test Set (red circles) without exercising rejection option.  }
	\label{fig:expt 1 on prediction on validity of the outputs }
\end{figure*}
\begin{figure*}[!tb]
	\begin{subfigure}{.48\textwidth}
		\centering
		\includegraphics[width=.98\linewidth]{raw_swiss_roll_2000_ver1}
		\caption{}%
		\label{fig:swiss roll1}
	\end{subfigure}
	\begin{subfigure}{.48\textwidth}
		\centering
		\includegraphics[width=.98\linewidth]{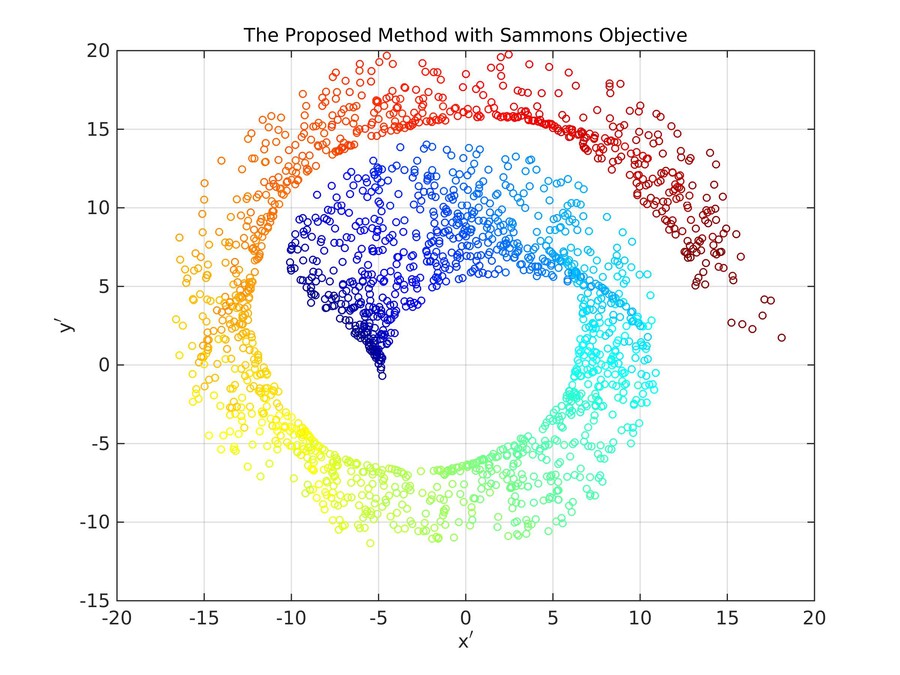}
		\caption{}%
		\label{fig: proposed with sammons obj }
	\end{subfigure}
	\caption{Visualization of Swiss Roll data with: (a) original input space, and (b) proposed method with Sammon's Objective. }
	\label{fig:generalization of proposed model}
\end{figure*}
\begin{figure*}[!tb]
	\centering
	\begin{subfigure}{.48\textwidth}
		\centering
		\includegraphics[width=.98\linewidth]{raw_s_curve_2000}
		\caption{ Original Data}
		\label{fig:SCurve}
	\end{subfigure}
	\begin{subfigure}{.48\textwidth}
		\centering
		\includegraphics[width=.98\linewidth]{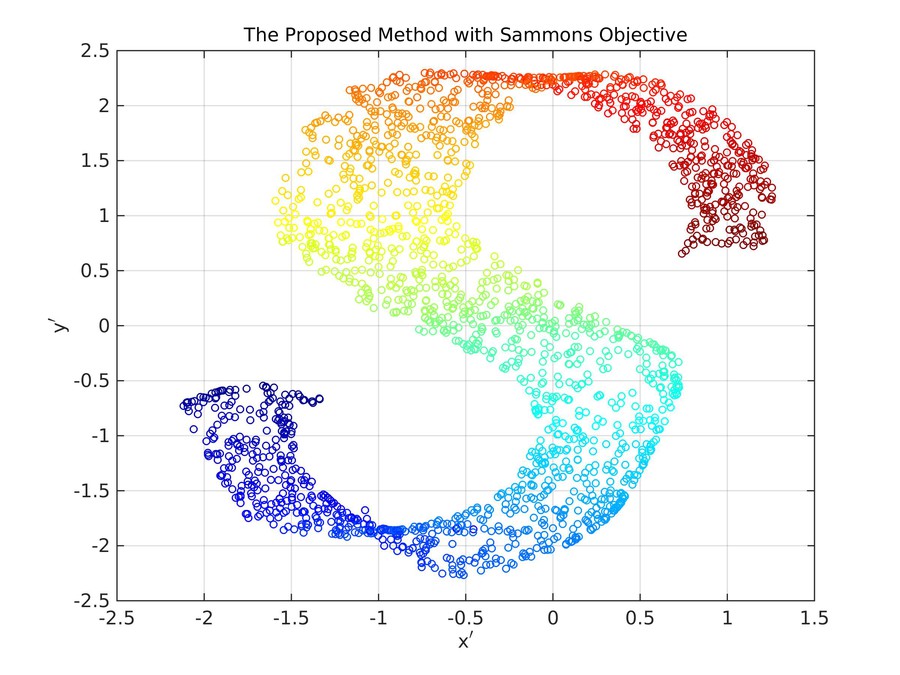}
		\caption{ Proposed Method with Sammon's Objective }
		\label{fig:SCurveSammonsWithProposedMethod}
	\end{subfigure}
	\caption{Visualization of S Curve data with: (a) original input space, and (b) proposed method with Sammon's objective.  }
	\label{fig:SCurveProposedMethodGeneralization}
\end{figure*}

\begin{figure*}[!tb]
	\centering
	\begin{subfigure}{.48\textwidth}
		\centering
		\includegraphics[width=.98\linewidth]{raw_helix_2000}
		\caption{ Original Data}
		\label{fig:Helix}
	\end{subfigure}
	\begin{subfigure}{.48\textwidth}
		\centering
		\includegraphics[width=.98\linewidth]{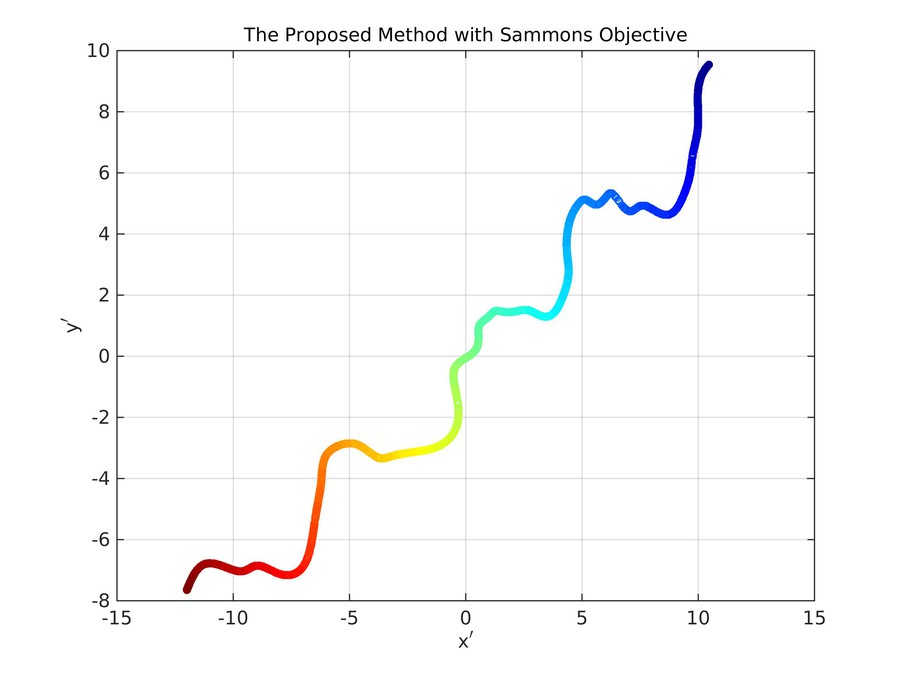}
		\caption{ Proposed Method with Sammon's Objective }
		\label{fig:HelixSammonsWithProposedMethod}
	\end{subfigure}
	\caption{Visualization of Helix data with: (a) original input space, and (b) proposed method with Sammon's objective.  }
	\label{fig:HelixProposedMethodGeneralization}
\end{figure*}
\makeatletter\@input{other1.tex}\makeatother